\documentclass[lettersize,journal]{IEEEtran}

\usepackage{setspace}
\usepackage{indentfirst}
\usepackage[numbers,sort&compress]{natbib}
\usepackage{xcolor}
\usepackage[pdftex]{graphicx}
\usepackage{amsmath}
\usepackage{amsthm}
\usepackage{amssymb}
\usepackage[ruled,linesnumbered]{algorithm2e}
\usepackage{array}
\usepackage[caption=true,font=footnotesize]{subfig}
\usepackage{fixltx2e}
\usepackage{stfloats}
\usepackage{url}
\usepackage{comment}
\usepackage{multirow}
\usepackage{ulem}
\usepackage{verbatim}
\allowdisplaybreaks[2]

\newtheorem{thm}{Theorem}
\newtheorem{lem}{Lemma}

\newtheorem{assum}{Assumption}
\newtheorem{cor}{Corollary}
\theoremstyle{definition}
\newtheorem{defn}{Definition}

\usepackage[T1]{fontenc}

\begin{document}
\title{Semi-Synchronous Personalized Federated Learning over Mobile Edge Networks}
\author{Chaoqun You,~\IEEEmembership{Member,~IEEE,}
        Daquan Feng,~\IEEEmembership{Member,~IEEE,}
        Kun Guo,~\IEEEmembership{Member,~IEEE,}\\
        Howard H. Yang~\IEEEmembership{Member,~IEEE,}
        Chenyuan Feng,
        and~Tony~Q.~S.~Quek,~\IEEEmembership{Fellow,~IEEE}
\thanks{This paper was supported in part by the National Research Foundation, Singapore and Infocomm Media Development Authority under its Future Communications Research \& Development Programme, in part by MOE ARF Tier 2 under Grant T2EP20120$-$0006, in part by the National Science and Technology Major Project under Grant 2020YFB1807601, in part by the Shenzhen Science and Technology Program under Grants JCYJ20210324095209025, in part by Shanghai Pujiang Program under Grant No. 21PJ1402600, in part by the National Natural Science Foundation of China under Grant 62201504, in part by the Zhejiang Provincial Natural Science Foundation of China under Grant LGJ22F010001. \textit{(Corresponding author: Daquan Feng)}}
\thanks{C. You and T. Quek are with the Wireless Networks and Design Systems Group, Singapore University of Design and Technology, 487372, Singapore (e-mail: chaoqun\_you, tonyquek@sutd.edu.sg).}
\thanks{D. Feng and C. Feng are with the Shenzhen University, Shenzhen 518052, China (e-mail:fdquan, fengchenyuan@szu.edu.cn)}
\thanks{K. Guo is with the East China Normal University, Shanghai 200241, China (e-mail: kguo@cee.ecnu.edu.cn).}
\thanks{H. H. Yang is with the Zhejiang University/University of Illinois at Urbana-Champaign Institute, Zhejiang University, Haining 314400, China (email: haoyang@intl.zju.edu.cn).}}

\markboth{Journal of \LaTeX\ Class Files,~Vol.~14, No.~8, August~2015}%
{Shell \MakeLowercase{\textit{et al.}}: Bare Demo of IEEEtran.cls for IEEE Journals}

\maketitle
\normalem
\begin{abstract}
Personalized Federated Learning (PFL) is a new Federated Learning (FL) approach to address the heterogeneity issue of the datasets generated by distributed user equipments (UEs). However, most existing PFL implementations rely on synchronous training to ensure good convergence performances, which may lead to a serious straggler problem, where the training time is heavily prolonged by the slowest UE. To address this issue, we propose a semi-synchronous PFL algorithm, termed as Semi-Synchronous Personalized FederatedAveraging (PerFedS$^2$), over mobile edge networks. By jointly optimizing the wireless bandwidth allocation and UE scheduling policy, it not only mitigates the straggler problem but also provides convergent training loss guarantees. We derive an upper bound of the convergence rate of PerFedS$^2$ in terms of the number of participants per global round and the number of rounds. On this basis, the bandwidth allocation problem can be solved using analytical solutions and the UE scheduling policy can be obtained by a greedy algorithm. Experimental results verify the effectiveness of PerFedS$^2$ in saving the training time as well as guaranteeing the convergence of training loss, in contrast to synchronous and asynchronous PFL algorithms.
\end{abstract}


\IEEEpeerreviewmaketitle

\section{Introduction} \label{sec:1}

\IEEEPARstart{F}{ederated} Learning (FL) is a new distributed machine learning paradigm that enables model training across multiple user equipments (UEs) without uploading their raw data to a central parameter server~\cite{mcmahan2017communication}.
Since its advent, FL has been widely adopted as a powerful tool to exploit the wealth of data available at the end-user devices~\cite{yang2021federated, yang2020energy} and foster new applications such as Artificial Intelligence (AI) medical diagnosis~\cite{rieke2020future} and autonomous vehicles~\cite{xiao2021vehicle}.
Training a FL model contains three typical steps: ($i$) a set of UEs conduct local computing based on their own dataset, and upload the resultant parameters to the server, ($ii$) the server aggregates the UEs' parameters and improve the global model, and ($iii$) the server feeds back the new model to UEs for another round of local computing.
This procedure repeats until the loss function starts to converge and a certain model accuracy is achieved.
%
%

With the substantial improvement in sensing capabilities and computational power of edge devices, UEs are producing abundant but diverse data~\cite{song2020artificial}.
The increasingly diverse datasets breed a demand for customized services on individual UEs.
Typical examples of potential applications include Vehicle-to-everything (V2X) communications, where vehicles in the network may experience various road conditions and driving habits, making the local model disparate to the global model~\cite{samarakoon2019distributed, prathiba2021cybertwin}; and recommendation systems, where local servers have potentially heterogeneous customers and share non-independent and identically distributed (non-i.i.d.) item popularities, and thus requiring fine-grained recommendations~\cite{yang2020federated, wang2021fast}.
However, conventional FL algorithms are proposed to learn a \textit{common} model which may have mediocre performance on certain UEs.
And the situation is exacerbating as the ever-developing mobile UEs are generating increasingly diverse data.
To address this issue, Personalized Federated Learning (PFL)~\cite{dinh2020personalized, jiang2019improving} has been proposed.
Specifically, PFL provides an \textit{initial} model that is good enough for the UEs to start with.
Using this initial model, each UE can fastly adapt to its local dataset with one or more gradient descent steps using only a few data points. As a result, the UEs (especially with heterogeneous datasets) are able to enjoy fast personalized models by adapting the global model to local datasets.

Nonetheless, most PFL implementations adopt synchronous training to ensure good convergence performance~\cite{fallah2020personalized, dinh2020personalized, deng2020adaptive, shamsian2021personalized, achituve2021personalized}.
In the synchronous setting, the central server has to wait until the arrival of the parameters of the slowest UE before it can update the global model.
As a consequence, synchronous training may cause severe \textit{straggler} problem in PFL, where the deceleration of any UE can delay all other UEs.
On the other hand, parameters of the UEs may arrive at the server at different speeds due to reasons such as various CPU processing capabilities and different wireless channel conditions.
This difference begets another operation mechanism: asynchronous training.
The key idea of asynchronous implementation is to allow all UEs work independently and the server updates the global model every time it receives an update from any UE~\cite{lian2015asynchronous, xu2021asynchronous, chen2020asynchronous}.
Although this model updating strategy avoids the waiting time of UEs, the gradient staleness caused by asynchronous updating will further degrade the performance of the model training.
At this point, a semi-synchronous PFL has been a natural choice to balance the disadvantages caused by the synchronous as well as the asynchronous PFL algorithms.

Although there have been several works on semi-synchronous FL algorithms~\cite{wu2020safa, ma2021fedsa, stripelis2021semi, zhang2021csafl}, the semi-synchronous PFL problem is not well understood.
\cite{wu2020safa} studied the semi-asynchronous protocol for fast FL. \cite{ma2021fedsa} proposed a semi-asynchronous FL algorithm in heterogeneous edge computing. \cite{stripelis2021semi} introduced a novel energy-efficient semi-asynchronous FL protocol that mixes local models periodically with minimal idle time and fast convergence. At last, \cite{zhang2021csafl} proposed a clustered semi-asynchronous FL algorithm that groups UEs by the delay and direction of clients' model update to make the most of the advantage of both synchronous and asynchronous FL.
Designing a semi-synchronous PFL in mobile edge networks, however, is particularly
challenging due to the following reasons: (1) The convergence rate of a semi-synchronous PFL is \textit{unclear}. Moreover, the loss function of a deep learning model is usually \textit{non-convex}, and whether a semi-synchronous PFL can converge and under what conditions can the algorithm converge is of much interest. (2) The practical wireless communication environments need to be considered. It is \textit{non-trivial} to decide the UE scheduling policy of a semi-synchronous PFL algorithm while considering the wireless bandwidth allocation.

In this paper, we propose a semi-synchronous PFL algorithm over mobile edge networks, named Semi-Synchronous Personalized FederatedAveraging (PerFedS$^2$) that mitigates the straggler problem in PFL.
This is done by optimizing a joint bandwidth allocation and UE scheduling problem.
To solve this problem, we first analyse the convergence rate of PerFedS$^2$ with non-convex loss functions.
Our analysis characterizes the upper bound of the convergence rate in terms of two decision variables: the number of scheduled UEs in each communication round, and the number of communication rounds.
Based on this upper bound, the joint bandwidth allocation and UE scheduling optimization problem can be solved separately.
For the bandwidth allocation problem, we find that for a given UE scheduling policy, there exists infinitely many bandwidth solutions to minimize the overall training time.
For the UE scheduling problem, facilitated by the results obtained from the convergence analysis, the optimal number of UEs that are scheduled to update the global model in each communication round and the optimal number of communication rounds can be estimated.
These results lead us to designing a greedy algorithm that gives the UE scheduling policy.
Finally, with the optimal bandwidth allocation and the UE scheduling policy, we are able to implement PerFedS$^2$ over mobile edge networks.

To summarize, in this paper we make the following contributions:
\begin{itemize}
  \item We propose a new semi-synchronous PFL algorithm, i.e., the PerFedS$^2$, over mobile edge networks. The PerFedS$^2$ strikes a good balance between synchronous and asynchronous PFL algorithms. Particularly, by solving a joint bandwidth allocation and UE scheduling problem, it not only mitigates the straggler problem caused by the synchronous training but also abbreviates potential divergence issue in asynchronous training.
  \item We derive the convergence rate of the PerFedS$^2$. Our analysis characterizes the upper bound of convergence rate as a function with respect to the number of UEs that are scheduled to update the global model in each communication round and the number of communication rounds.
  \item We solve the optimization problem by decoupling it into two sub-problems: bandwidth allocation problem and UE scheduling problem. While the optimal bandwidth is proved to minimize the overall training time within a range of values, the UE scheduling policy can also be determined using a greedy online algorithm.
  \item We conduct extensive experiments by using MNIST, CIFAR-100 and Shakespeare datasets to demonstrate the effectiveness of PerFedS$^2$ in saving the overall training time as well as providing a convergent training loss, compared with four baselines, namely, the synchronous and asynchronous, FL and PFL algorithms, respectively.
\end{itemize}

The rest of the paper has been organized as follows. In Section~\ref{sec:2} we introduce the basic learning process of PerFedS$^2$. Then in Section~\ref{sec:3} we formulate a joint bandwidth allocation and UE scheduling problem to quantify and maximize the benefits PerFedS$^2$ could bring compared with synchronous and asynchronous training. In order to solve the optimization problem, we first analyse the convergence rate of PerFedS$^2$ in Section~\ref{sec:4}. Then, we solve the joint optimization problem in Section~\ref{sec:5}. At last, we evaluate the performance of PerFedS$^2$ in Section~\ref{sec:6}.

\section{Semi-Synchronous Personalized Federated Learning Mechanism} \label{sec:2}

In this section, we propose PerFedS$^2$ to mitigate the drawbacks of synchronous and asynchronous PFL algorithms.
For a better understanding of the proposed algorithm, we commence with reviewing FL and PFL in Section~\ref{sec:2.1} and Section~\ref{sec:2.2}, respectively.
Then, we formally introduce PerFedS$^2$ in Section~\ref{sec:2.3}.

\subsection{Review: Federated Learning} \label{sec:2.1}
Consider a set of $n$ UEs connected to the server via a BS, where each UE has a local data $(x,y)\in\mathcal{X}_i\times\mathcal{Y}_i$.
If we define $f_i:\mathbb{R}^m \rightarrow \mathbb{R}$ as the loss function corresponding to UE $i$, and $w$ as the model parameter that the server needs to learn, then the goal of the server is to solve
\begin{equation}\label{equ:FL}
  \min_{w\in\mathbb{R}^m} f(w):=\frac{1}{n}\sum_{i=1}^{n}f_i(w),
\end{equation}
where $f_i$ represents the expected loss over the data distribution of UE $i$, which is formalized as follows,
\begin{equation}\label{equ:f_i}
  f_i(w):= \mathbb{E}_{(x,y)\sim \mathcal{H}_i}[l_i(w;x,y)],
\end{equation}
where $l_i(w;x,y)$ measure the error of model $w$ in predicting the true label $y$, and $\mathcal{H}_i$ is the distribution over $\mathcal{X}_i\times\mathcal{Y}_i$.

Because the dataset resided on different UEs are usually non-i.i.d. and unbalanced, while the global model trained by FedAvg concentrates on the average performance of all the UEs.
The resultant model may perform very poor on certain individual UEs.
In response, PFL is proposed to capture the statistical heterogeneity among UEs by adapting the global model to local datasets.
We review this scheme in the next subsection.

\subsection{Review: Personalized Federated Learning} \label{sec:2.2}
In contrast to the standard FL, PFL approaches the solution of \eqref{equ:FL} via the Model-Agnostic Meta-Learning (MAML).
Specifically, the target of PFL is to learn an \textit{initial} model that adapts quickly to each UE through one or more gradient steps with only a few data points on the UEs.
Such an initial model is commonly known as the meta model, and the local model after adaptation is referred to as the fine-tuned model.

Formally, if each UE intakes the initial model and updates it via one step of gradient using its own loss function,  problem (\ref{equ:FL}) can be written as
\begin{equation}\label{equ:MAML}
  \min_{w\in\mathbb{R}^m} F(w):=\frac{1}{n}\sum_{i=1}^{n} f_i(w-\alpha\nabla f_i(w)),
\end{equation}
where $\alpha \geq 0$ is the learning rate at individual UEs. Note that we use the same learning rate for all UEs in this paper for simplification. This assumption can be easily extended to the general case when UEs have diverse learning rate $\alpha_i$ as long as $\alpha_i\geq 0$.
For each UE $i$, its optimization objective $F_i$ can be computed as
\begin{equation}\label{equ:F_i}
  F_i(w) := f_i(w-\alpha \nabla f_i(w)).
\end{equation}

Unlike conventional FL, after receiving the current global model, a UE in PFL first adapts the global model to its local data with one step of gradient descent, and then computes local gradients with respect to the model after the adaptation.
This step of local adaptation captures the difference between UEs, and the model learned with this new formulation (\ref{equ:MAML}) is proved to be a good initial point for any UE to start with for fast adaptation~\cite{finn2017model, fallah2020convergence}.

Many existing works on PFL is limited to the context of synchronous learning, where the faster UEs have to wait until all the others arrive the server to move to the next communication round~\cite{fallah2020personalized, dinh2020personalized, deng2020adaptive, shamsian2021personalized, achituve2021personalized}.
As a result, the synchronous PFL often suffers from the \emph{straggler} problem due to the prolonged waiting time for the slowest UE.
On the other hand, the PFL can also be trained in an asynchronous manner, where the server performs global updating as soon as it receives a local model from any UE.
In this scenario, some slower UEs will bring stale gradient updates to the server, thereby degrading the convergence performance of the model training.
Therefore, in this paper, we propose a semi-synchronous PFL mechanism that seeks a trade-off between synchronous and asynchronous PFL algorithms, which is detailed in the following subsection.

\subsection{Semi-Synchronous Personalized Federated Learning} \label{sec:2.3}

\begin{algorithm}[t]
\caption{Semi-Synchronous Personalized Federated Averaging (PerFedS$^2$)}
\label{alg:AutoFLSA}
\For{$k=0,1,\dots,K-1$}{
    \textbf{Processing at Each UE} $i$ \\
    \If{Receive $w_k$ from the server}{
        Compute local gradient $\tilde{\nabla} F_i(w_k)$ by Eq. (\ref{equ:stochastic_local_gradient})
        Upload $\tilde{\nabla} F_i(w_k)$ to the server\\
    }
    \textbf{Processing at the Parameter Server} \\
    $\mathcal{A}_k = \varnothing$ \\
    \While{$|\mathcal{A}_k| < A$}{
        Receive local gradient $\tilde{\nabla} F_i(w_k)$ from UE $i$ \\
        $\mathcal{A}_k = \mathcal{A}_k \cup \{i\}$
    }
    Update global model to $w_{k+1}$ by Eq.~(\ref{equ:stochastic_global_update}) \\
    \For{$i\in\mathcal{U}$}{
        \If{$i\in\mathcal{A}_k$ or $\tau_k^i > S$}{
            Distribute $w_{k+1}$ to UE $i$ \\
        }
    }
}
\end{algorithm}

We propose a semi-synchronous PFL mechanism, which is a trade-off between synchronous and asynchronous PFL.
We term this semi-synchronous PFL algorithm as Semi-Synchronous Personalized FederatedAveraging (PerFedS$^2$).
PerFedS$^2$ is formally described in Alg.~\ref{alg:AutoFLSA}.
At the UE side (Line 2-5), upon receiving a global model, or equivalently, the meta model $w_k$, the UE adapts $w_k$ to its local dataset to obtain the gradient of local functions, which in this case, the gradient $\nabla F_i$, that is given by
\begin{equation}\label{equ:local_update}
  \nabla F_i(w_k) = (I - \alpha \nabla^2 f_i(w_k)) \nabla f_i(w_k - \alpha \nabla f_i(w_k)).
\end{equation}

At the server side (Line 6-12), let $\mathcal{A}_k$ be the set of UEs participating in the global updating in round $k$, with the carnality being $\vert \mathcal{A}_k \vert = A$.
Let $\tau_k^i$ be the interval between the current round $k$ and the last received global model version by UE $i$.
Such an interval reflects the \emph{staleness} of local updates.
With this notion, we can write the gradient received by the BS at round $k$ from UE $i$ as $\nabla F(w_{k-\tau_k^i})$.
Upon receiving $A$ local gradients, the server updates the global model parameter as follows:
\begin{equation}\label{equ:global_update}
  w_{k+1} = w_k - \frac{\beta}{A}\sum_{i\in\mathcal{A}_k}\nabla F_i(w_{k-\tau_k^i}),
\end{equation}
where $\beta>0$ is the global step size.
Then, the server distributes the new global model $w_{k+1}$ to either ($a$) the UEs in $\mathcal{A}_{k}$ or ($b$) those with a staleness larger than the staleness threshold $S$.

Due to the vast volume of dataset, computing the exact gradient for each UE is costly.
Therefore, we use the stochastic gradient descent (SGD)~\cite{bottou2012stochastic} as a proxy.
Specifically, a generic UE $i$ samples a subset of data points to calculate an unbiased estimate $\tilde{\nabla}f_i(w_k;\mathcal{D}_i)$ of $\nabla f_i(w_k)$, where $\mathcal{D}_i$ represents a portion of UE $i$'s local dataset with size $|\mathcal{D}_i| = D_i$.
Similarly, the Hessian $\nabla^2$ in (\ref{equ:local_update}) can be replaced by its unbiased estimate $\tilde{\nabla}^2 f_i(w_k;\mathcal{D}_i)$.
At this point, the actual gradient computed by UE $i$ is the stochastic gradient of local loss function $\tilde{\nabla} F_i(w_k)$, which is given by
\begin{align}\label{equ:stochastic_local_gradient}
  & \tilde{\nabla}  F_i(w_k) = \nonumber \\
  & (I-\alpha \tilde{\nabla}^2 f_i(w_k;\mathcal{D}_i^{\text{h}})) \tilde{\nabla}f_i(w_k - \alpha \tilde{\nabla} f_i(w_k;\mathcal{D}_i^{\text{in}});\mathcal{D}_i^{\text{o}}),
\end{align}
where $\mathcal{D}_i^{\text{in}}$, $\mathcal{D}_i^{\text{o}}$ and $;\mathcal{D}_i^{\text{h}}$ are independently sampled datasets with total size denoted by $d_i = D_i^{\text{in}} + D_i^{\text{o}} + D_i^{\text{h}}$.
This stochastic gradient is then uploaded to the central server for global model update as follows: \begin{equation}\label{equ:stochastic_global_update}
  w_{k+1} = w_k - \frac{\beta}{A} \sum_{i\in\mathcal{A}_k} \tilde{\nabla} F_i(w_{k-\tau_k^i})
\end{equation}

\section{System Model and Problem Formulation}\label{sec:3}

In the last section, we introduce the basic learning process of PerFedS$^2$. This alone is not enough to quantify the benefits a semi-synchronous training manner brings to implementation, because the communication related parameters and the training hyperparameters remain to be unclear. Therefore, our next step is to formulate an optimization problem for PerFedS$^2$, with the wireless bandwidth allocation and the UE scheduling policy to be determined. In this section, We introduce some notations and concepts in Section~\ref{sec:3.1} and~\ref{sec:3.2} that are used to formulate the optimization problem in Section~\ref{sec:3.3}.

\subsection{Communication Model} \label{sec:3.1}

To implement PerFedS$^2$ in mobile edge networks, the wireless communication environments should also be considered to maximize the benefit a semi-asynchronous learning manner brings to the learning algorithm.
Note that in PerFedS$^2$, one local iteration of UE $i$ may last for a few global communication rounds, we focus on describing the wireless communication processes of UE $i$ within such a local iteration.
The learning time of UE $i$ during one local iteration consists of two parts: communication time and computation time.
As for the communication time over mobile edge networks, we consider that UEs access the BS through a channel partitioning scheme, such as orthogonal frequency division multiple access (OFDMA)~\cite{yin2006ofdma}, with total bandwidth $B$.
Meanwhile, the bandwidth allocation to UE $i$ in round $k$ is denoted as $b_k^i$.
The uplink rate of UE $i$ transmitting its local gradients to the BS can be computed as follows~\cite{shi2020joint, chen2020joint},
\begin{equation}\label{equ:uplink_rate}
  r_k^i = b_k^i \ln(1+\frac{p_i h_k^i \|c_i\|^{-\kappa}}{b_k^i N_0}),
\end{equation}
where $p_i$ is the transmit power of UE $i$, $\kappa$ is the path loss exponent, and $N_0$ is the noise power spectral density. $h_k^i \|c_i\|^{-\kappa}$ is the channel gain between UE $i$ and the BS at round $k$ with $c_i$ being the distance between UE $i$ and the BS and $h_k^i$ being the small-scale channel coefficient. In this paper, we assume that the small-scale channel coefficients across communication rounds $h_k^i$ follow Rayleigh distribution~\cite{sklar1997rayleigh}. With $r_k^i$, the uplink transmission delay of UE $i$ can be specified as follows,
\begin{equation}\label{equ:com_delay}
  {Tcom}_k^i = \frac{Z_k^i}{r_k^i},
\end{equation}
where $Z_k^i$ denotes the number of bits UE $i$ transmits in round $k$.
Meanwhile, $Z$ denotes total size of the gradient UE $i$ transmits each time.
Since the transmit power of the BS is much higher than the UEs', the downlink transmission latency is much smaller than that in the uplink.
Meanwhile, we care more about the transmit power allocation on individual UEs rather than that on the server, hence we ignore the downlink delay for simplicity.

As for the computation time, let $c_i$ denote the number of CPU cycles for UE $i$ to execute one sample of data, $\vartheta_i$ denote the CPU-cycle frequency of UE $i$, and $d_i$ denote the number of sampled data points on UE $i$, then the computation time of UE $i$ per local iteration can be expressed as follows~\cite{shi2020joint},
\begin{equation}\label{equ:cmp_delay}
  {Tcmp}^i_k = \frac{c_i d_i}{\vartheta_i}.
\end{equation}
As such, given that for semi-synchronous training, each local iteration of UE $i$ may last several global rounds, the total time UE $i$ spent in round $k$ is given by
\begin{equation}\label{equ:round_delay}
T_k^i = \left\{
    \begin{aligned}
        & {Tcom}_k^i + {Tcmp}^i_k, \\
        & \text{          when UE } i \text{ starts a new local iteration in round } k, \\
        & {Tcom}_k^i, \quad \text{otherwise.}
    \end{aligned}
\right.
\end{equation}


\begin{figure*}
  \centering
  \includegraphics[width=\linewidth]{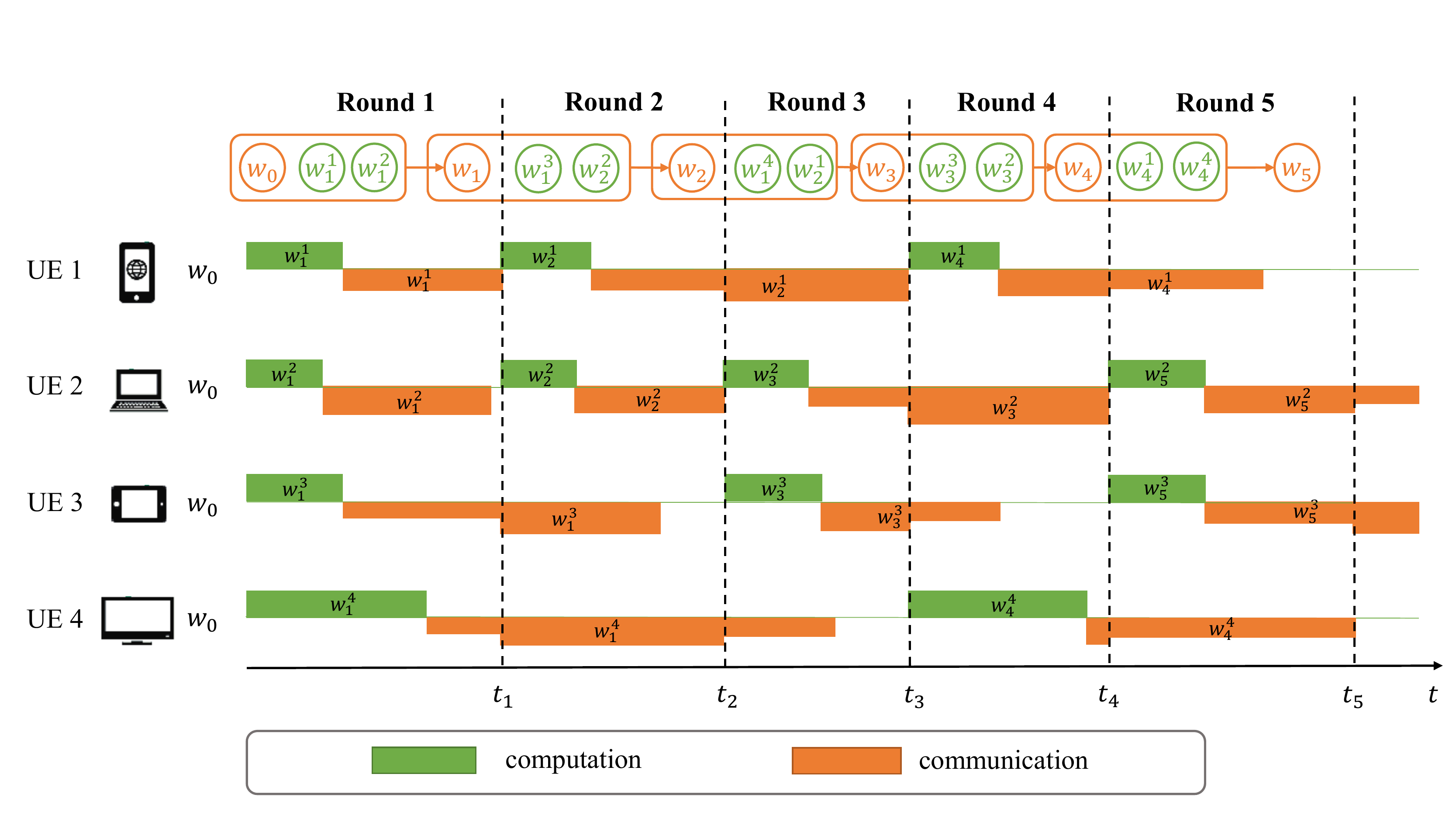}
  \caption{Example of the PerFedS$^2$ mechanism when $A=2$.}
  \label{fig:example}
\end{figure*}

\subsection{Illustrative Example} \label{sec:3.2}

We give an example to facilitate the understanding of PerFedS$^2$.
Consider the scenario depicted in Fig.~\ref{fig:example}, where $A=2$.
This network has four UEs.
In the first communication round, UE $3$ and $4$ are stragglers. Therefore, once the stochastic gradients uploaded by UE $1$ and $2$ arrive at the server in round $1$, the server updates the global model from $w_0$ to $w_1$, leaving the gradients computed by UE $3$ and $4$ to be integrated into the global model in round $2$ and round $3$, respectively.

\vspace{0.2cm}
\noindent\textbf{Scheduling policy}: Let $\pi_k^i \in \{0, 1\}$ be an indicator to denote whether the gradient uploaded from UE $i$ arrives at the server in round $k$.
That is, $\pi_k^i=1$ if the update from UE $i$ is included in the global model in round $k$, and $\pi_k^i=0$ otherwise. Then, $\mathbf{\Pi} \triangleq [\mathbf{\Pi}_1,\mathbf{\Pi}_2,\dots,\mathbf{\Pi}_K]$ denotes the scheduling decision matrix up to round $K$, where $\mathbf{\Pi}_k \triangleq [\pi_k^1,\pi_k^2,\dots,\pi_k^n]$.
For the example given in Fig.~\ref{fig:example}, the computation has been carried out five rounds and the scheduling decision matrix $\mathbf{\Pi}$ can be written as
\begin{equation}\label{matrix:pi}
\mathbf{\Pi}=
\begin{pmatrix}
  1 & 1 & 0 & 0 \\
  0 & 1 & 1 & 0 \\
  1 & 0 & 0 & 1 \\
  0 & 1 & 1 & 0 \\
  1 & 0 & 0 & 1 \\
\end{pmatrix}.
\end{equation}
From the above, we can see that the entries in each row of $\mathbf{\Pi}$ satisfy the following relationship
\begin{equation}\label{equ:A}
  \sum_{i=1}^{n} \pi_k^i = A.
\end{equation}

We further introduce a concept, coined as the relative participation frequency, to reflect the statistical property of the scheduling policy.
Specifically, for UE $i$, we denote its relative participation frequency as $\eta_i$, which represents the fraction of time this UE participates in the global iteration.
Such a notion is formally defined as
\begin{equation}\label{equ:eta}
\eta_i = \frac{\sum_{k=1}^{K} \pi_k^i}{\sum_{k=1}^{K}\sum_{i=1}^{n} \pi_k^i} = \frac{\sum_{k=1}^{K} \pi_k^i}{AK}.
\end{equation}

Notably, the staleness bound $S$ provides a lower bound of $\eta_i$, that is, $\eta_i \geq S/K$ ($\forall i\in \mathcal{U}$).

\subsection{Problem Formulation} \label{sec:3.3}

PerFedS$^2$ significantly increases the proportion of time UEs spend on computing, as opposed to waiting.
Meanwhile, PerFedS$^2$ also upper bounds the staleness caused by updates from slow UEs.
Let $T$ be the overall training time over $K$ communication rounds. Then the objective of PerFedS$^2$ is to minimize the loss function as well as the overall training time.
Formally, the optimization problem of PerFedS$^2$ is formulated as follows~\footnote{Besides bandwidth allocation and UE scheduling policy, other decision variables such like transmit power can also be included in the problem formulation. The logic keeps the same, but the parameters that need to be considered might change. Problem (P1) shows the case when we consider the bandwidth allocation and UE scheduling policy as variables, and it is free for the researcher to extend this general formulation to other forms.},

\begin{subequations}\label{equ:opt1}
\begin{align}
  \min_{\mathbf{b},\Pi, A, K} \quad & F(w) \label{equ:opt1_obj} \tag{P1}\\
  \text{s.t.} \quad & \min_{\mathbf{b}} \sum_{k=1}^{K}\max_{i\in\mathcal{A}_k} \{T_k^i\} = T, \quad \forall i\in \mathcal{U}, \label{equ:opt1_const1} \tag{C1.1}\\
  & \sum_{i=1}^{n} b_k^i \leq B, \quad k=1,2,\dots, K, \label{equ:opt1_const2}\tag{C1.2}\\
  & \sum_{j=k-\tau_k^i}^{k-\tau_k^i+S} \pi_j^i \geq 1, \quad \forall i\in \mathcal{U} \label{equ:opt1_const3} \tag{C1.3} \\
  & \sum_{j=k-\tau_k^i}^{k} Z_j^i \leq Z \label{equ:opt1_const4} \tag{C1.4} \\
  & K \geq \frac{S}{\eta_i}, \quad \forall i\in \mathcal{U}, \label{equ:opt1_const5} \tag{C1.5}
\end{align}
\end{subequations}
where $\mathbf{b}\triangleq [\mathbf{b}_1,\mathbf{b}_2,\dots,\mathbf{b}_K]$ denotes the bandwidth allocation matrix up to round $K$, and $\mathbf{b}_k = [b_k^1, b_k^2,\dots, b_k^n]$. (\ref{equ:opt1_const1}) is the overall training time constraint, that for each communication round $k$, the round time is determined by the maximum of $T_k^i$ over $i
\in\mathcal{A}_k$, and the total time up to round $K$ is equal to $T$.
(\ref{equ:opt1_const2}) is the bandwidth constraint, that the bandwidth allocation to all UEs in every communication round shall not exceed the available bandwidth $B$.
(\ref{equ:opt1_const3}) stipulates the staleness constraint on the updates, that the during any $S$ rounds of communication, UE $i$ must be scheduled to update the global model at least once.
(\ref{equ:opt1_const4}) limits the number of bit transmitted, note that $Z_k^i$ is determined by $b_k^i$, and the number of bits that are transmitted during $\tau_k^i$ rounds shall not be larger than the size of model parameters.
Finally, (\ref{equ:opt1_const5}) follows from the lower bound we drawn in the previous subsection.

\section{Convergence Analysis} \label{sec:4}

In this section, we first introduce some definitions and assumptions on the loss functions of PerFedS$^2$ in Section~\ref{sec:4.1}. Then we analyse its convergence rate in Section~\ref{sec:4.2}.

\subsection{Preliminaries} \label{sec:4.1}
We consider the \textit{non-convex} loss functions in this paper.
Our goal is to find an $\epsilon$-approximate first-order stationary point (FOSP) for PerFedS$^2$ \cite{fallah2020convergence, fallah2020personalized}.
The formal definition of FOSP is given as follows.
\begin{defn}
  A random vector $w_{\epsilon}\in\mathbb{R}^m$ is called an $\epsilon$-FOSP for PerFedS$^2$ if it satisfies $\mathbb{E}[\|\nabla F(w_{\epsilon})\|^2] \leq \epsilon$.
\end{defn}

To make the convergence analysis consistent with that of Per-FedAvg, we make the following assumptions~\cite{fallah2020personalized}.

\begin{assum}[Bounded Staleness] \label{assum:1}
  All delay variables $\tau_k^i$'s are bounded, i.e., $\max_{k,i} \tau_k^i \leq S$.
\end{assum}
\begin{assum} \label{assum:2}
    For each UE $i\in\mathcal{U}$, its gradient $\nabla f_i$ is $L$-Lipschitz continuous and is bounded by a nonnegative constant $C$, namely,
    \begin{align}\label{equ:assum_2}
    \|\nabla f_i(w)-\nabla f_i(u)\|  & \leq L\|w-u\|, \qquad w,u\in \mathbb{R}^m \\
    \|\nabla f_i(w)\| & \leq C, \qquad w\in\mathbb{R}^m.
    \end{align}
\end{assum}
\begin{assum} \label{assum:3}
  For each UE $i\in\mathcal{U}$, the Hessian of $f_i$ is $\rho$-Lipschitz continuous:
  \begin{equation}\label{equ:assum_3}
    \|\nabla^2 f_i(w) - \nabla^2 f_i(u)\|\leq \rho\|w-u\|, \qquad w,u\in\mathbb{R}^m.
  \end{equation}
\end{assum}
\begin{assum}\label{assum:4}
For any $w\in \mathbb{R}^m$, $\nabla l_i(w;x,y)$ and $\nabla^2 l_i(w;x,y)$, computed \textit{w.r.t.} a single data point $(x,y)\in \mathcal{X}_i\times\mathcal{Y}_i$, have bounded variance:
\begin{align}\label{equ:assum_4}
  \mathbb{E}_{(x,y)\thicksim p_i}[\|\nabla l_i(w;x,y)-\nabla f_i(w)\|^2] & \leq \sigma^2_G, \nonumber \\
  \mathbb{E}_{(x,y)\thicksim p_i}[\|\nabla^2l_i(w;x,y)-\nabla^2 f_i(w)\|^2]& \leq \sigma^2_H.
\end{align}
\end{assum}
\begin{assum}\label{assum:5}
  For any $w\in \mathbb{R}^m$, the gradient and Hessian of local loss function $f_i(w)$ and the average loss function $f(w) = 1/n \sum_{i=1}^{n}f_i(w)$ satisfy the following conditions:
  \begin{align}\label{equ:assum_5}
    \frac{1}{n}\sum_{i=1}^{n}\|\nabla f_i(w) -\nabla f(w)\|^2 & \leq \gamma_G^2, \nonumber \\
    \frac{1}{n}\sum_{i=1}^{n}\|\nabla^2 f_i(w) - \nabla^2 f(w)\|^2 & \leq \gamma_H^2.
  \end{align}
\end{assum}

While Assumption~\ref{assum:1} limits the maximum of the staleness, Assumptions~\ref{assum:2} to~\ref{assum:5} characterize the properties of the gradient and Hessian of $f_i(w)$, which are necessary to deduce the following lemmas and convergence rate analysis.

\subsection{Analysis of Convergence Bound} \label{sec:4.2}

Before delving into the full details of convergence analysis, we introduce three lemmas inherited from~\cite{fallah2020personalized} to quantify the smoothness of $F_i(w)$ and $F(w)$, the deviation between $\nabla F_i(w)$ and its estimate $\tilde{\nabla} F_i(w)$, and the deviation between $\nabla F_i(w)$ and $\nabla F(w)$, respectively.
\begin{lem}\label{lem:1}
  If Assumptions 2-4 hold, then $F_i$ is smooth with parameter $L_F:=4L+\alpha \rho C$. As a consequence, the average function $F(w) = 1/n \sum_{i=1}^n F_i(w)$ is also smooth with parameter $L_F$.
\end{lem}
\begin{lem}\label{lem:2}
  If Assumptions 2-4 hold, then for any $\alpha_i\in(0,1/L]$ and $w\in\mathbb{R}^m$, we have
  \begin{align}\label{equ:lem2}
    \left\|\mathbb{E}\left[ \tilde{\nabla}F_i(w) - \nabla F_i(w) \right]\right\| & \leq \frac{2\alpha L \sigma_G}{\sqrt{D^{\text{in}}}}, \\
      \mathbb{E}\left[\|\tilde{\nabla}F_i(w) - \nabla F_i(w)\|^2\right] & \leq \sigma_F^2.
  \end{align}
  where $\sigma_F^2$ is defined as
  \begin{equation} \label{equ:sigma_F}
    \sigma_F^2:= 12\left[C^2 + \sigma_G^2\left[ \frac{1}{D^{\text{o}}} + \frac{(\alpha L)^2}{D^{\text{in}}}\right]\right]\left[1+\sigma_H^2 \frac{\alpha^2}{4D^{\text{h}}}\right] - 12C^2,
  \end{equation}
  where $D^{\text{in}} = \max_{i\in \mathcal{U}} D_i^{\text{in}}$, $D^{\text{o}} = \max_{i\in \mathcal{U}} D_i^{\text{o}}$ and $D^{\text{h}} = \max_{i\in \mathcal{U}} D_i^{\text{h}}$.
\end{lem}
\begin{lem} \label{lem:3}
  Given the loss function $F_i(w)$ shown in (\ref{equ:F_i}) and  $\alpha \in (0,1/L]$, if the conditions in Assumptions \ref{assum:2},~\ref{assum:3}, and \ref{assum:5} are all satisfied, then for any $w\in \mathbb{R}^m$, we have
  \begin{equation}\label{equ:lem3}
    \frac{1}{n} \sum_{i=1}^{n}\|\nabla F_i(w) - \nabla F(w)\|^2 \leq \gamma_F^2,
  \end{equation}
  where $\gamma_F^2$ is defined as
  \begin{equation}\label{equ:gamma_F}
    \gamma_F^2:=3C^2\alpha^2\gamma_H^2 +192\gamma_G^2,
  \end{equation}
  where $\nabla F(w) = 1/n\sum_{i=1}^{n} \nabla F_i(w)$.
\end{lem}

Based on the three lemmas, we obtain the following theorem to

\begin{thm} \label{thm:1}
  If Assumptions~\ref{assum:1} to~\ref{assum:5} hold and the steplength $L_F$ in Lemma~\ref{lem:1} satisfies
  \begin{equation}\label{equ:thm_constraint}
    L_F\beta^2-\beta+2L_F^2\beta^2S^2 \leq 1,
  \end{equation}
  then the following FOSP condition holds,
  \begin{align}\label{equ:convergence_rate}
    & \frac{1}{K} \sum_{k=0}^{K-1} \mathbb{E}[\|\nabla F(w_k)\|^2] \leq \frac{2(F(w_0) - F(w^*))}{\beta K}\nonumber \\
    & + 4(L_F\beta +2 L_F^2\beta^2S^2)(\sigma_F^2+\gamma_F^2)\sqrt{A}.
  \end{align}
\end{thm}
\begin{IEEEproof}
See the Appendix.
\end{IEEEproof}

\begin{cor} \label{cor:1}
  Assume the conditions in Theorem~\ref{thm:1} are satisfied. Then, if we set the number of total communication rounds as $K=\mathcal{O}(\epsilon^{-3})$, the global learning rate as $\beta=\mathcal{O}(\epsilon^2)$, the staleness threshold as $S=\mathcal{O}(\epsilon^{-1})$, and the number of UEs that updates the global model as $A = \mathcal{O}(\epsilon^{-2})$, Algorithm 1 finds an $\epsilon$-FOSP for PerFedS$^2$.
\end{cor}

\begin{IEEEproof}
  Note that $2(F(w_0)-F(w^*))$ is constant, then $K=\mathcal{O}(\epsilon^{-3})$ and $\beta=\mathcal{O}(\epsilon^2)$ ensure the first term of right-hand-side of (\ref{equ:convergence_rate}) to be equal to $\mathcal{O}(\epsilon)$. Next we examine the second term of (\ref{equ:convergence_rate}). Note that $(\sigma_F^2+\gamma_F^2)$ is constant, then $\beta = \mathcal{O}(\epsilon^2)$ and $S=\mathcal{O}(\epsilon^{-1})$ together make $(2L_F\beta + 4 L_F^2\beta^2S^2) = \mathcal{O}(\epsilon^2)$. At this point, if $A= \mathcal{O}(\epsilon^{-2})$, the second term of (\ref{equ:convergence_rate}) is equivalent to $\mathcal{O}(\epsilon)$.
\end{IEEEproof}

\section{Joint Bandwidth Allocation and UE Scheduling} \label{sec:5}

In this section, we present the steps to solve the optimization problem P1.
Particularly, we decouple P1 into P2, a bandwidth allocation problem, and P3, a UE scheduling problem.
Note that individually solving the two sub-problems is equivalent to solving the original P1, which will be elaborated in the sequel.

\subsection{Problem Decoupling} \label{sec:5.1}
We begin with the bandwidth allocation problem. Given a scheduling pattern $\mathbf{\Pi}$, the bandwidth allocation problem can be written as follows:
\begin{subequations}\label{equ:opt2}
\begin{align}
  \min_{\mathbf{b}} \quad & T(\mathbf{\Pi}) \label{equ:opt2_obj} \tag{P2} \\
  s.t. \quad & \sum_{k=1}^{K} \max_{i\in\mathcal{A}_k} \{T_k^i\} \leq T(\mathbf{\Pi})\label{equ:opt2_const1} \tag{C2.1} \\
  & \sum_{i=1}^n b_k^i \leq B, k=1,2,\dots, K \label{equ:opt2_const2} \tag{C2.2} \\
  & \sum_{j=k-\tau_k^i}^{k} Z_j^i \leq Z , \quad \forall i \in \mathcal{U}. \label{equ:opt2_const3} \tag{C2.3}
\end{align}
\end{subequations}
Then, with the optimal bandwidth allocation and the corresponding minimal overall training time $T^*(\mathbf{\Pi})$, the UE scheduling problem can be written as follows,
\begin{subequations}\label{equ:opt3}
\begin{align}
  \min_{K, A, \mathbf{\Pi}} \quad & F(w) \label{equ:opt3_obj} \tag{P3} \\
  s.t. \quad & \sum_{k=1}^{K} \max_{i\in\mathcal{A}_k} \{T_k^i\} = T^*(\mathbf{\Pi}), \quad \forall i \in \mathcal{U} \label{equ:opt3_const1} \tag{C3.1}\\
  & \sum_{j=k-\tau_k^i}^{k-\tau_k^i+S} \pi_j^i \geq 1,  \quad \forall i\in\mathcal{U} \label{equ:opt3_const2} \tag{C3.2} \\
  & K \geq \frac{S}{\eta_i},  \quad \forall i\in\mathcal{U} \label{equ:opt3_const3} \tag{C3.3}.
\end{align}
\end{subequations}

\subsection{Optimal Bandwidth Allocation} \label{sec:5.2}
In order to solve P2, we introduce the following theorems to explore the relationship between $b_k^i$ and $T(\mathbf{\Pi})$ step by step.

\begin{thm} \label{thm:2}
  If the server updates the global model after receiving $A$ gradients from the UEs in each round, then the optimal bandwidth allocation can be achieved if and only if all the scheduled UEs have the same finishing time.
\end{thm}

\begin{IEEEproof}
Recall the expression of $r_k^i$ defined in (\ref{equ:uplink_rate}),
we take a derivative with respect to $b^i_k$ and arrive at the following
\begin{align}
  & \frac{\text{d}}{\text{d} b_k^i}\left( b_k^i \ln \left(1+\frac{p_ih_i\|c_i\|^{-\kappa}}{b_k^i N_0}\right)\right) \nonumber \\
  = & \ln \left(1+\frac{p_ih_i\|c_i\|^{-\kappa}}{b_k^i N_0}\right) - \frac{p_ih_i\|c_i\|^{-\kappa}}{b_k^i N_0 + p_ih_i\|c_i\|^{-\kappa}} \\
  > & \frac{\frac{p_ih_i\|c_i\|^{-\kappa}}{b_k^i N_0}}{1+\frac{p_ih_i\|c_i\|^{-\kappa}}{b_k^i N_0}} - \frac{p_ih_i\|c_i\|^{-\kappa}}{b_k^i N_0 + p_ih_i\|c_i\|^{-\kappa}}\nonumber \\
  = & 0,
\end{align}
where the inequality follows from the fact that $\ln(1+x) > \frac{x}{1+x}$, for $x>0$. Therefore, $r_k^i$ monotonically increases with $b_k^i$. While it is obvious that $r_k^i > 0$, and thus ${Tcmp}_k^i + {Tcom}_k^i = {Tcmp}_k^i + \frac{Z_k^i}{r_k^i}$ monotonically decreases with $b_k^i$.
Therefore, at round $k$, if any UE $i\in\mathcal{A}_k$ has finished its whole local model update process than the others, we can decrease its bandwidth allocation to make it up for the other slower UEs in $\mathcal{A}_k$.
As a result, the round latency which is determined by the slowest UE in $\mathcal{A}_k$ can be reduced.
Such a bandwidth compensation is performed until all scheduled UEs in $\mathcal{A}_k$ finish their local iterations at the same time.
Consequently, the optimal bandwidth allocation in round $k$ is achieved when all scheduled UEs in $\mathcal{A}_k$ have the same finishing time.
\end{IEEEproof}

\begin{thm} \label{thm:3}
  Given the relative participation frequency $\eta_i$ ($i\in\mathcal{U}$), the UEs would be scheduled in an order with a recurrence pattern. That is, the UEs would periodically participate into the global model update.
\end{thm}

\begin{IEEEproof}
  Recall the formulation of $\eta_i$ defined in (\ref{equ:eta}), it is obvious that $\eta_i$ is computed by the number of times UE $i$ has been scheduled during all $K$ rounds.
  Therefore, if $\eta_i$ is settled, then $\sum_{k=0}^{K-1} \pi_k^i$ is settled.
  As a result, if the UEs are scheduled periodically, the times of each UE involved in the global update can be settled, thus matching the relative participation rate it has been assigned with.
\end{IEEEproof}

\begin{thm} \label{thm:4}
  The optimal bandwidth allocation that achieves the minimum learning time is given by the following
  \begin{equation} \label{equ:thm4}
  \left\{
  \begin{aligned}
    & \sum_{i\in\mathcal{U}} b_k^i = B, \quad k=1,\dots,K \\
    & b_k^i > \frac{Bn\eta_iZ}{(T_i^*(\Pi) - {Tcmp}_i)(W(-\Gamma_{i} e^{-\Gamma_i}) + \Gamma_i)}, \\
    & \sum_{i\in\mathcal{A}_k} b_k^i \leq B,
  \end{aligned}
  \right.
  \end{equation}
  where $\Gamma_i \triangleq \frac{N_0 Z}{(T_i^*(\Pi) - {Tcmp}_i)p_ih_i\|c_i\|^{-\kappa}}$, $W(\cdot)$ is Lambert-W function, and $T_i^*(\Pi)$ is the objective value of (P2).
\end{thm}

\begin{IEEEproof}
    From Theorem~\ref{thm:3}, we know that all UEs update the global model periodically. Let $K_p$ denote the number of communication rounds in each period, then inferring from Theorem~\ref{thm:2}, all UEs have the same finishing time in each period without any waiting time. That is, we have
    \begin{equation}\label{equ:thm4_1}
      \sum_{k=1}^{K_p} T_k^i = \sum_{k=1}^{K_p} T_k^i,\quad \forall i,j \in\mathcal{U}, i\neq j,
    \end{equation}
    Meanwhile, we have
    \begin{equation}\label{equ:thm4_2}
      \sum_{k=1}^{K_p} Z_k^i = \eta_i ZAK_P, \quad \forall i\in\mathcal{U},
    \end{equation}
    where $ZAK_P$ denotes the number of bits that needs to be transmitted during the $K_p$ rounds. This equation indicates that the number of bits transmitted by UE $i$ during $K_p$ rounds is equal to the product of its relative participation frequency $\eta_i$ and the total number of bits transmitted during that $K_p$ communication rounds. From equation (\ref{equ:thm4_2}), it is easy to indicate that
    \begin{equation}\label{equ:thm4_3}
      \sum_{k=1}^{K_p} \frac{Z_k^i}{\eta_i} = \sum_{k=1}^{K_p} \frac{Z_k^j}{\eta_j}, \quad \forall i,j \in\mathcal{U}, i\neq j.
    \end{equation}
    Now combing (\ref{equ:thm4_1}) and (\ref{equ:thm4_3}), we have
    \begin{equation}\label{equ:thm4_4}
      \frac{\sum_{k=1}^{K_p} Z_k^i}{\eta_i\sum_{k=1}^{K_p} T_k^i} =  \frac{\sum_{k=1}^{K_p} Z_k^j}{\eta_j\sum_{k=1}^{K_p} T_k^j}
      ,\quad \forall i,j \in\mathcal{U}, i\neq j.
    \end{equation}
    From equation (\ref{equ:thm4_4}) we observe that $\frac{\sum_{k=1}^{K_p} Z_k^i}{\sum_{k=1}^{K_p} T_k^i}$ denotes the average rate of UE $i$ during $K_p$ rounds. That is, we have
    \begin{equation}\label{equ:thm4_5}
      \frac{\mathbb{E}(r_k^i)}{\eta_i} = \frac{\mathbb{E}(r_k^j)}{\eta_j},\quad \forall i,j \in\mathcal{U}, i\neq j.
    \end{equation}
    The above equation states a fact that as long as the average rate of each UE is weighted equalized, the optimal solution is achieved. Therefore, there exists \emph{infinitely many solutions} of $r_k^i$ to the above equation.
    The simplest solution is $\frac{\eta_i}{r_k^i} = \frac{\eta_j}{r_k^j}$ in each round $k$.
    Note that $r_k^i$ is determined by $b_k^i$, and thus there exits infinitely many solutions of $b_k^i$ in each round $k$.

    Our next step is to compute the boundary values of $b_k^i$. To do this, we first divide UEs into two categories: UEs in $\mathcal{A}_k$ and UEs do not in $\mathcal{A}_k$.
    \begin{itemize}
      \item At one extreme case, only UEs in $\mathcal{A}_k$ are assigned with bandwidth. That is, $\sum_{i\in\mathcal{A}_k} b_k^i= B$. Under this case, the PerFedS$^2$ algorithm turns out to be a synchronous PerFedAvg algorithm where in each round $A$ UEs are selected to update the global model. Meanwhile, the bandwidth is allocated proportionally to the UEs in $\mathcal{A}_k$ such that $\frac{r_k^i}{\eta_i} = \frac{r_k^j}{\eta_j}$, $\forall i,j\in\mathcal{A}_k, i \neq j$. This extreme case is corresponding to the third inequation of (\ref{equ:thm4}).
      \item At the other extreme case, all UEs in round $k$ share the available bandwith $B$ at a rate $\frac{r_k^i}{\eta_i} = \frac{r_k^j}{\eta_j}$, $\forall i,j\in\mathcal{A}_k, i \neq j$. This case indicates the least bandwidth allocation to UEs in $\mathcal{A}_k$ to ensure their orders to arrive the server in the scheduling pattern. Under this case, $\sum_{i\in\mathcal{U}} b_k^i = B$. Therefore, a closed form of $b_k^i$ is obtained, which is corresponding to the lower bound of $b_k^i$ shown in the second inequation of (\ref{equ:thm4}).
    \end{itemize}
\end{IEEEproof}

\begin{figure*}[!t]
  \centering
  \subfloat[The largest bandwidth allocation to UEs in $\mathcal{A}_k$]{
      \includegraphics[width=3in]{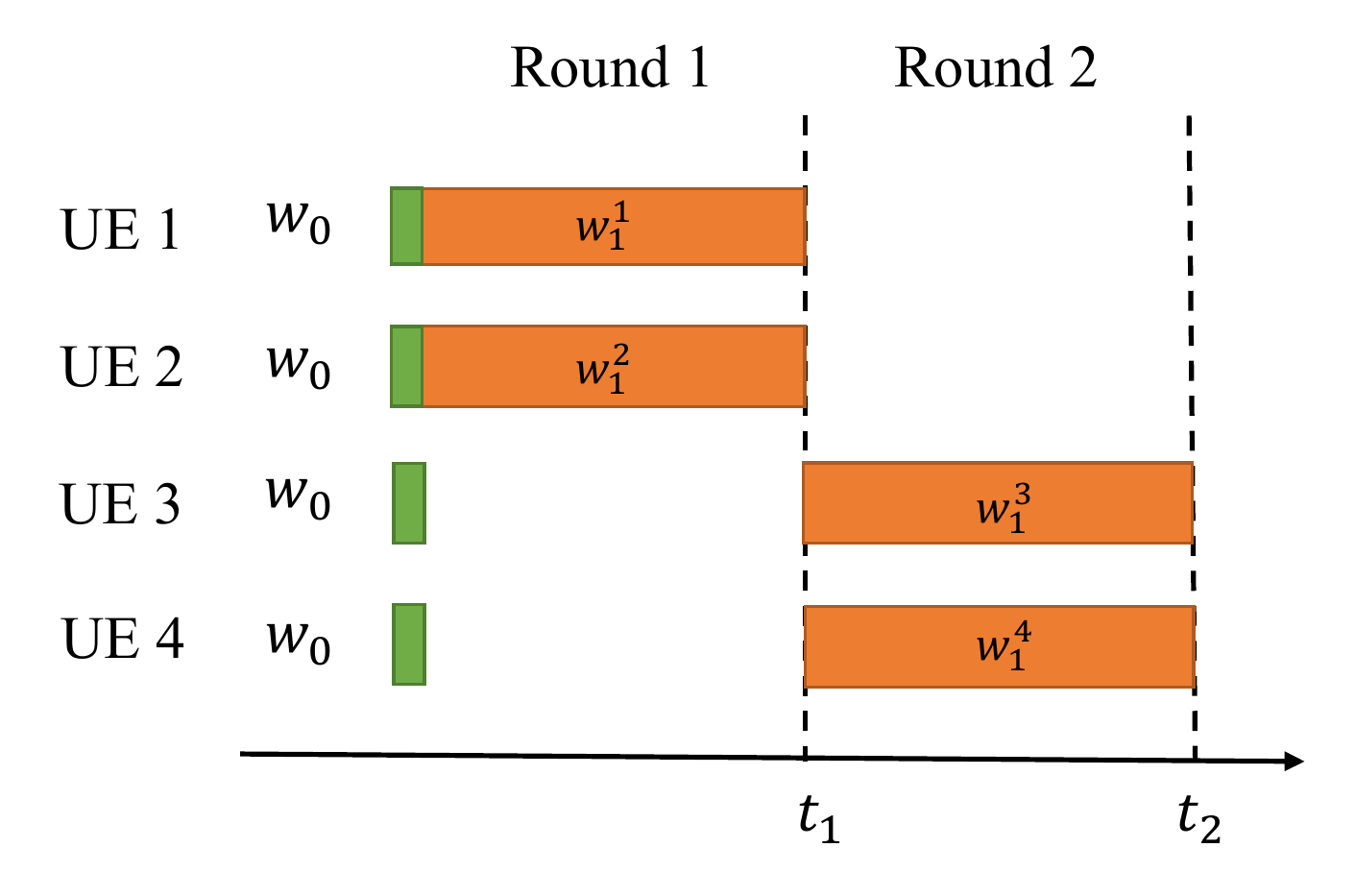}
      \label{fig_bandwidth_allocation:subfig:a}}
  \subfloat[The least bandwidth allocation to UEs in $\mathcal{A}_k$]{
      \includegraphics[width=3in]{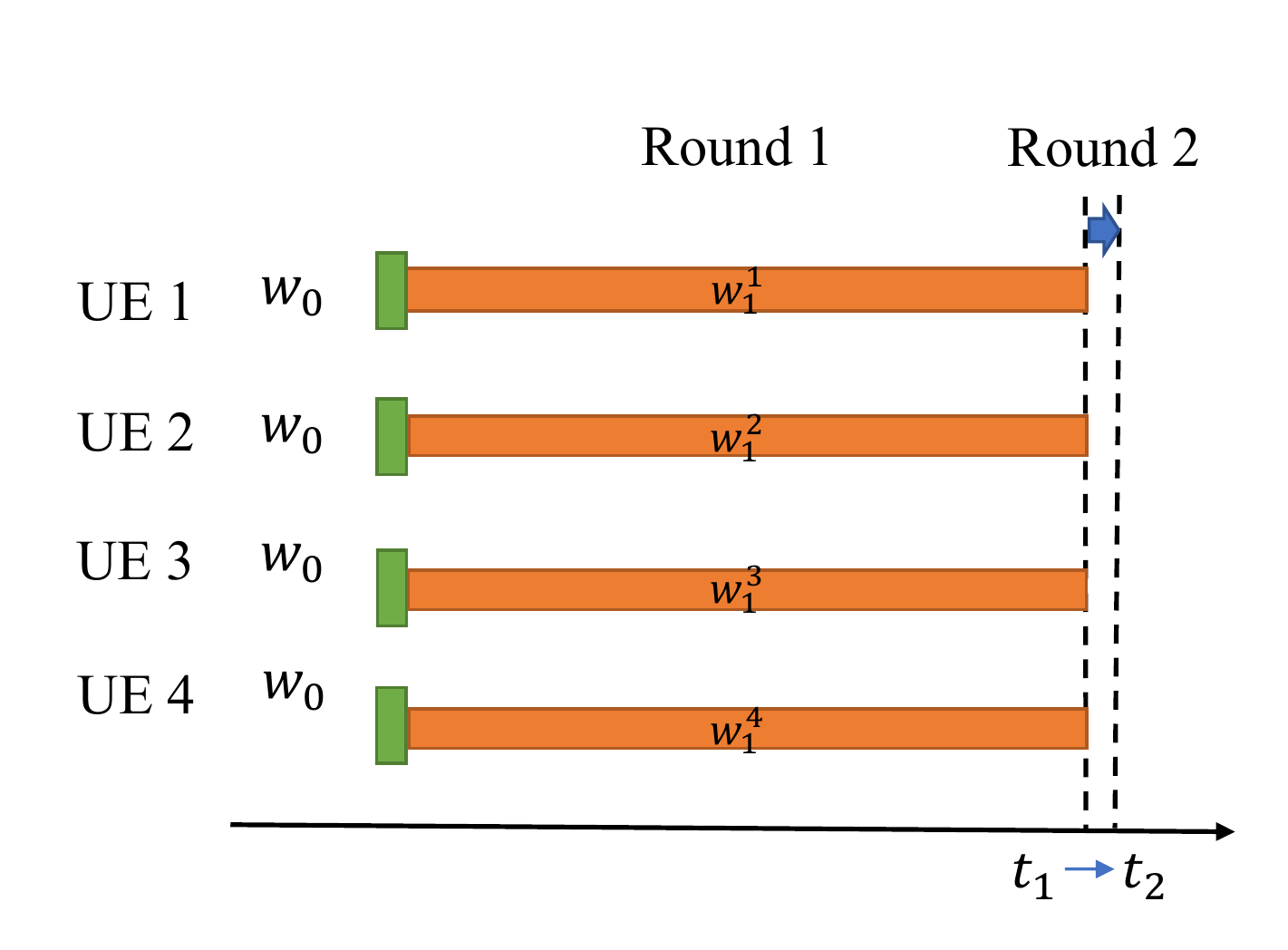}
      \label{fig_bandwidth_allocation:subfig:b}}
  \caption{Bandwidth allocation example, where all UEs have the same parameters, and $A=2$.}
  \label{fig:bandwidth_allocation}
\end{figure*}

To better illustrate these approaches, let us take the example in Fig.~\ref{fig:bandwidth_allocation}.
Assume $A=2$ and the four UEs have the same $\eta_i$, $p_i$, $h_i$, and $c_i$.
We can write the scheduling pattern $\Pi$ of the four UEs as follows:
\begin{equation}\label{matrix:example}
\begin{pmatrix}
  1 & 1 & 0 & 0 \\
  0 & 0 & 1 & 1 \\
  1 & 1 & 0 & 0 \\
  0 & 0 & 1 & 1 \\
  \hdotsfor{4}
\end{pmatrix}.
\end{equation}
The length of the scheduling period is $K_p=2$. Meanwhile, according to Theorem~\ref{thm:4}, we have $\mathbb{E}(r_k^1) = \dots = \mathbb{E}(r_k^4)$.
One extreme case of bandwidth allocation is UE 1 and UE 2 share the total bandwidth $B$ in the first round, each of which is assigned $\frac{B}{2}$. At the same time, UE 3 and UE 4 can complete their local computation during round 1. Then, at round 2, all bandwidth $B$ is allocated to UE 3 and UE 4 for their gradients transmission. In this case, according to Theorem~\ref{thm:2}, in each round, both UEs will finish their gradient transmission at the same time. That is, the duration of round 1 will be minimized when UE 1 and UE 2 share the total bandwidth $B$ equally. At this point, the round duration is $\frac{Z}{r(B/2)}$, where $r(B/2) = \frac{B}{2} \ln(1+\frac{2p_ih_i\|c_i\|^{-\kappa}}{B N_0})$. Similarly, the duration of round 2 is also $\frac{Z}{r(B/2)}$. Then, the total time of each period is $\frac{2Z}{r(B/2)}$.
The other extreme case of bandwidth allocation is for all the four UEs to share the bandwidth equally, then the UEs will finish one time of global update at the same time, which is computed by $\frac{Z}{r(B/4)}$. Note that we set $A=2$, but in this case if all UEs finish one communication round at the same time then $A=4$, therefore this extreme situation cannot be achieved but can only be approached infinitely.
It is obvious $\frac{Z}{r(B/4)} = \frac{2Z}{r(B/2)}$, this equation indicates that all bandwidth allocation policies between the two extreme cases can lead to the same minimized overall training time.

At this point, according to the features of the optimal bandwidth solutions, we obtain four corollaries. Corollary~\ref{cor:2} and \ref{cor:3} are two direct conclusions derived from Theorem~\ref{thm:2}, which are shown as follows,

\begin{cor} \label{cor:2}
  From Theorem~\ref{thm:2}, we find that in each round $k$, UEs in $\mathcal{A}_k$ will finish the communication round at the same time. That is, none of the UEs have to wait for the others under the optimal bandwidth allocation policy. Therefore, we have $\sum_{k=1}^{K}\max_{i\in\mathcal{A}}\{T_k^i\} = \sum_{k=1}^{K} T_k^{i*} = T_i^*$ ($\forall i\in\mathcal{U}$).
\end{cor}

\begin{cor} \label{cor:3}
  The optimal overall training time is equivalent to the optimal total training time of any UE $i$ from a long-term perspective when $K\rightarrow + \infty$. That is, $T^*(\mathbf{\Pi}) = T_i^*$ ($\forall i\in\mathcal{U}$ and a large $K$).
\end{cor}

Next, according to Theorem~\ref{thm:4}, we extract Corollary~\ref{cor:4} to characterize the optimal solutions of $Z_k^i$, which is determined right after the computation of $b_k^i$.

\begin{cor} \label{cor:4}
  There exists infinitely many solutions of $Z_k^i$ as long as the bandwidth allocation follows the results shown in Theorem~\ref{thm:4}. Meanwhile, $Z_k^i$ is in a range of values from $0$ to $Z$.
\end{cor}

At last, we introduce Corollary~\ref{cor:5} to describe the relationship between the relative participation frequency $\eta_i$ and the optimal overall training time $T^*(\mathbf{\Pi})$.

\begin{cor} \label{cor:5}
  There is a tradeoff between the relative participation frequency $\eta_i$ ($i\in\mathcal{U}$) and the optimal overall training time $T^*(\mathbf{\Pi})$. As long as $\eta$ is defined or determined, then according to Theorem~\ref{thm:3} the circular scheduling pattern $\mathbf{\Pi}$ can be determined. With the scheduling pattern $\mathbf{\Pi}$, according to Theorem~\ref{thm:4}, the optimal bandwidth allocation and the corresponding optimal overall training time $T^*(\mathbf{\Pi})$ can be determined.
\end{cor}

\subsection{Scheduling Policy} \label{sec:5.3}

Based on the optimal bandwidth $b_k^i$ obtained from P2, we turn to P3 to solve the UE scheduling problem.
From (C3.2) we have
\begin{equation}
  \eta_i AK = \sum_{k=1}^{K} \pi_k^i \geq \frac{K}{S}, \quad \forall i\in \mathcal{U},
\end{equation}
which can be further simplified to $A \geq \frac{1}{\eta_i S}$.
Meanwhile, note that the minimization of $F(w)$ can be approximated by minimizing the upper bound of $\frac{1}{K} \sum_{k=0}^{K-1} \mathbb{E}[\|\nabla F(w_k)\|^2] $ according to Theorem~\ref{thm:1}. Therefore, P3 can be approximated by P4 as follows:
\begin{subequations}\label{equ:opt4}
\begin{align}
  \min_{K, A, \mathbf{\Pi}} \quad & \frac{2(F(w_0) - F(w^*))}{\beta K} \nonumber \\
  & + 4(L_F\beta +2 L_F^2\beta^2S^2)(\sigma_F^2+\gamma_F^2)\sqrt{A} \label{equ:opt4_obj} \tag{P4} \\
  s.t. \quad & T_i^* = T^*(\mathbf{\Pi}), \quad \forall i\in\mathcal{U} \label{equ:opt4_const1} \tag{C4.1} \\
  & A \geq \frac{1}{\eta_i S}, \quad  \forall i\in\mathcal{U}  \label{equ:opt4_const2} \tag{C4.2} \\
  & K \geq \frac{S}{\eta_i},  \quad  \forall i\in\mathcal{U}, \label{equ:opt4_const3} \tag{C4.3}
\end{align}
\end{subequations}
where (\ref{equ:opt4_const1}) is derived from Corollary~\ref{cor:2} and~\ref{cor:3}.

The relationship between $A$ and $K$ has been coarsely analysed in Corollary~\ref{cor:1}, where $K=\mathcal{O}(\epsilon^{-3})$ and $A = \mathcal{O}(\epsilon^{-2})$. This means that the optimal $K^*$ and $A^*$ can only be estimated in the implementation. Let the first term and the second term of the objective of P4 be equal to $\epsilon$ respectively, the optimal solution of $K$ and $A$ can be approximated by
\begin{align}
  K^* & \thickapprox \min_{i\in \mathcal{U}}\{\frac{2(F(w_0)-F(w^*))}{\beta \epsilon}, \frac{S}{\eta_i}\} \\
  A^* & \thickapprox \min_{i\in \mathcal{U}} \{\frac{\epsilon^2}{16(L_F\beta +2 L_F^2\beta^2S^2)^2(\sigma_F^2+\gamma_F^2)^2}, \frac{1}{\eta_i S}\}.
\end{align}

With the optimal value $A^*$, we use a greedy algorithm to generate the scheduling policy matrix $\Pi$, which is shown in Algorithm~\ref{alg:schduling_policy}. In each round $k$, the algorithm is always picking up the UE $i$ with the smallest current relative participation frequency $\hat{\eta}_i$, if $\hat{\eta}_i < \eta_i$ then the algorithm sets $\pi_k^i=1$. Then the algorithm picks up the second poorest UE $j$ and set $\pi_k^j=1$. This process repeats until $A^*$ UEs are picked up in round $k$. For the next round $k+1$, the same process repeats. In this way, the circular scheduling pattern can be achieved and $\mathbf{\Pi}$ is obtained.

\begin{algorithm}[t]
\caption{Greedy PerFedS$^2$ Scheduling Algorithm}
\label{alg:schduling_policy}
\SetKwInOut{Input}{Input}
\Input{$\eta = \{\eta_1,\eta_2,\dots,\eta_n\}$, $A^*$ }
Initialize $\Pi \leftarrow \varnothing$ \;
\For{$k=1$ \KwTo $K$}{
    \For{$i=1$ \KwTo $N$}{
        \eIf{the total number of global updates $sum(\Pi) = 0$}{
            $\hat{\eta}_i = 0$ \;
        }{
            $\hat{\eta}_i =\frac{\text{number of overall updates of UE } i}{\text{number of overall global updates}}= \frac{sum(\Pi[:,i])}{sum(\Pi)}$\;
        }
        \eIf{current number of updates in round $k$ $sum(\Pi[k,:]) < A^*$ and current relative participation frequency of UE $i$ $\hat{\eta}_i \leq \eta_i$}{
            Set $\Pi[k][i] \leftarrow 1$\;
            \If{current number of updates in round $k$ $sum(\Pi[k,:])<A^*$}{
                Schedule the first $A^*-sum(\Pi[k,:])$ UEs in current round $k$\;
                i.e., $\Pi[k][0:A^*-sum(\Pi[k,:])] = 1$ \;
            }
        }{
           $\Pi[k][i] = 0$\;
        }
    }
}
\end{algorithm}

\section{Performance Evaluation} \label{sec:6}

In this section, we conduct extensive experiments to ($i$) verify the effectiveness of PerFedS$^2$ in saving the overall training time and ($ii$) examine the effects of different system parameters on the performance of PerFedS$^2$.

\subsection{Setup} \label{sec:6.1}

\subsubsection{Datasets and Models}

\begin{table}
\centering
\caption{System Parameters}
\label{tab:sys_para}
\begin{tabular}{|c|c|}
  \hline
  \textbf{Parameter} & \textbf{Value}  \\
  \hline
  $\alpha$ (MNIST) & $0.03$ \\
  \hline
  $\beta$ (MNIST) & $0.07$ \\
  \hline
  $\alpha$ (CIFAR-100)& $0.02$ \\
  \hline
  $\beta$ (CIFAR-100)& $0.06$ \\
  \hline
  $\alpha$ (Shakespeare)& $0.03$ \\
  \hline
  $\beta$ (Shakespeare) & $0.07$\\
  \hline
  $B$ & 1 MHz \\
  \hline
  $\kappa$ & $3.8$\\
  \hline
  $N_0$ & $-174$ dBm/Hz \\
  \hline
  $p_i$ & $0.01$ W\\
  \hline
\end{tabular}
\end{table}

We consider an FL system that contains multiple UEs located in a cell of radius $R=200$ m and a BS located at the center. Meanwhile, the Rayleigh distribution parameter of $h_k^i$ across communication rounds is $40$. We conduct the experiments using three datasets: MNIST~\cite{MNIST}, CIFAR-100~\cite{CIFAR10} and the Shakespeare~\cite{caldas2018leaf} datasets.
The network model we used for MNIST is a 2-layer deep neural network (DNN) with hidden layer of size 100.
The network model we used for CIFAR-100 is LeNet-5~\cite{lecun1998gradient} that contains two convolutional layers and three fully connected layers.
And the network model we used for the Shakespeare dataset is an LSTM classifier. The number of UEs under the MNIST and the CIFAR-100 datasets is set to be 20, and the number of UEs under the Shakespeare dataset for next-character prediction is 188.
The other parameters used in the experiments are summarized in Table~\ref{tab:sys_para}.

\subsubsection{Baselines}

We compare PerFedS$^2$ with three benchmarks: synchronous, semi-synchronous, and asynchronous FL algorithms. For the synchronous FL benchmark, we consider three algorithms, FedAvg, FedProx~\cite{li2020federated}, and Per-FedAvg (termed as FedAvg-SYN, FedProx-SYN and PerFed-SYN in the figures). FedProx is a FL algorithm that deals with heterogenous datasets. For the semi-synchronous benchmark, we consider only two algorithms besides PerFedS$^2$, semi-synchronous Federated Learning (FedAvgS$^2$), which is a semi-asynchronous FL algorithm, and semi-synchronous FedProx (FedProxS$^2$). For the asynchronous FL benchmark we consider three algorithms, FedAvg-ASY, FedProx-ASY and PerFed-ASY. The above three algorithms are asynchronous FL mechanisms, where the server performs the global updating as soon as it receives a local model from any UE.

\subsubsection{Dataset Participation}
The level of divergence in the distribution of UEs' datasets will affect the overall performance of the system.
To reflect this feature, each UE is allocated a different local data size and has $l = {1,2,\dots,10}$ of the 10 labels, where $l$ denotes the level of data heterogeneity, the higher $l$ is, the more diverse the datasets are.

\subsubsection{Relative Participation Frequency Setting}
The relative participation frequency plays a critical role in the system performance as it determines not only the scheduling pattern but also the minimal overall training time.
In practice, there are many factors that may affect the value of $\eta$. For example, the distances from UEs to the server and the transmit power of each UE.
In this paper, we use two sets of $\eta$.
For the first one, we consider all the UEs have the same $\eta_i$, i.e., $\eta_1=\eta_2=\dots=\eta_n$.
For the second one, we consider the distances from the UEs to the server is uniformly distributed, while the other parameters of the UEs are the same.
Under this setting, the values of $\eta_i$ among the UEs are unbalanced.

\begin{figure*}[!t]
  \centering
  \subfloat[MNIST training loss]{
      \includegraphics[width=1.7in]{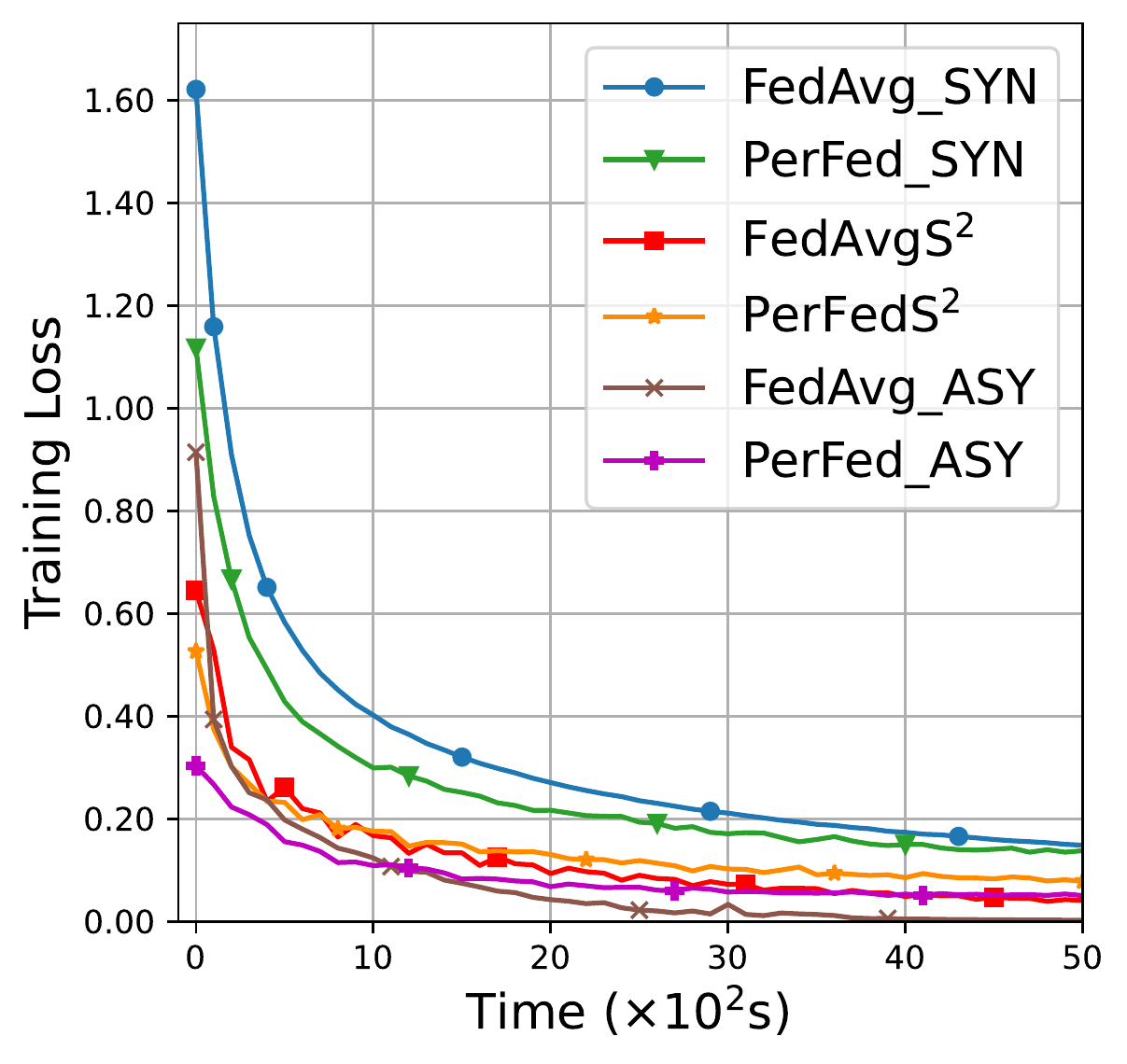}
      \label{fig_exp_1:subfig:a}}
  \subfloat[MNIST test accuracy]{
      \includegraphics[width=1.7in]{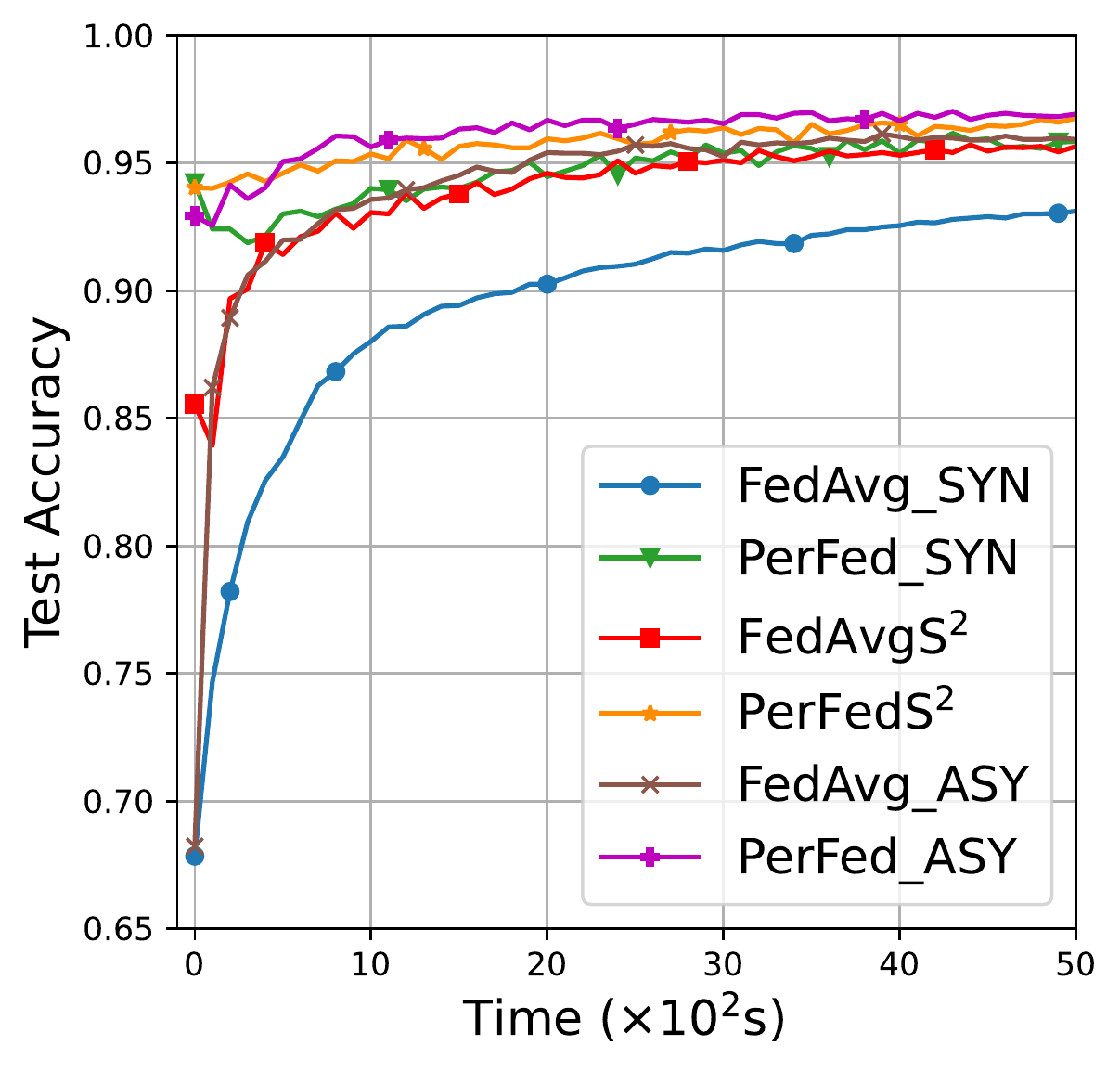}
      \label{fig_exp_1:subfig:b}}
  \subfloat[CIFAR-100 training loss]{
      \includegraphics[width=1.7in]{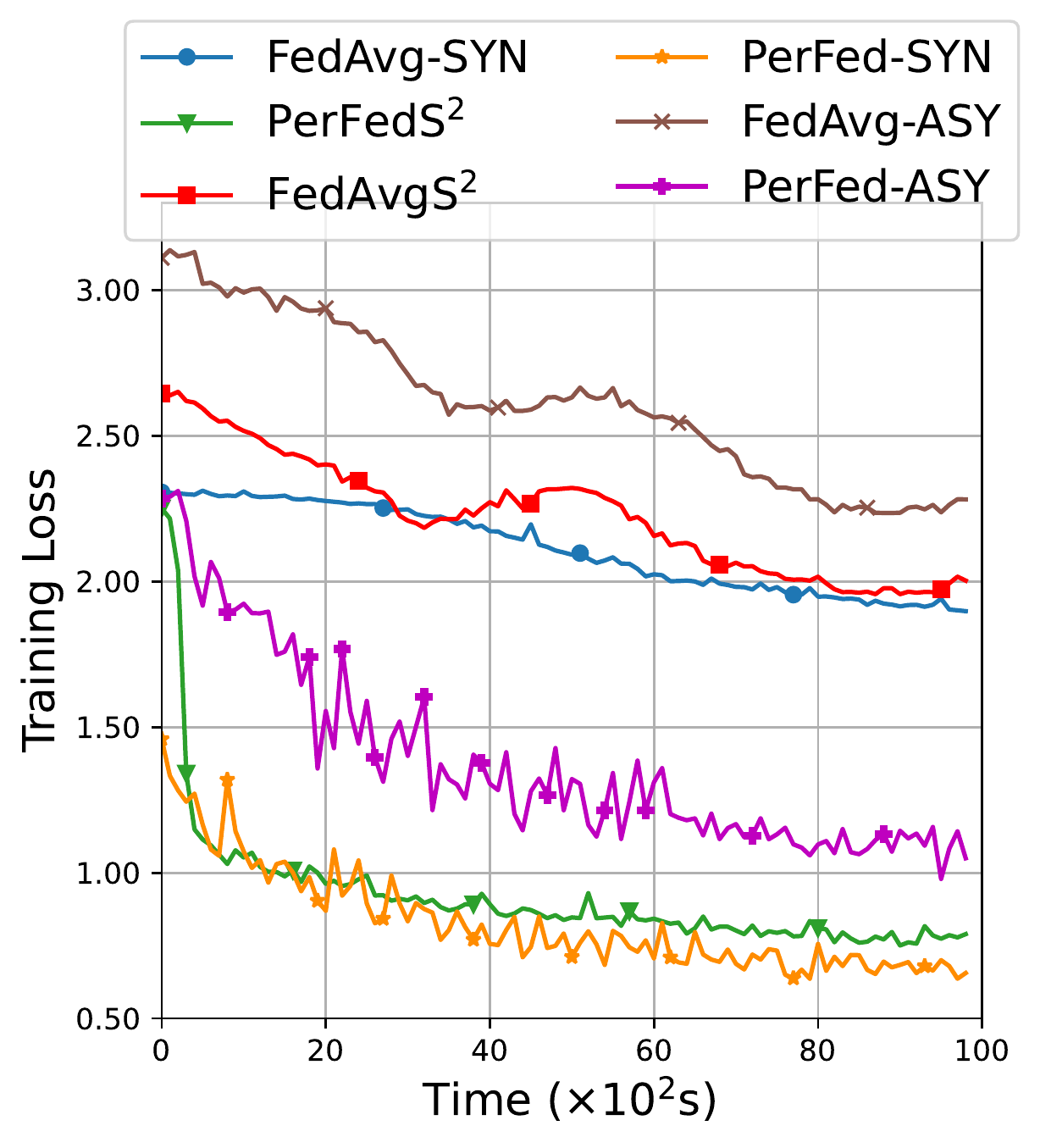}
      \label{fig_exp_1:subfig:c}}
  \subfloat[CIFAR-100 test accuracy]{
      \includegraphics[width=1.7in]{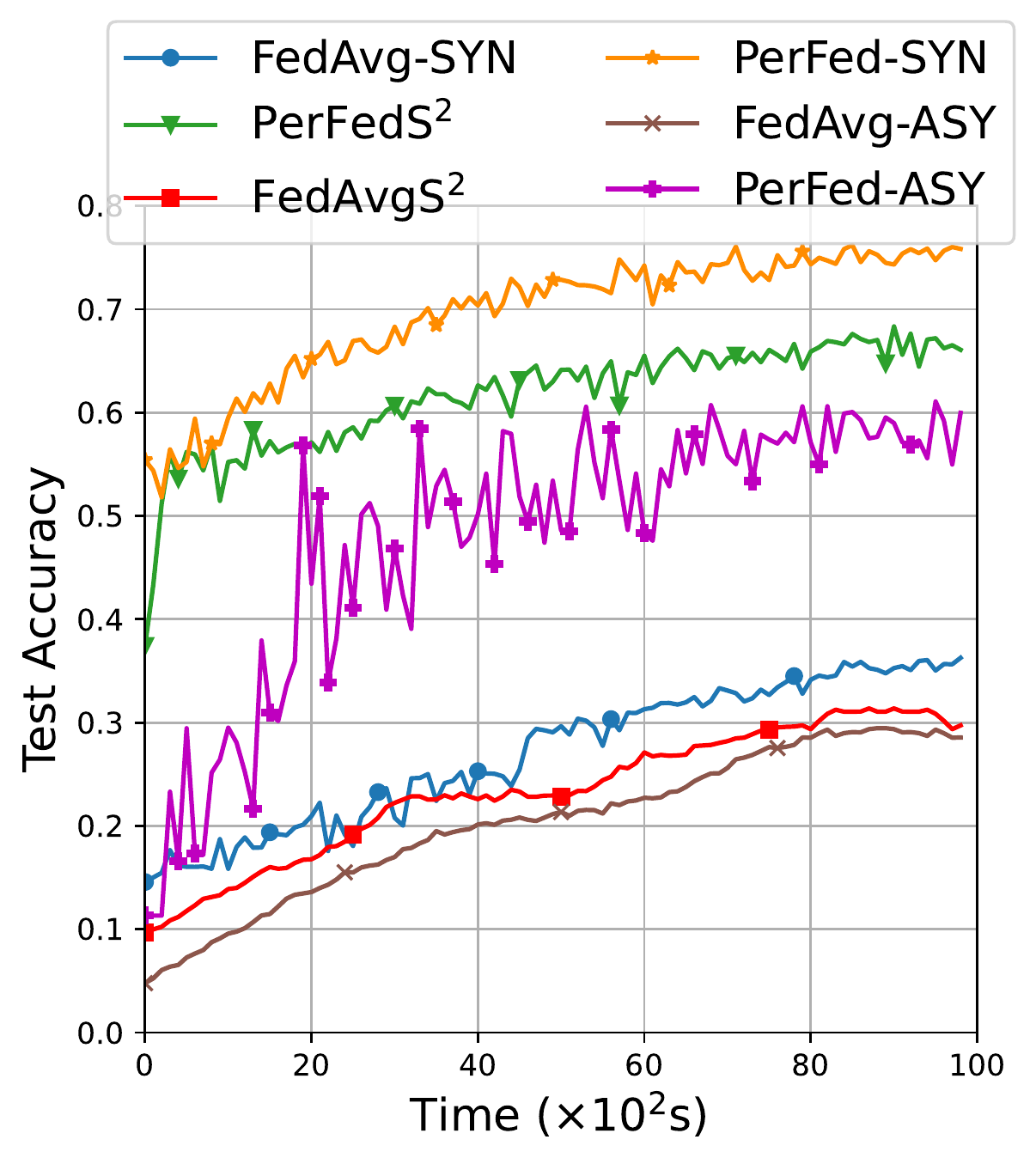}
      \label{fig_exp_1:subfig:d}}
  \caption{Convergence performance comparison of PerFedS$^2$, FedAvgS$^2$, FedAvg-SYN, PerFed-SYN, FedAvg-ASY and PerFed-ASY using MNIST and CIFAR-100 datasets. In this case, $\eta_1=\eta_2=\dots=\eta_n$. Meanwhile, as for the PerFedS$^2$ and FedAvgS$^2$ algorithms, we set $A=5$. }
  \label{fig:exp_1}
\end{figure*}

\begin{figure*}[!t]
  \centering
  \subfloat[MNIST training loss]{
      \includegraphics[width=1.7in]{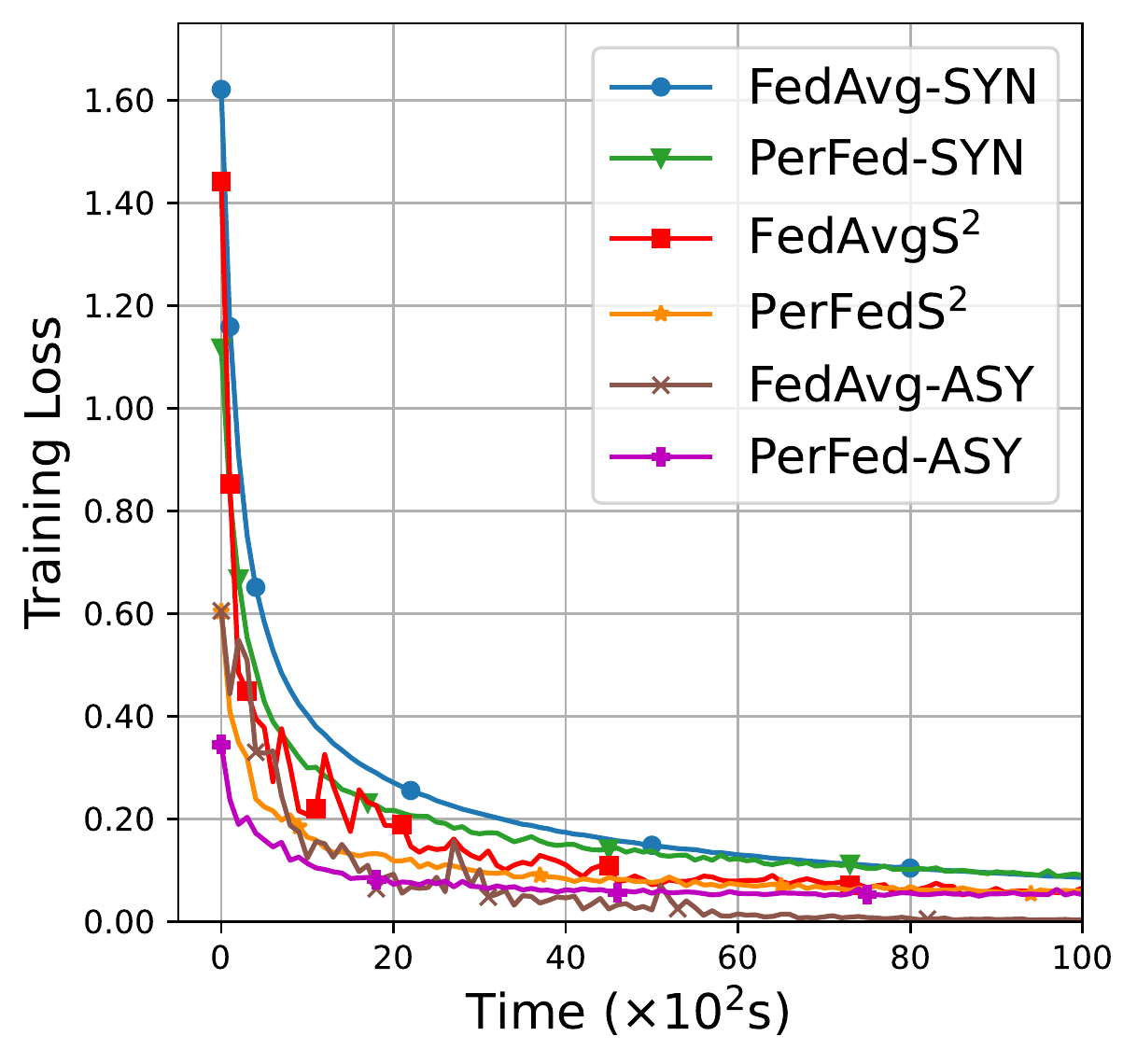}
      \label{fig_exp_2:subfig:a}}
  \subfloat[MNIST test accuracy]{
      \includegraphics[width=1.7in]{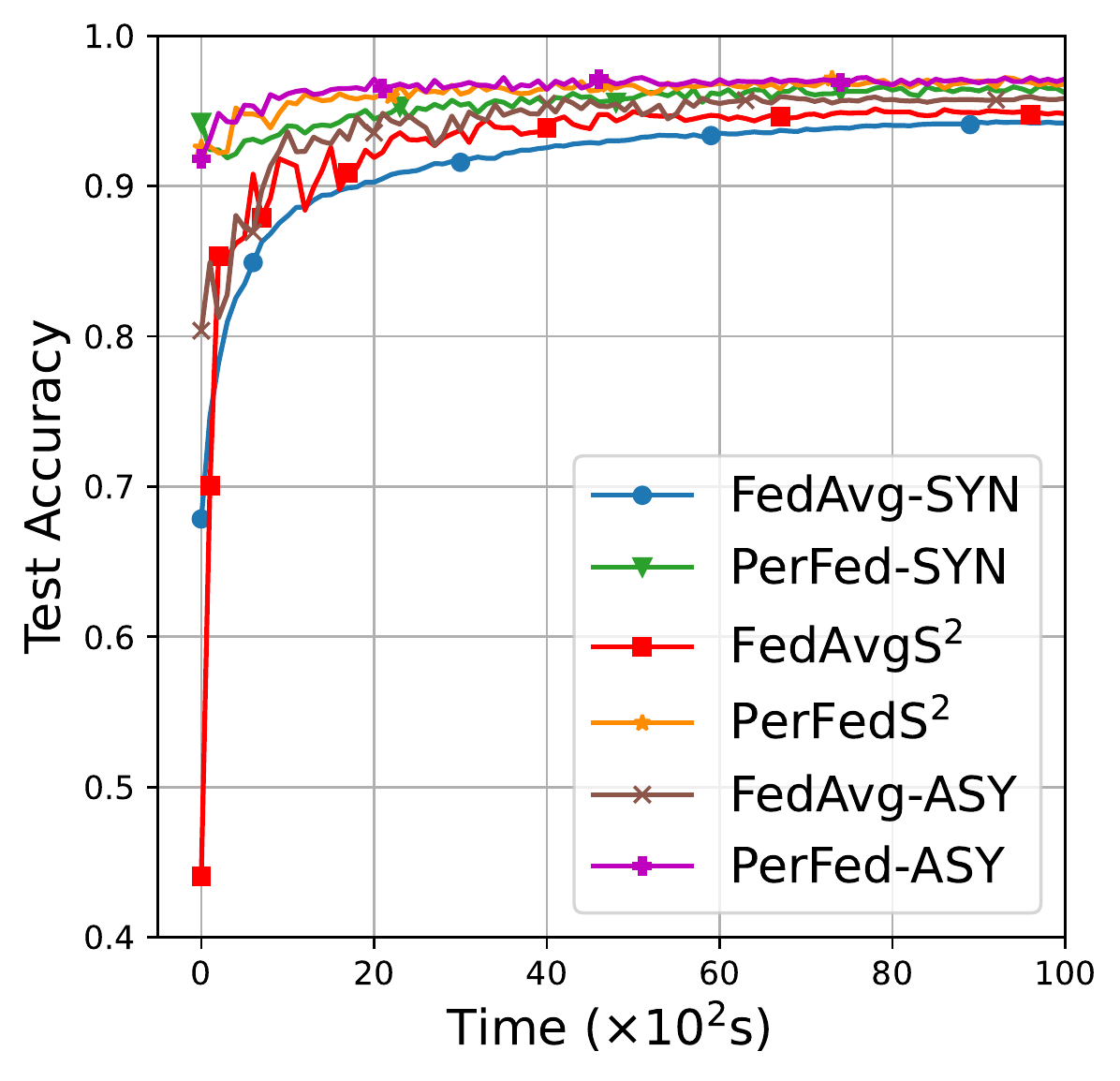}
      \label{fig_exp_2:subfig:b}}
  \subfloat[CIFAR-100 training loss]{
      \includegraphics[width=1.7in]{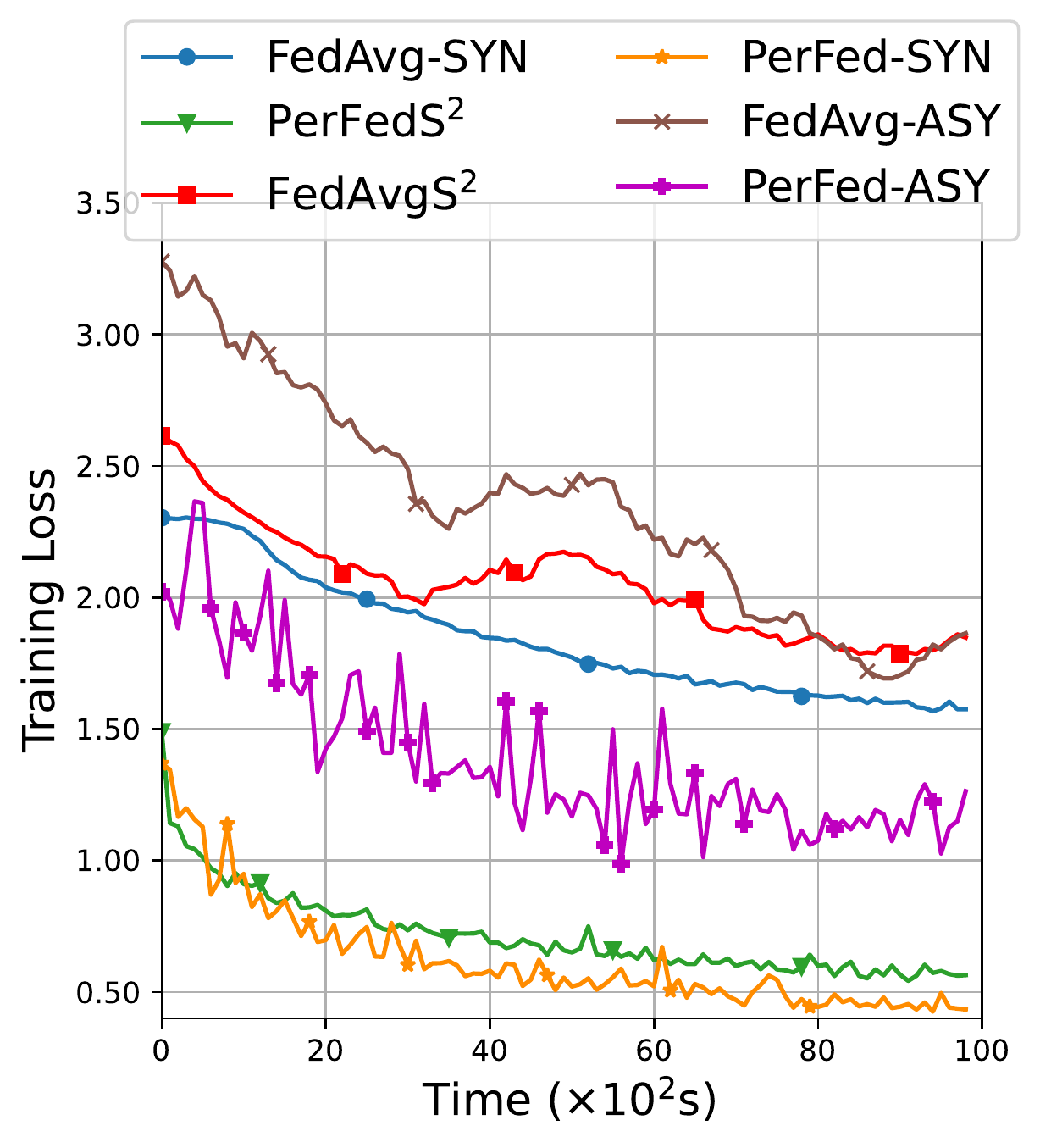}
      \label{fig_exp_2:subfig:c}}
  \subfloat[CIFAR-100 test accuracy]{
      \includegraphics[width=1.7in]{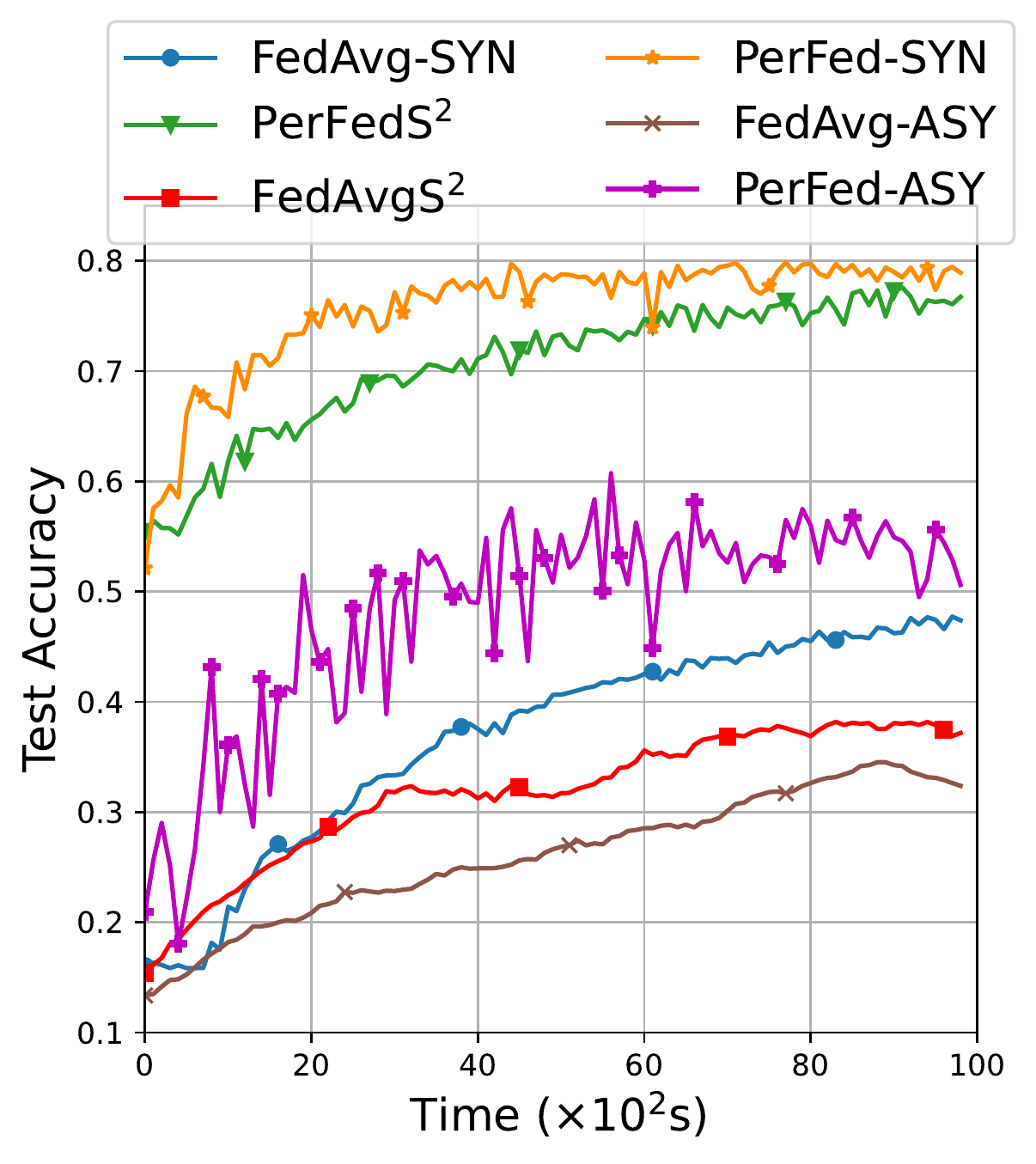}
      \label{fig_exp_2:subfig:d}}
  \caption{Convergence performance comparison of PerFedS$^2$, FedAvgS$^2$, FedAvg-SYN, PerFed-SYN, FedAvg-ASY and PerFed-ASY using MNIST and CIFAR-100 datasets. In this case, the distance from UEs to the server obeys the random distribution from 0 to 200 m. Meanwhile, as for the PerFedS$^2$ and FedAvgS$^2$ algorithms, we set $A=5$.}
  \label{fig:exp_2}
\end{figure*}

\begin{figure*}[!t]
  \centering
  \subfloat[Shakespeare training loss]{
      \includegraphics[width=1.7in]{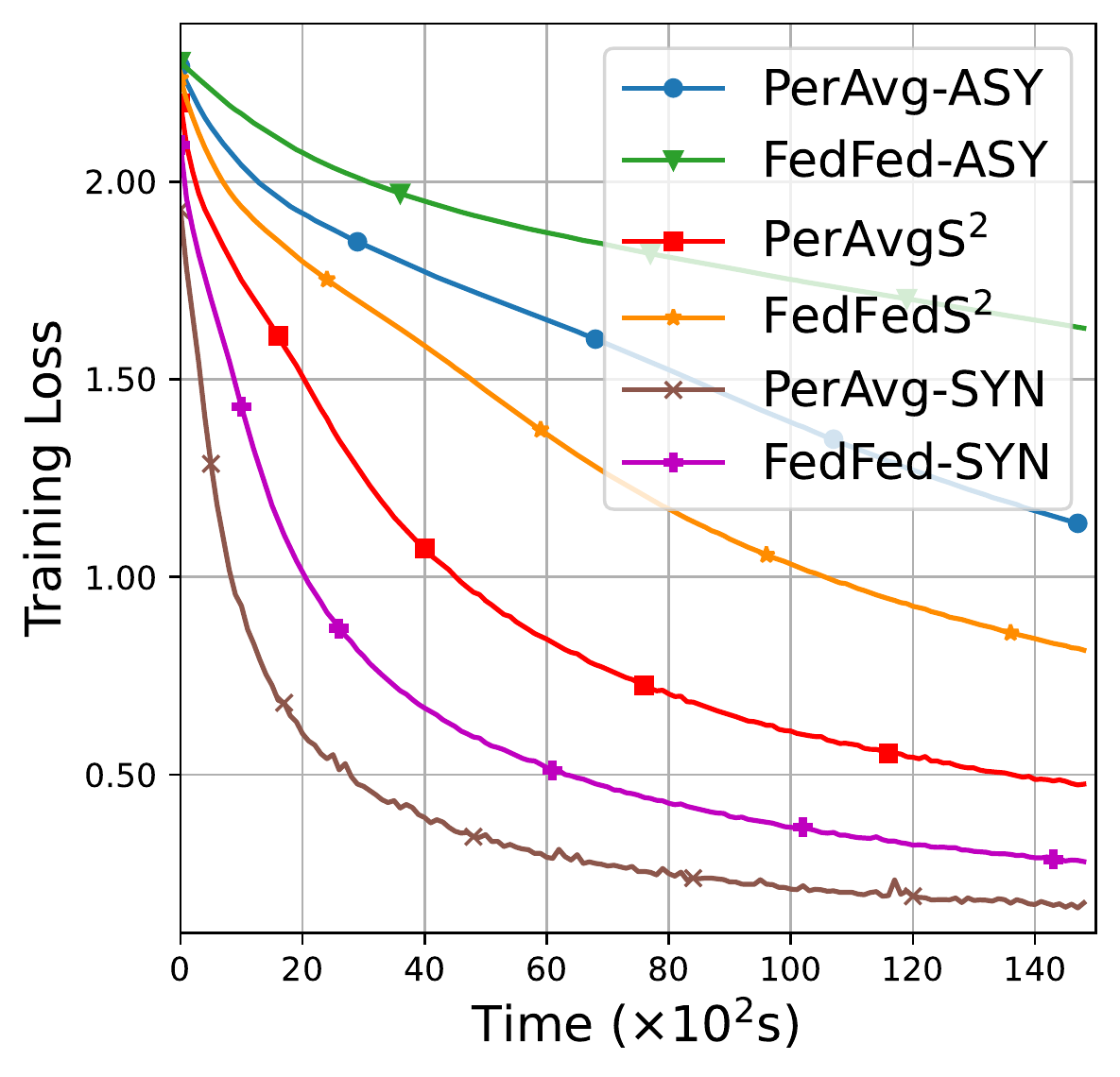}
      \label{fig_exp_7:subfig:a}}
  \subfloat[Shakespeare test accuracy]{
      \includegraphics[width=1.7in]{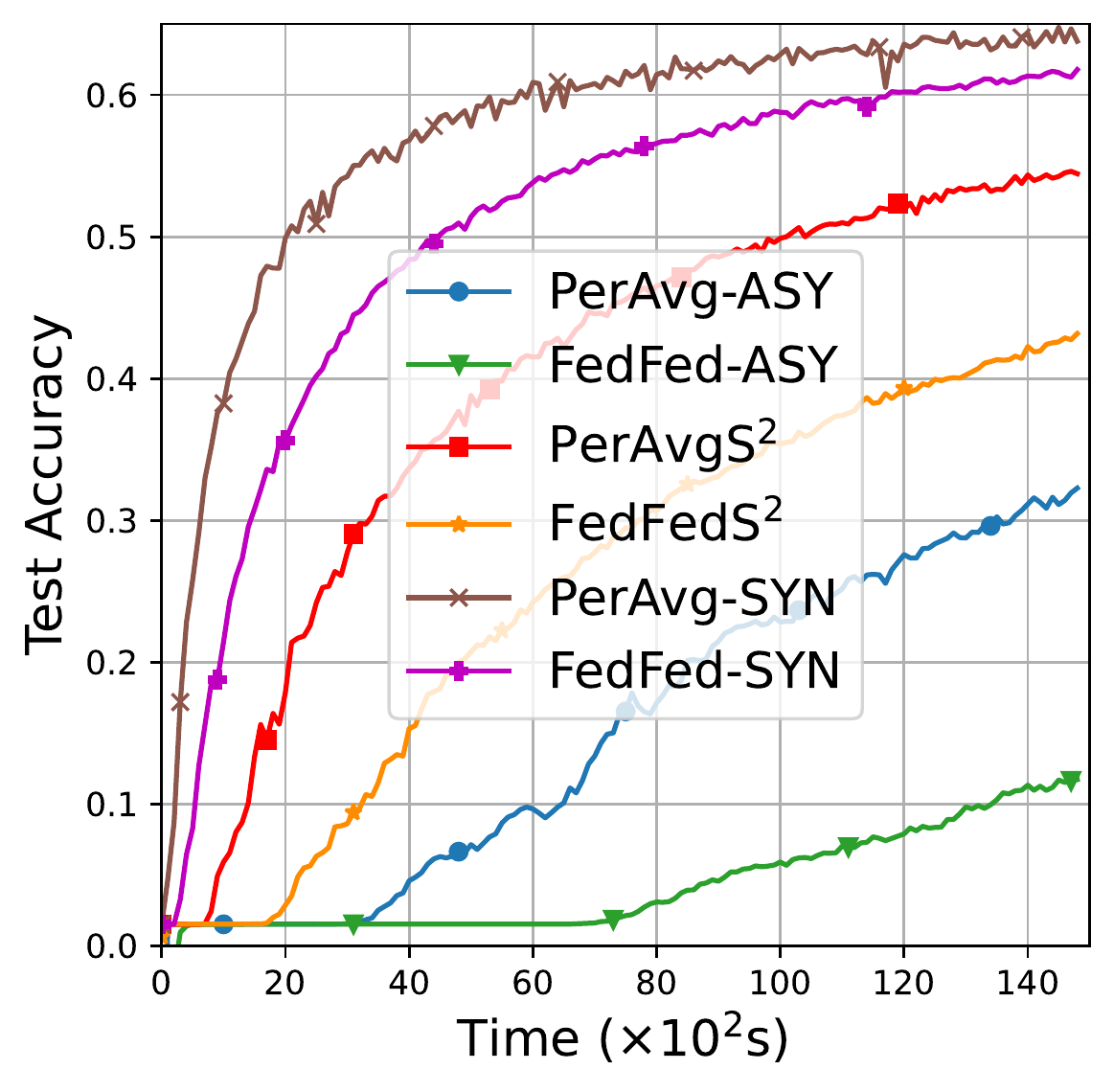}
      \label{fig_exp_7:subfig:b}}
  \subfloat[Shakespeare training loss]{
      \includegraphics[width=1.7in]{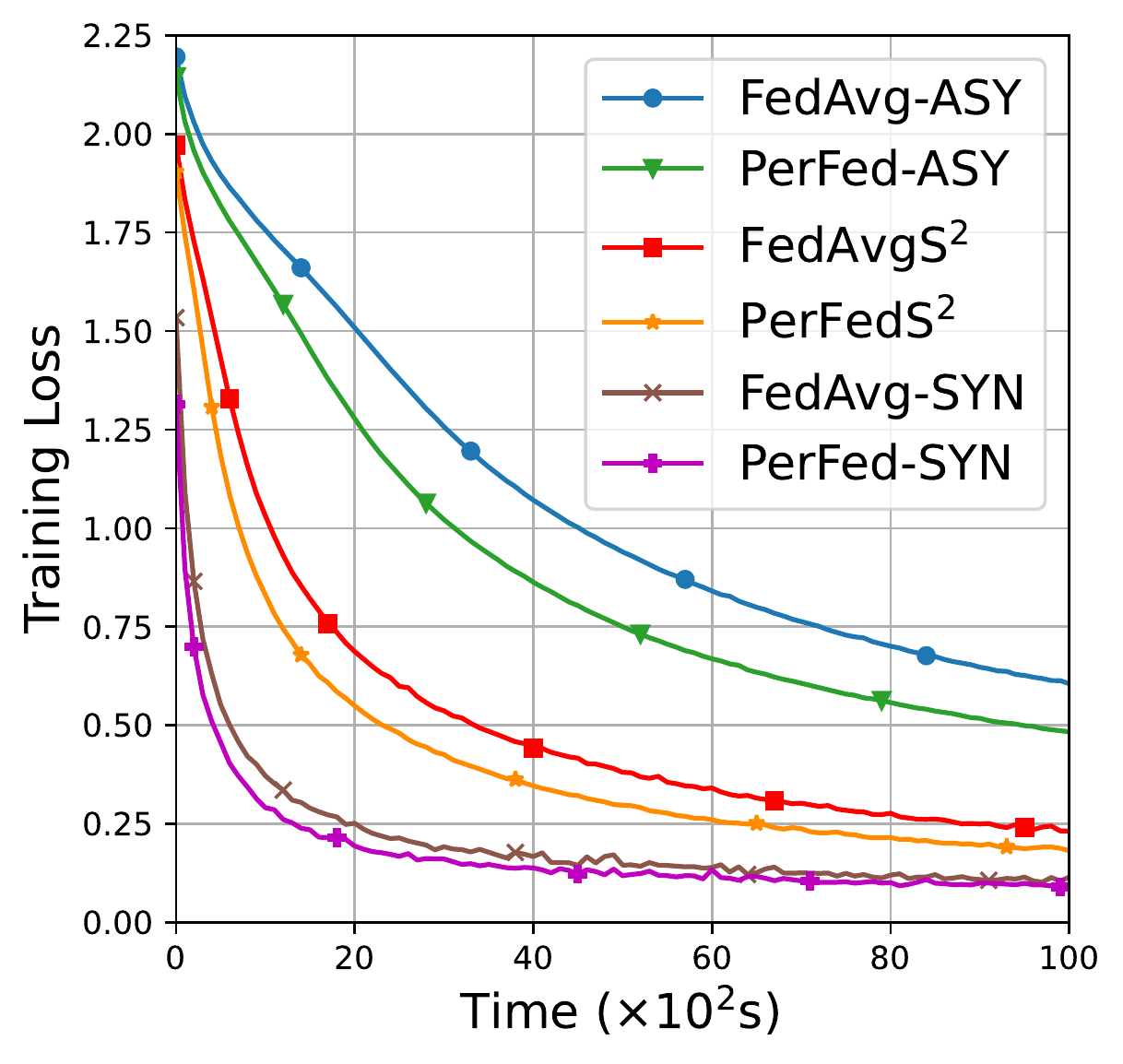}
      \label{fig_exp_7:subfig:c}}
  \subfloat[Shakespeare test accuracy]{
      \includegraphics[width=1.7in]{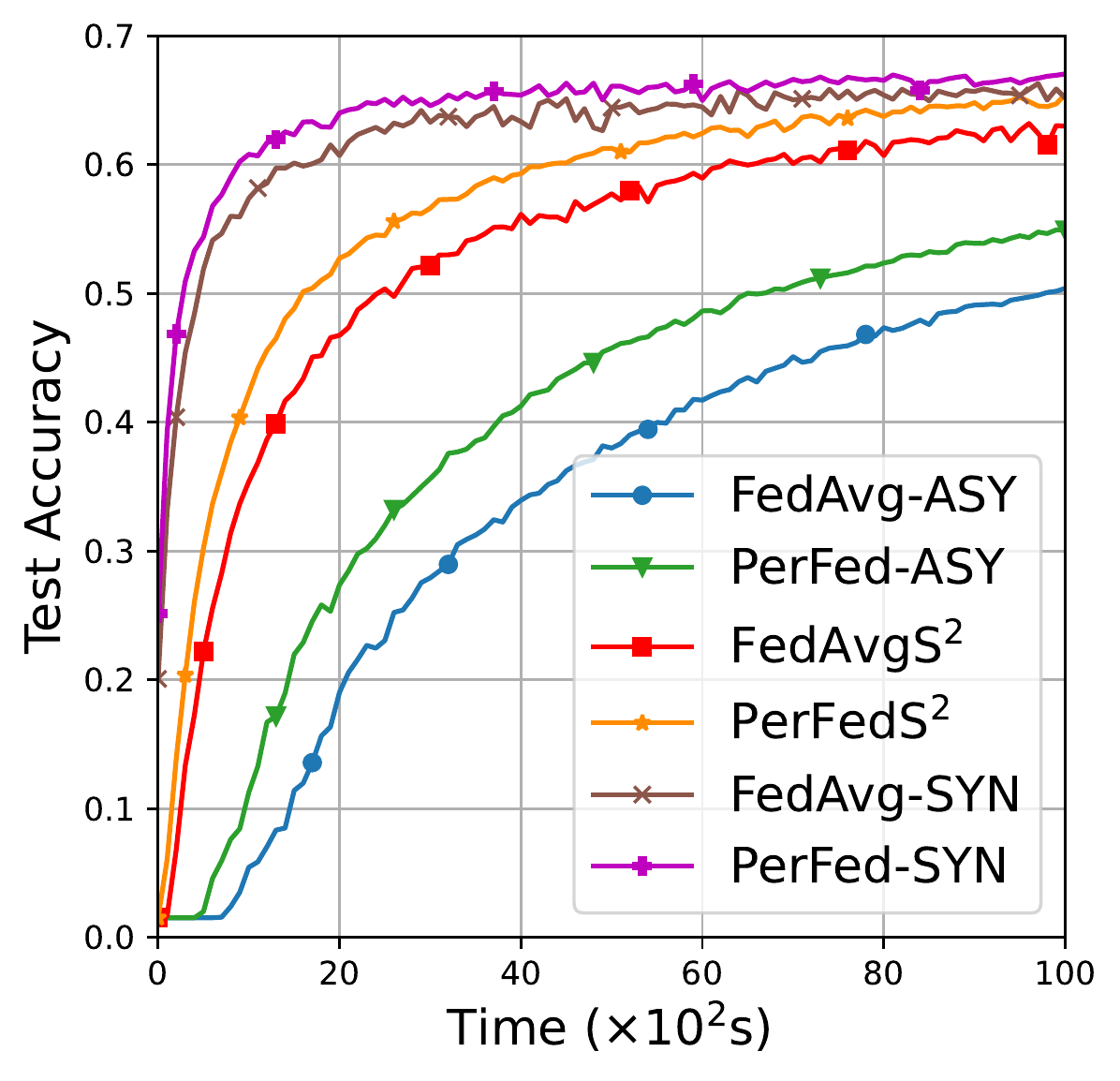}
      \label{fig_exp_7:subfig:d}}
  \caption{Convergence performance comparison of PerFedS$^2$, FedAvgS$^2$, FedAvg-SYN, PerFed-SYN, FedAvg-ASY and PerFed-ASY using the Shakespeare dataset. For (a) and (b), $\eta_1=\eta_2=\dots=\eta_n$, and for (c) and (d), the distance from UEs to the server obeys the random distribution from 0 to 200 m. Meanwhile, as for the PerFedS$^2$ and FedAvgS$^2$ algorithms, we set $A=50$.}
  \label{fig:exp_7}
\end{figure*}

\begin{figure*}[!t]
  \centering
  \subfloat[MNIST]{
      \includegraphics[width=1.7in]{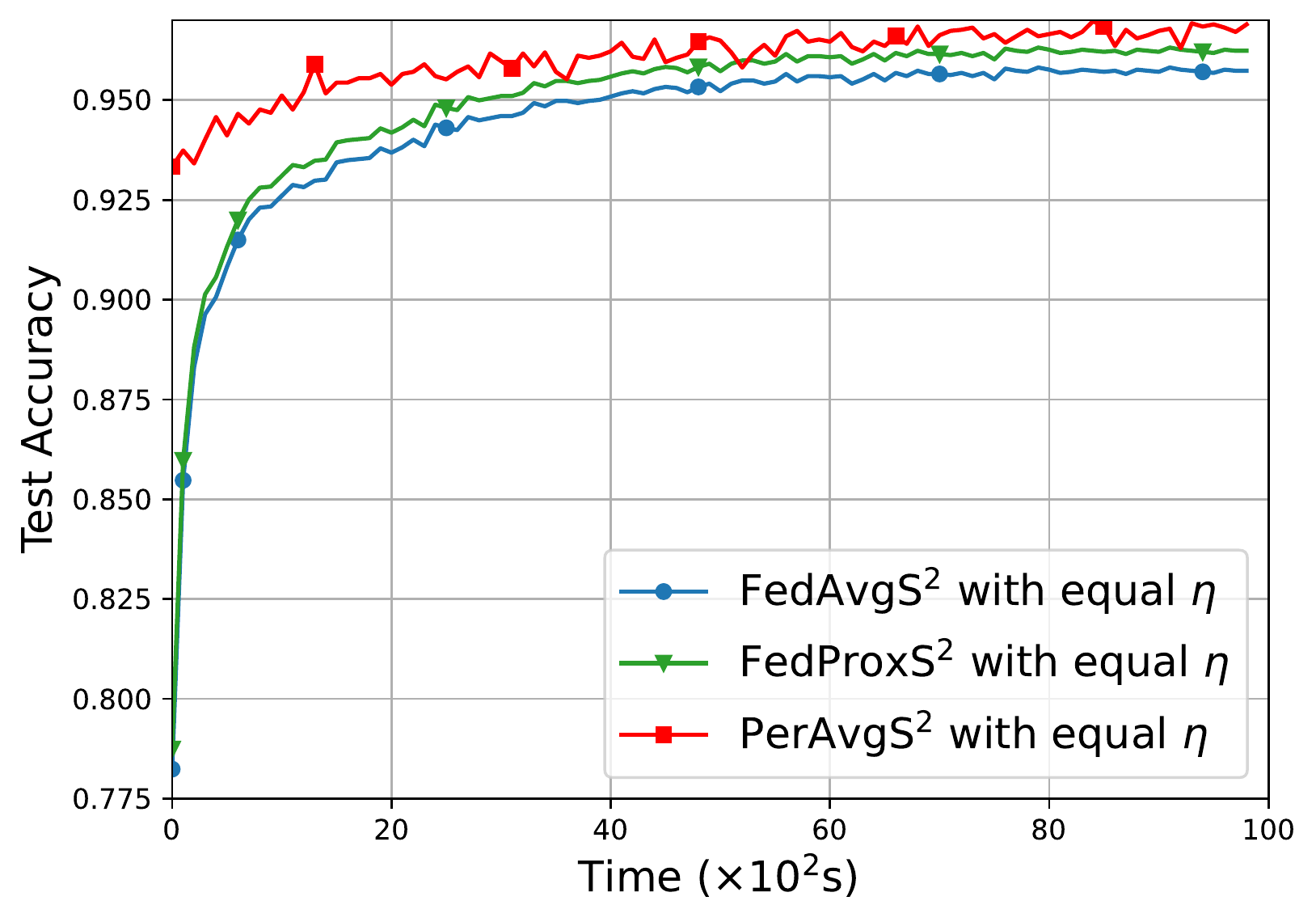}
      \label{fig_exp_8:subfig:a}}
  \subfloat[MNIST]{
      \includegraphics[width=1.7in]{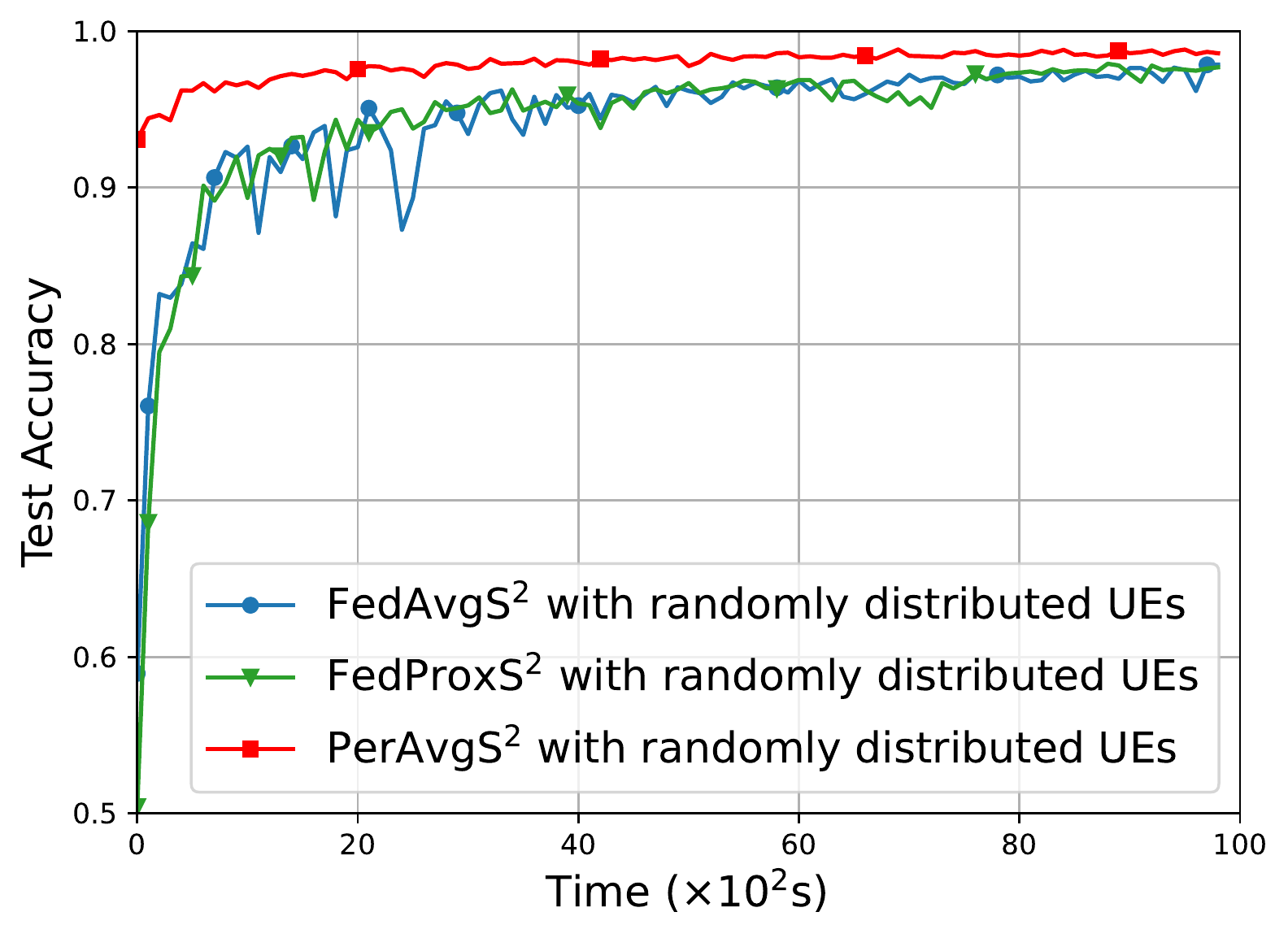}
      \label{fig_exp_8:subfig:b}}
  \subfloat[Shakespeare]{
      \includegraphics[width=1.7in]{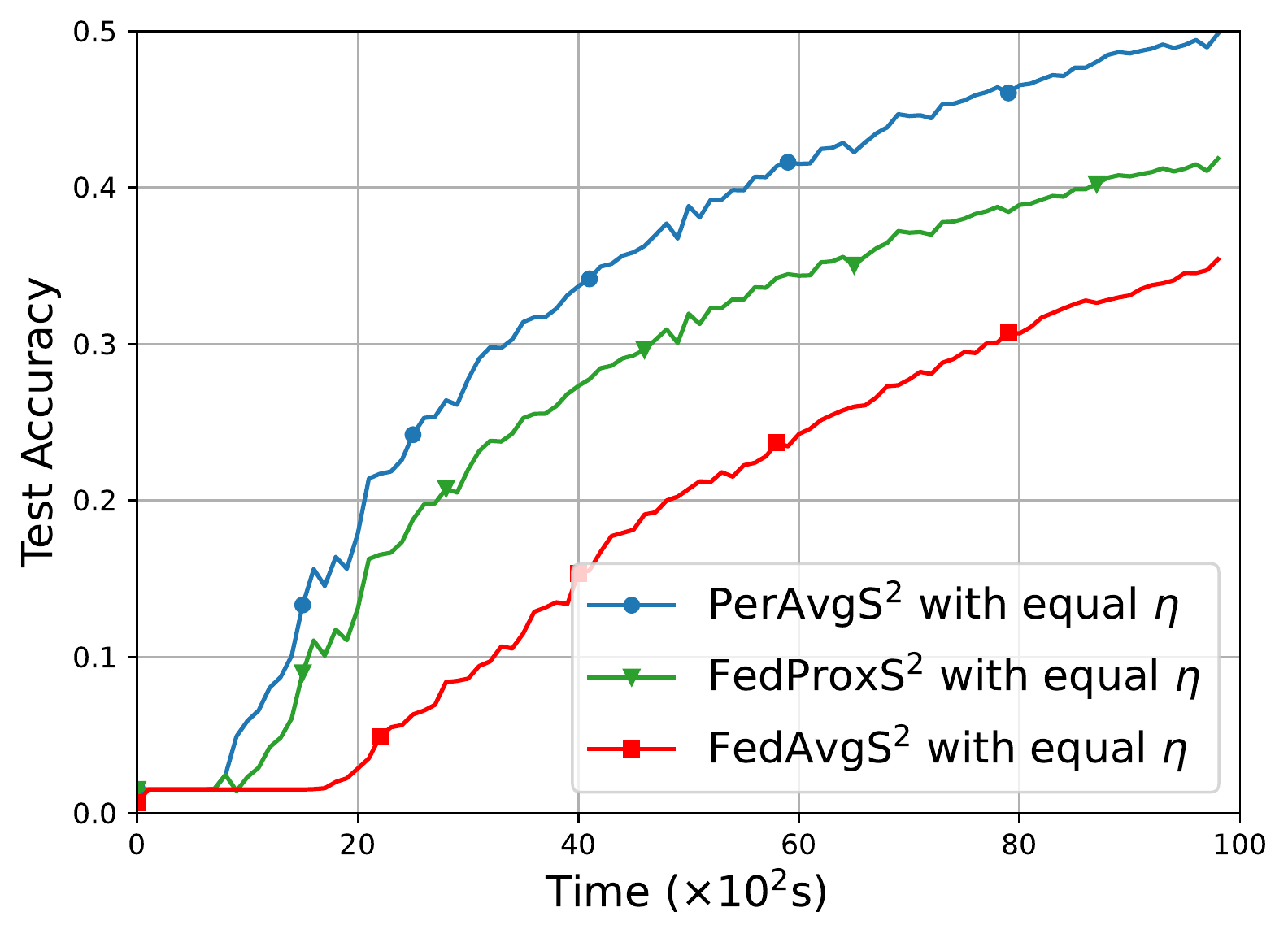}
      \label{fig_exp_8:subfig:c}}
  \subfloat[Shakespeare]{
      \includegraphics[width=1.7in]{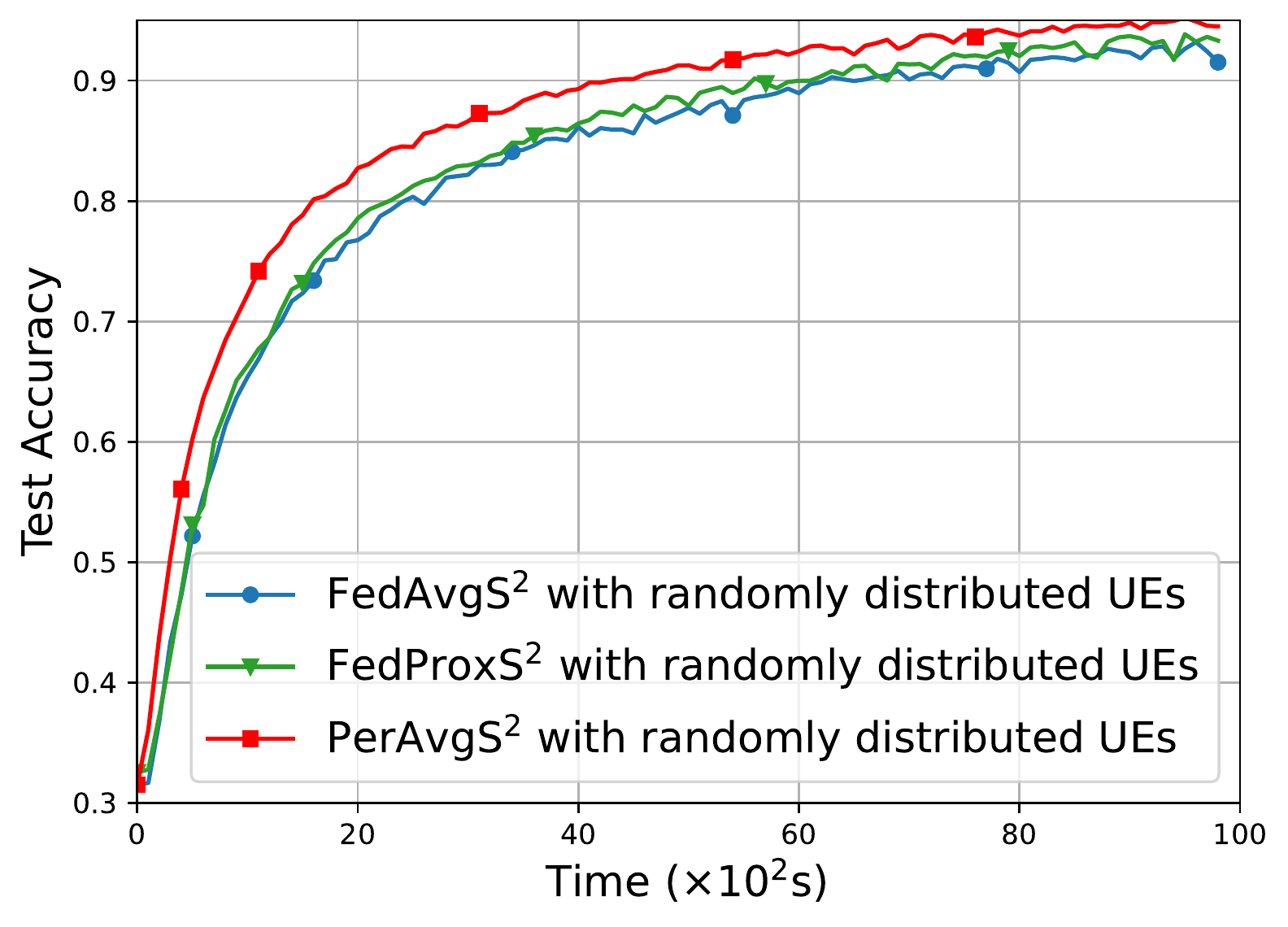}
      \label{fig_exp_8:subfig:d}}
  \caption{Convergence performance comparison of PerFedS$^2$, FedAvgS$^2$ and FedProxS$^2$. For (a), we use the MNIST dataset and $\eta_1=\eta_2=\dots=\eta_n$. For (b), we use the MNIST dataset and the distance from UEs to the server obeys the random distribution from 0 to 200 m. For (c), we use the Shakespeare dataset and $\eta_1=\eta_2=\dots=\eta_n$. And for (d), we use the Shakespeare dataset and the distance from UEs to the server obeys the random distribution from 0 to 200 m. Meanwhile, we set $A=5$ for the MNIST dataset and $A=50$ for the Shakespeare dataset.}
  \label{fig:exp_8}
\end{figure*}

\subsection{Evaluation Results} \label{sec:6.2}

\subsubsection{Effect of relative participation frequency $\eta$}
Fig.~\ref{fig:exp_1} shows the convergence performance comparison between PerFedS$^2$ and other five FL and PFL algorithms, where UEs have the same $\eta_i$, and $A=5$.
Then Fig.~\ref{fig:exp_2} shows the convergence performance comparison of the six algorithms, where the $\eta_i$ of each UE is determined by its distance to the server, and the distance is uniformly distributed from 0 to 200 m.
At last, Fig.~\ref{fig:exp_7} shows the convergence comparison of the six algorithms using Shakespeare dataset, where $A=50$.

From both figures, we find that for MNIST, generally, it takes synchronous algorithms the most time to achieve the same convergence performance compared with semi-synchronous and asynchronous algorithms, then asynchronous algorithms behaves the best. However, for the CIFAR-100 dataset, generally, semi-synchronous algorithms behaves the best. We attribute this confliction of behavior to the fact that MNIST is a much simpler dataset than CIFAR-100. Commonly, we use asynchronous algorithms to save waiting time for faster UEs and hope that the convergence performance will not be affected by the update staleness. This only works when the dataset is simple and easy to train. Therefore, as we can see in Fig.~\ref{fig:exp_1}, for the MNIST dataset with a two-layer DNN model, the asynchronous algorithms does behave the best, semi-synchronous algorithms is the second, and synchronous algorithms behave the worst. However, when it comes to the CIFAR-100 dataset with the LeNet-5 model, which is a much larger dataset with a much more complicated model, it is hard for the asynchronous algorithms to convergence. In this case, semi-synchronous algorithms behave the best. This evaluation performance verifies our theoretical result that a proper semi-synchronous algorithm not only mitigates the straggler problem that happened in synchronous algorithms, but also bounds the staleness caused by the stragglers, thereby ensuring the convergence of the learning process. Meanwhile, it is clear that PFL algorithms converge much faster than FL algorithms. This result is derived from the fact the PFL algorithms is designed to adapt and converge fast to new datasets.

Most importantly, we find that compared with Fig.~\ref{fig:exp_1}, the convergence performance shown in Fig.~\ref{fig:exp_2} is poorer. This is because the relative participation frequencies of UEs in Fig.~\ref{fig:exp_2} is not equalized. Since the UEs are uniformly distributed in the cell, their distances to the central server are different. The UEs with longer distances to the server have to transmit its gradients for a longer time to reach the server. Therefore, these UEs are naturally slower than the others, leading to smaller $\eta$ to participate in the global model updates. Given that the datasets among UEs are heterogenous, the less participation of long distance UEs will lead to inadequate training on these UEs, making the global model convergence performance poorer than the ones shown in Fig.~\ref{fig:exp_1}.

As for the shakespeare dataset, we find that all the conclusions about the comparisons between the 6 algorithms drawn from the above two datasets still stand.

The comparison between FedAvgS$^2$, FedProxS$^2$ and PerFedS$^2$ using the MNIST and Shakespeare datasets is shown in Fig.~\ref{fig:exp_8}. From the figure it is obvious that PerFedS$^2$ outperforms the other two algorithms. This is reasonable since Per-FedAvg has already been verified in previous works to provide a better convergence performance, and PerFedS$^2$ is designed based on Per-FedAvg. Therefore, PerFedS$^2$ inherits this benefit.

\vspace{0.2cm}
\subsubsection{Effect of the non-i.i.d. level $l$}

\begin{figure*}[!t]
  \centering
  \subfloat[MNIST training loss]{
      \includegraphics[width=1.7in]{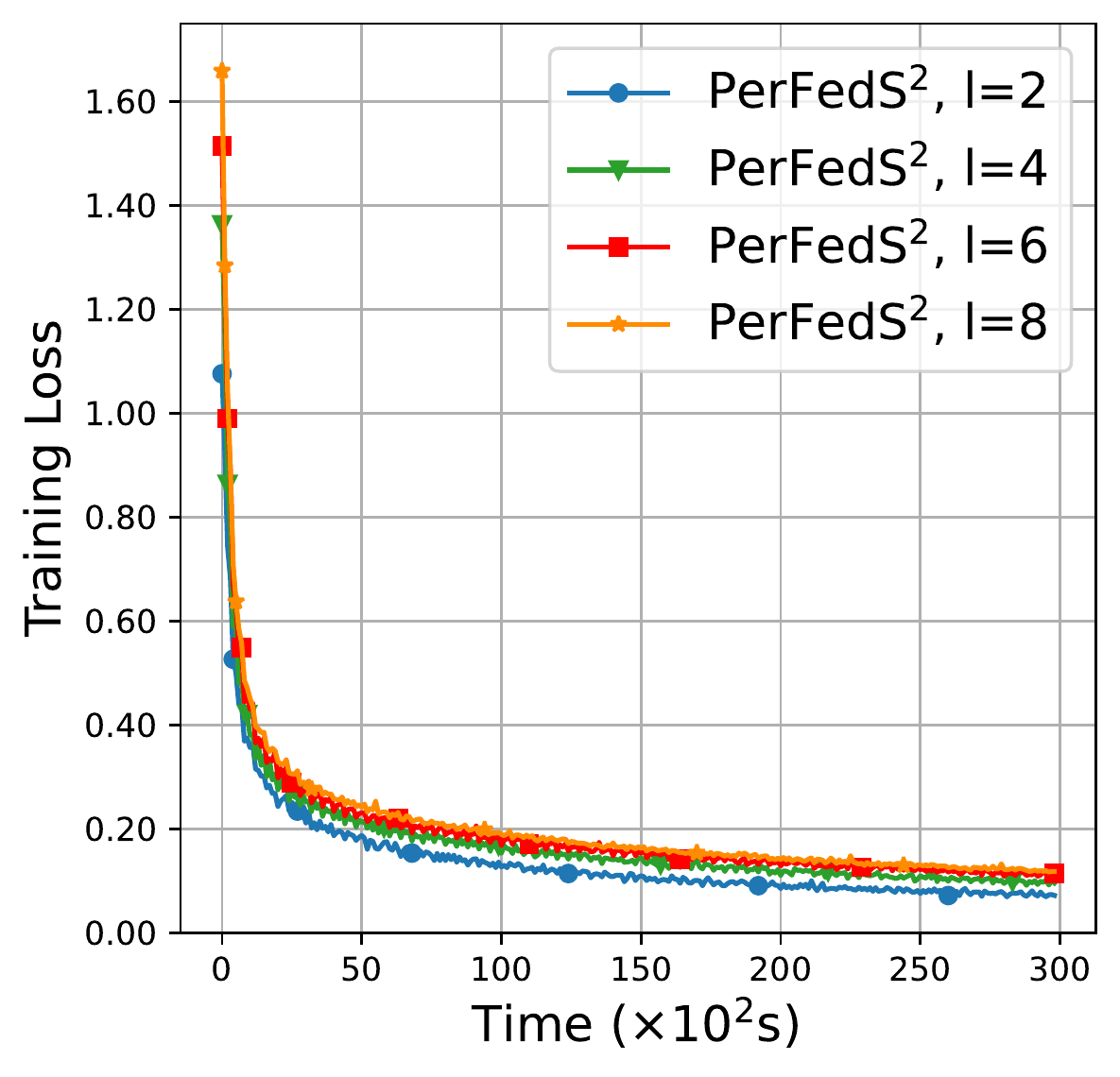}
      \label{fig_exp_3:subfig:a}}
  \subfloat[MNIST test accuracy]{
      \includegraphics[width=1.7in]{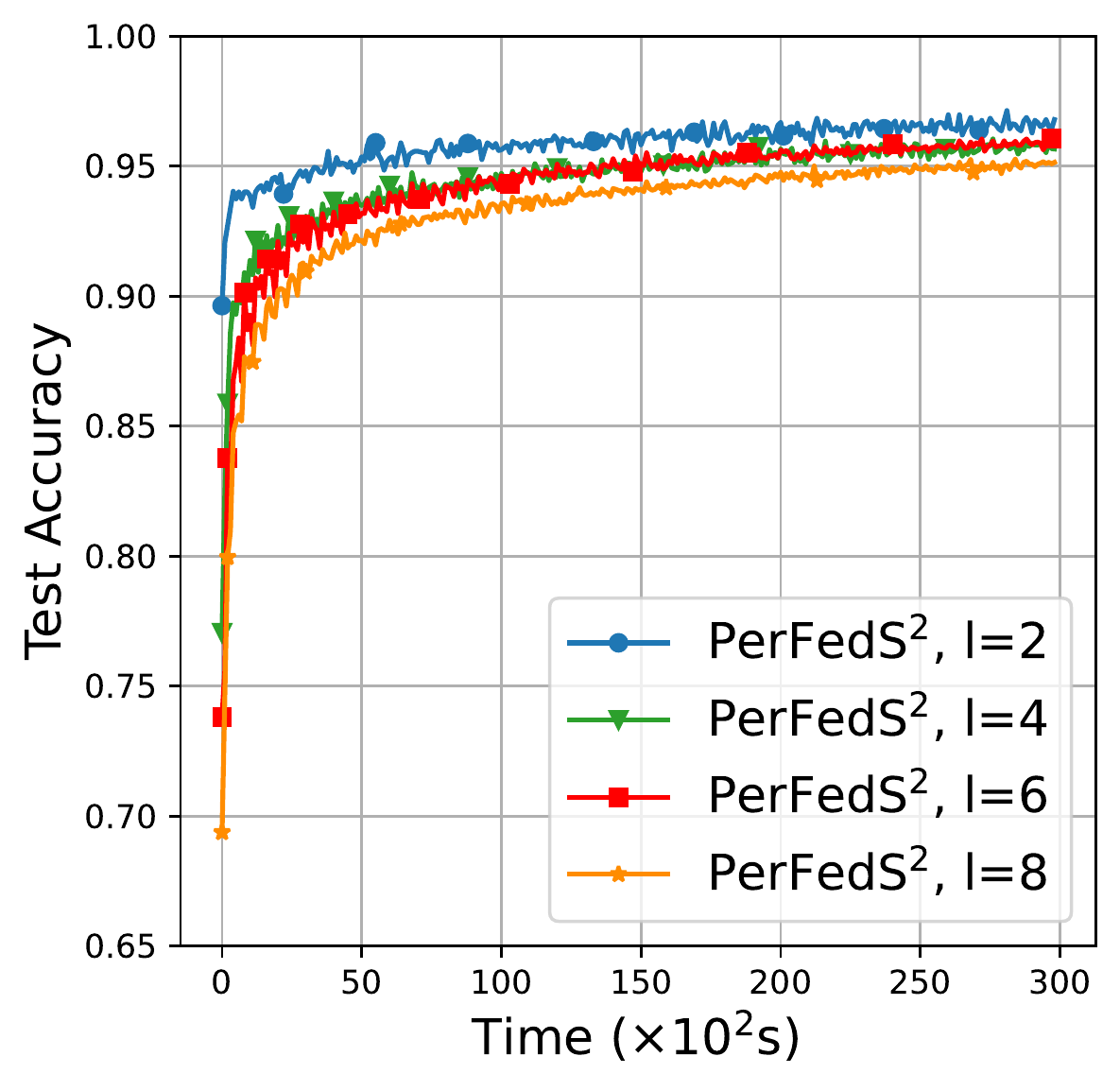}
      \label{fig_exp_3:subfig:b}}
  \subfloat[CIFAR-100 training loss]{
      \includegraphics[width=1.7in]{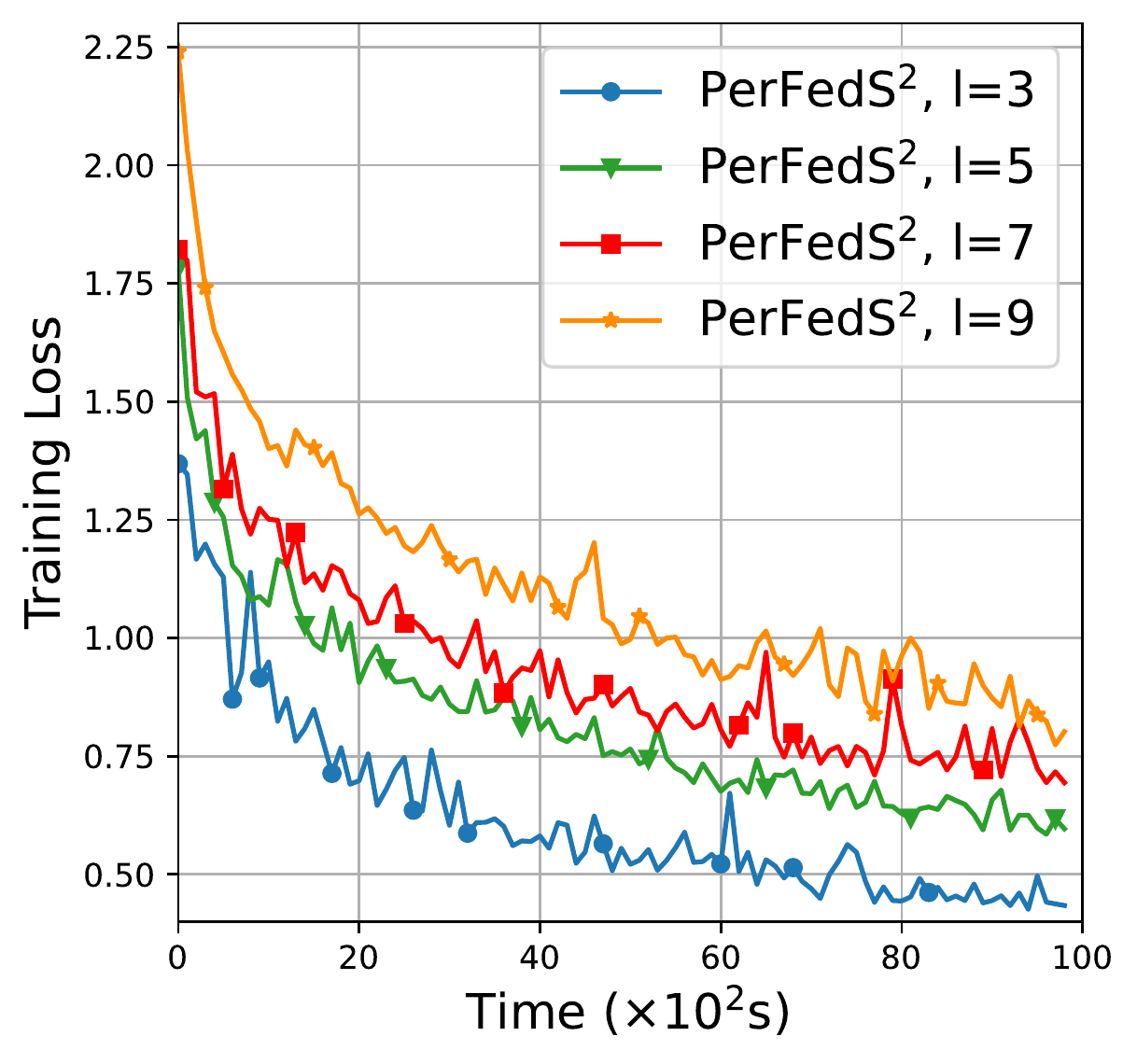}
      \label{fig_exp_3:subfig:c}}
  \subfloat[CIFAR-100 test accuracy]{
      \includegraphics[width=1.7in]{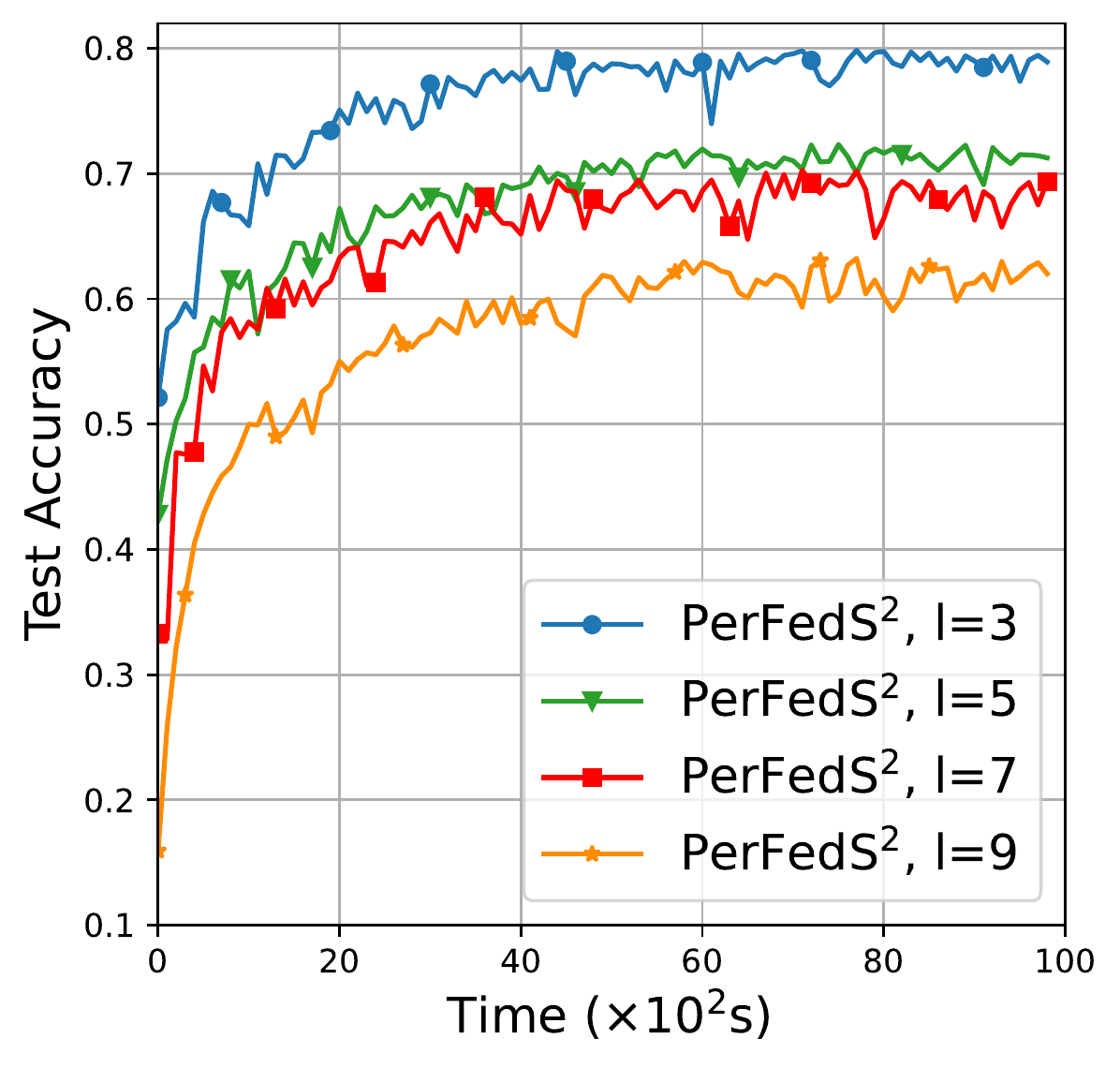}
      \label{fig_exp_3:subfig:d}}
  \caption{Convergence performance of PerFedS$^2$ with respect to the non-i.i.d level $l$ of data sampled from the MNIST and CIFAR-100 datasets. We compare the results when $l=2,4,6,8$ for data sampled from the MNIST dataset, and $l=3,5,7,9$ for data sampled from the CIFAR-100 dataset.}
  \label{fig:exp_3}
\end{figure*}

Fig.~\ref{fig:exp_3} shows the evaluation results of PerFedS$^2$ under different non-i.i.d. levels. It is obvious that for both datasets, the higher the heterogenous level is, the worse the convergence performances are. These results are natural and in line with the laws of theory.

\vspace{0.2cm}
\subsubsection{Effect of the number of participants in each round $A$}

\begin{figure*}[!t]
  \centering
  \subfloat[MNIST training loss]{
      \includegraphics[width=1.7in]{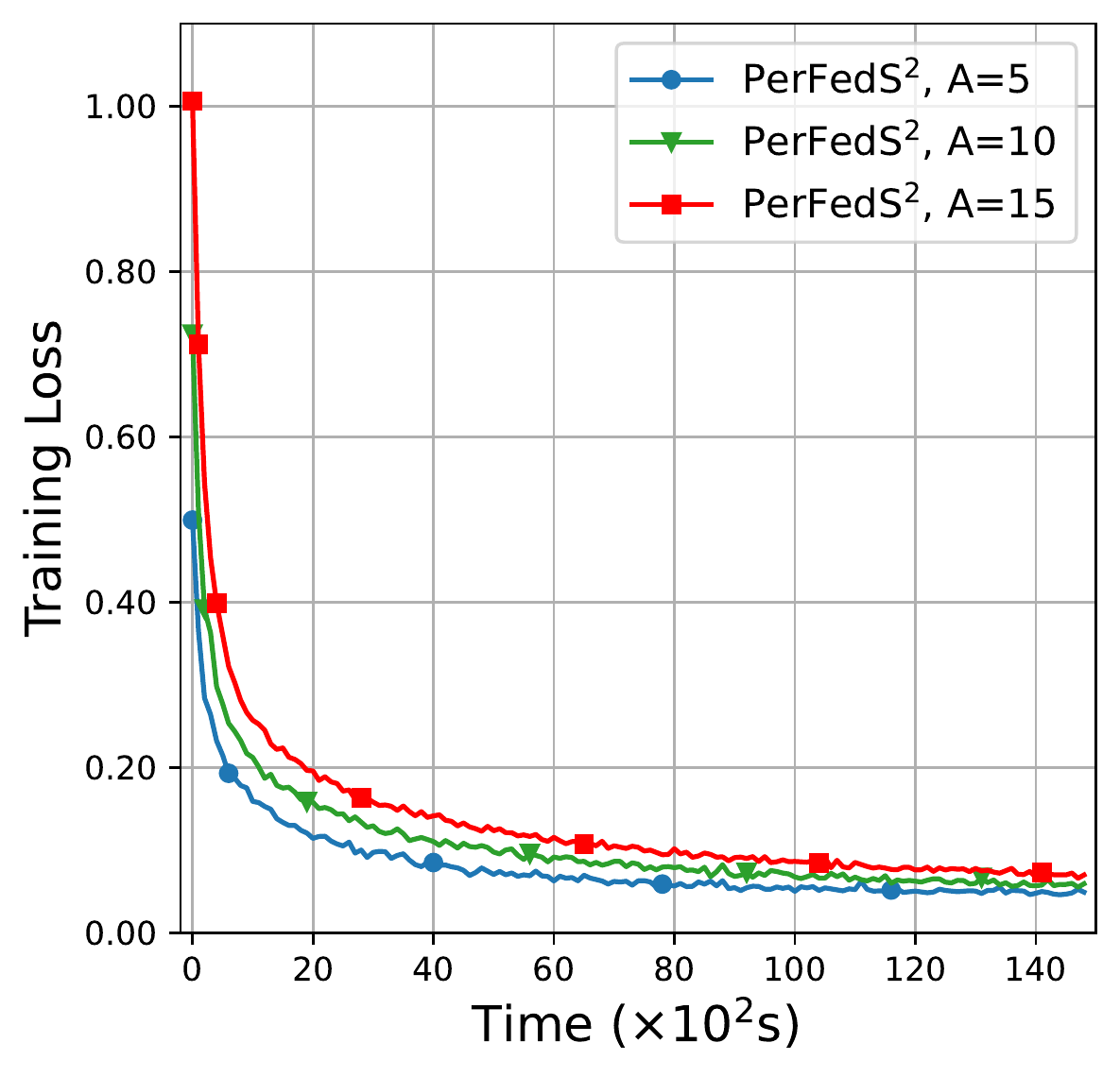}
      \label{fig_exp_4:subfig:a}}
  \subfloat[MNIST test accuracy]{
      \includegraphics[width=1.7in]{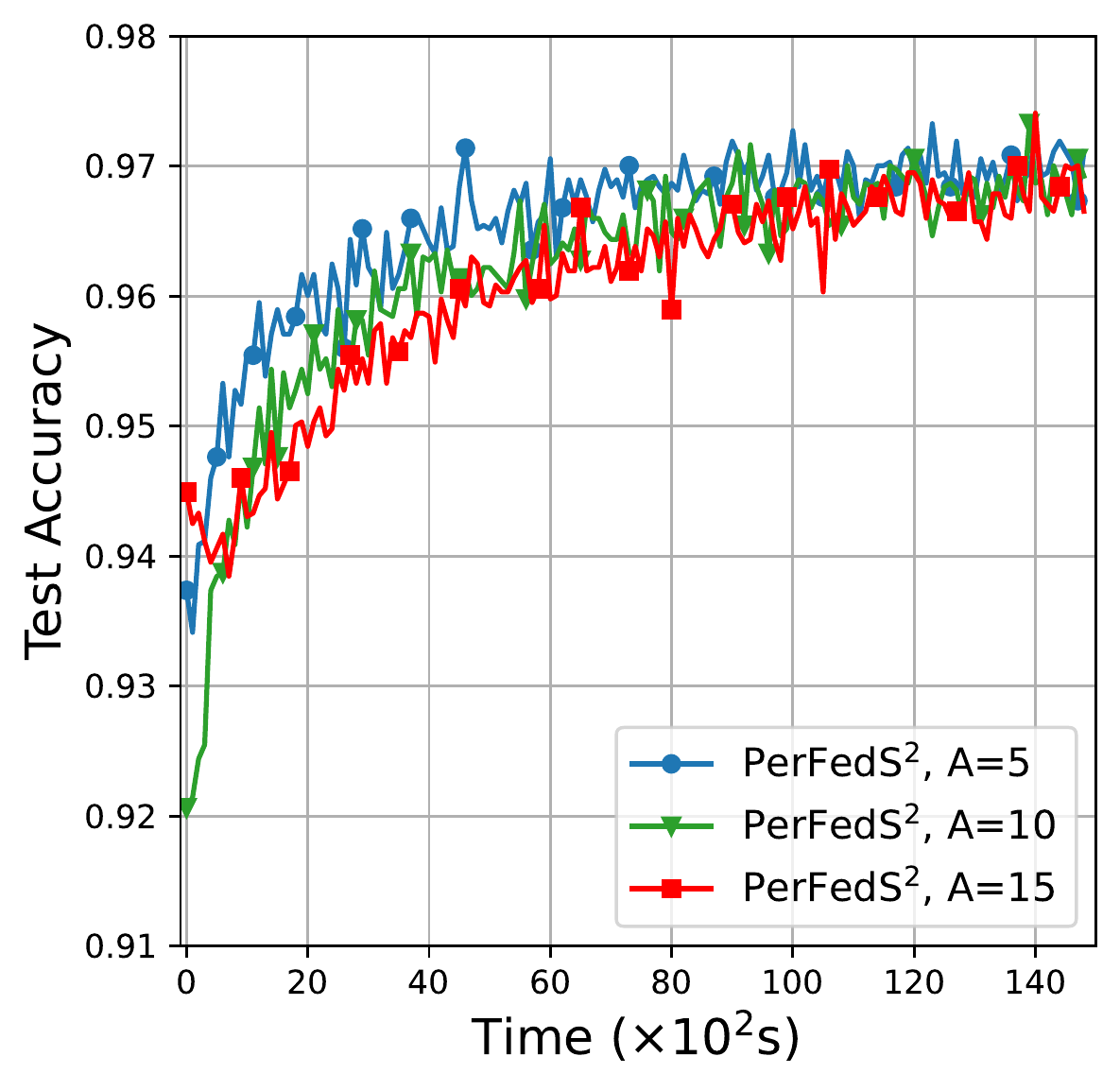}
      \label{fig_exp_4:subfig:b}}
  \subfloat[CIFAR-100 training loss]{
      \includegraphics[width=1.7in]{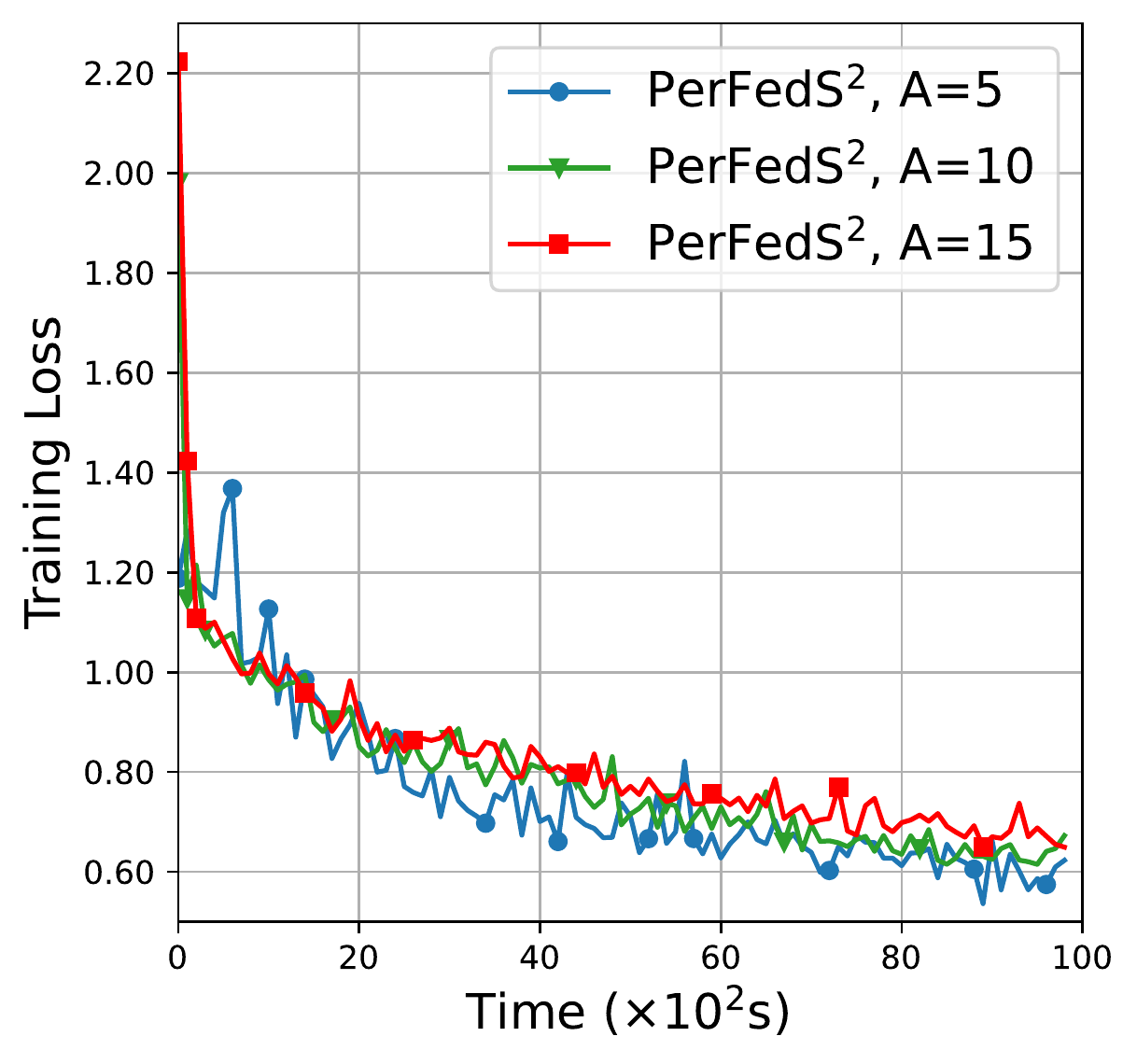}
      \label{fig_exp_4:subfig:c}}
  \subfloat[CIFAR-100 test accuracy]{
      \includegraphics[width=1.7in]{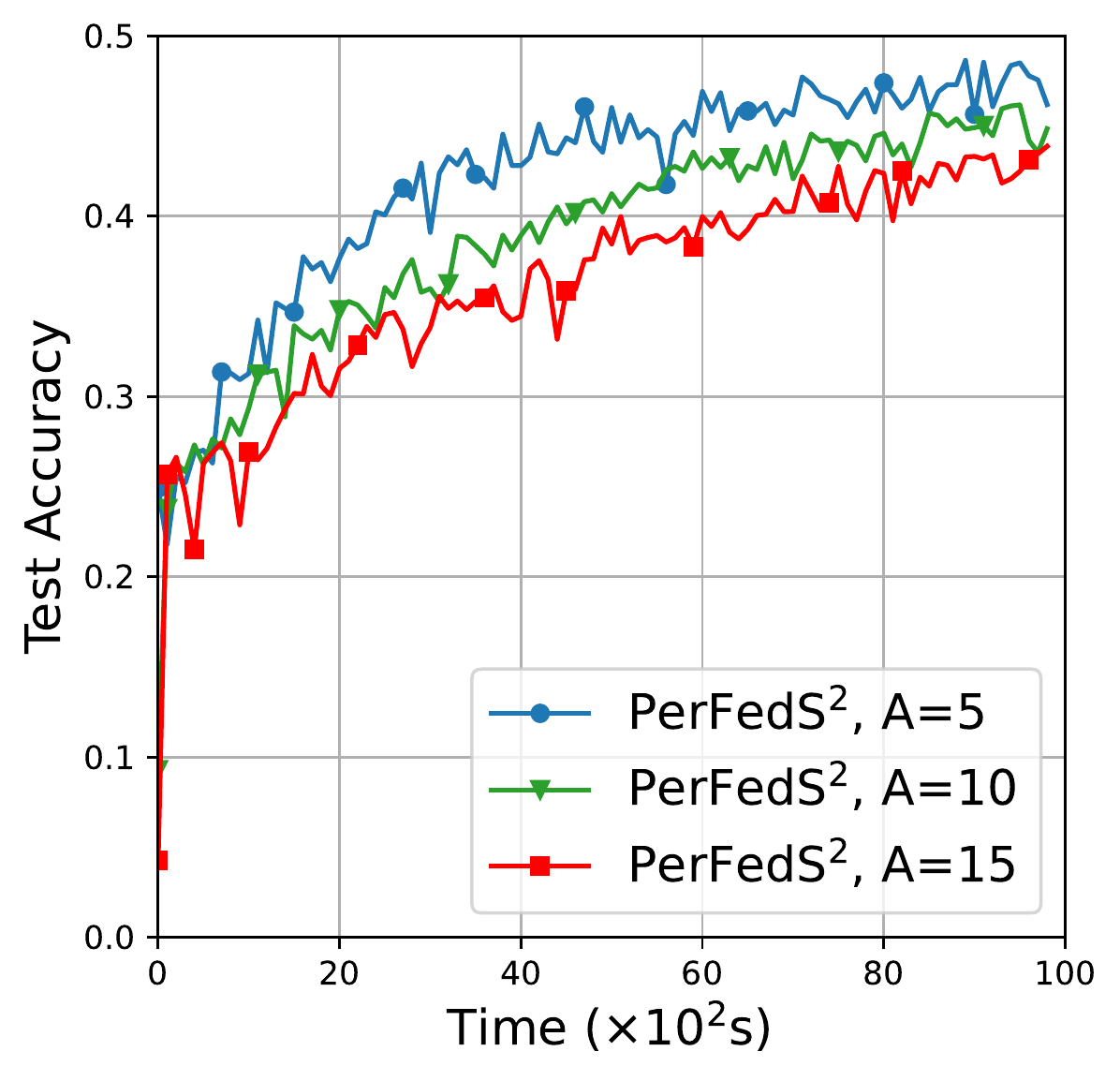}
      \label{fig_exp_4:subfig:d}}
  \caption{Convergence performance of PerFedS$^2$ with respect to the number of UEs $A$ that participate in the global model update in each round using MNIST and CIFAR-100 datasets. In this case, $\eta_1=\eta_2=\dots=\eta_n$. Meanwhile, we compare the results when $A=5,10,15$.}
  \label{fig:exp_4}
\end{figure*}

\begin{figure*}[!t]
  \centering
  \subfloat[MNIST training loss]{
      \includegraphics[width=1.7in]{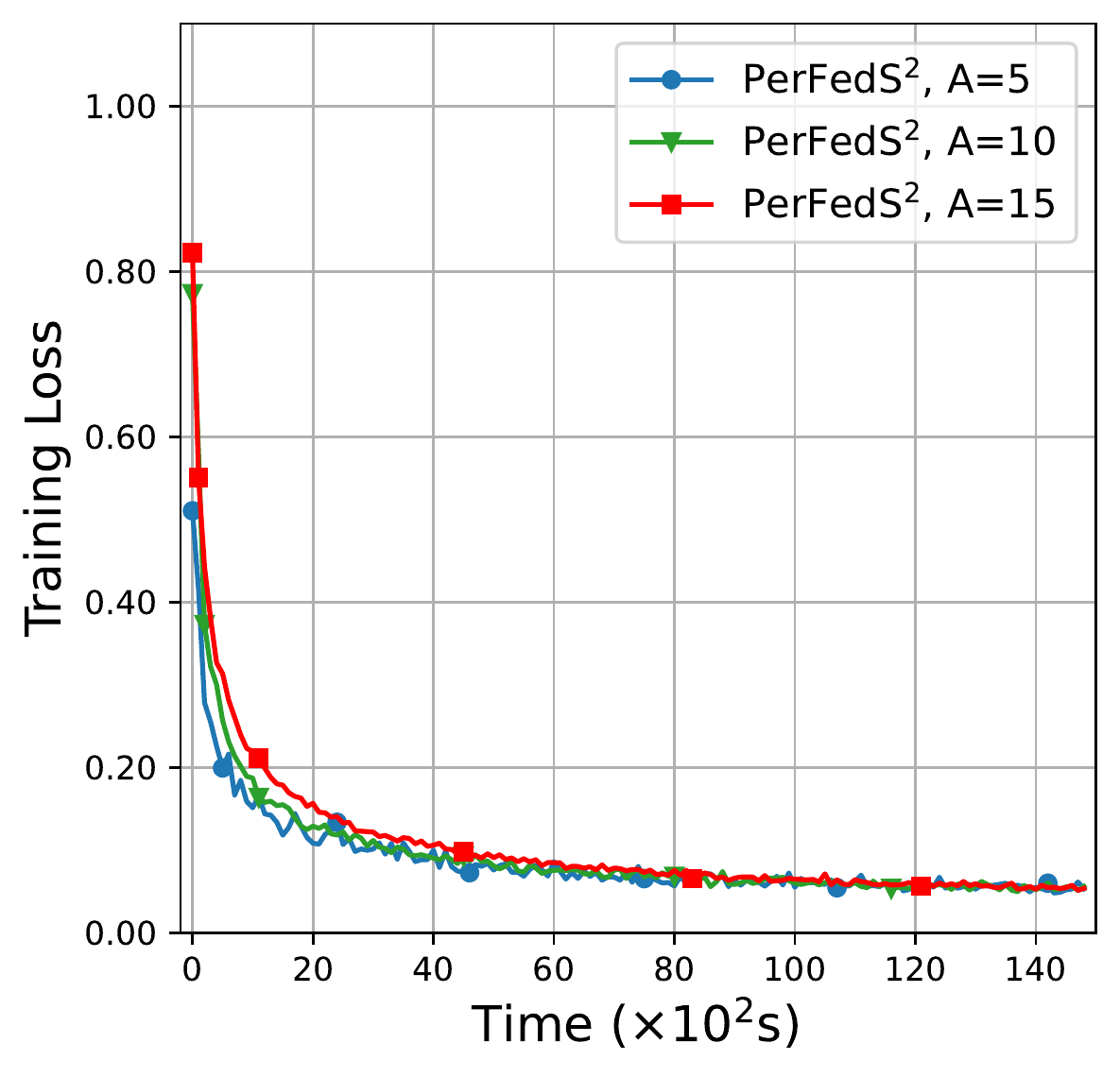}
      \label{fig_exp_5:subfig:a}}
  \subfloat[MNIST test accuracy]{
      \includegraphics[width=1.7in]{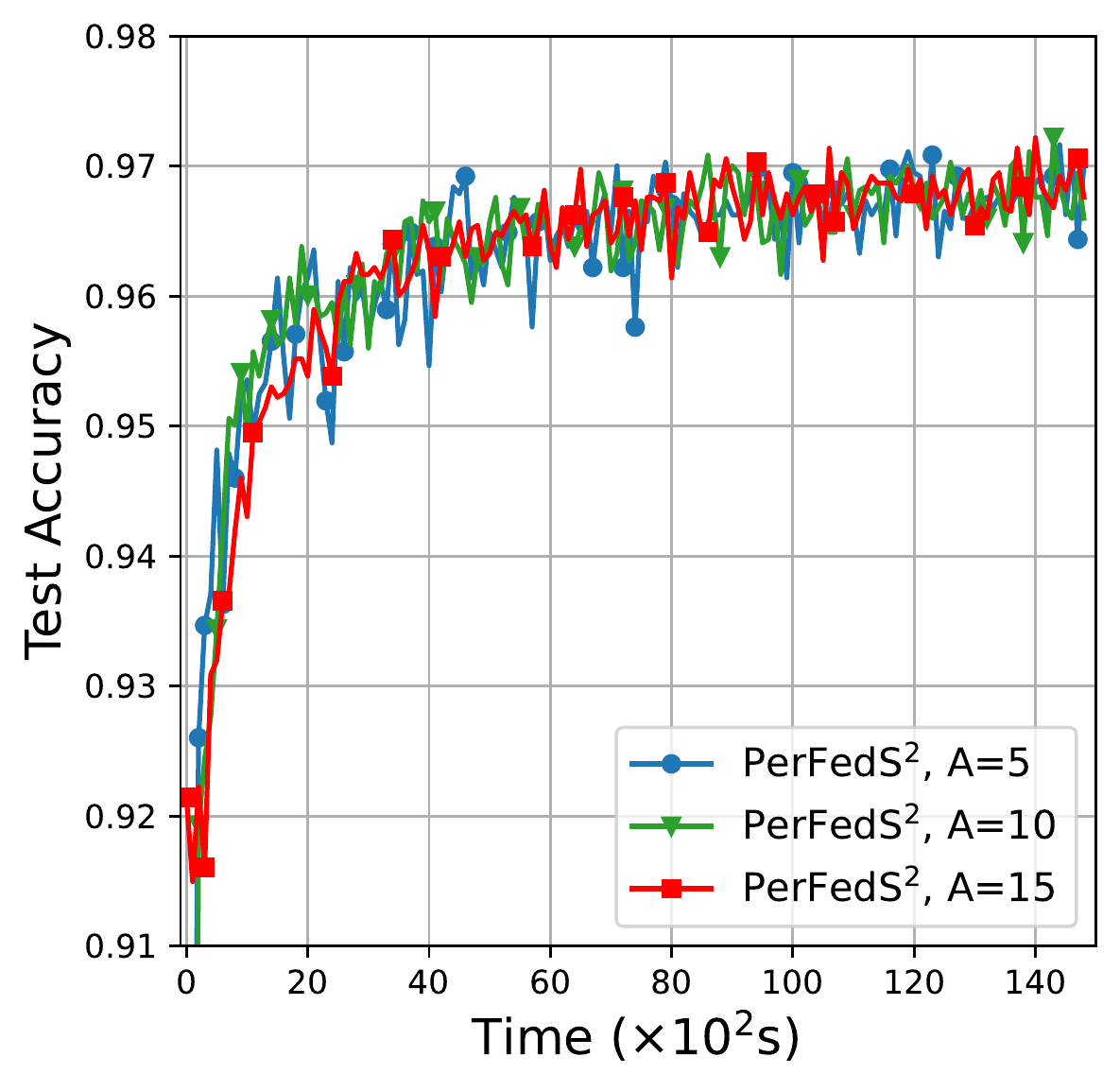}
      \label{fig_exp_5:subfig:b}}
  \subfloat[CIFAR-100 training loss]{
      \includegraphics[width=1.7in]{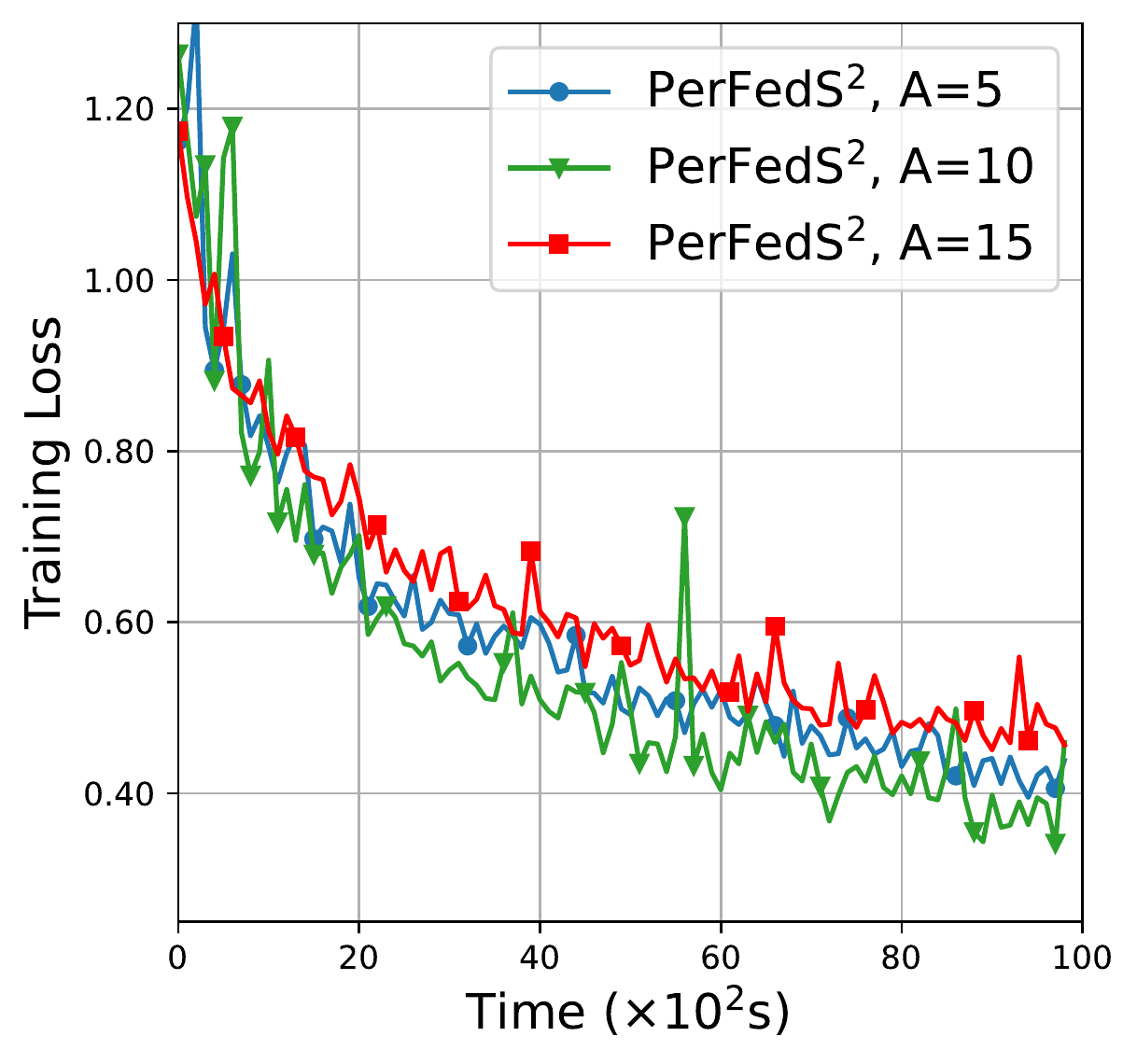}
      \label{fig_exp_5:subfig:c}}
  \subfloat[CIFAR-100 test accuracy]{
      \includegraphics[width=1.7in]{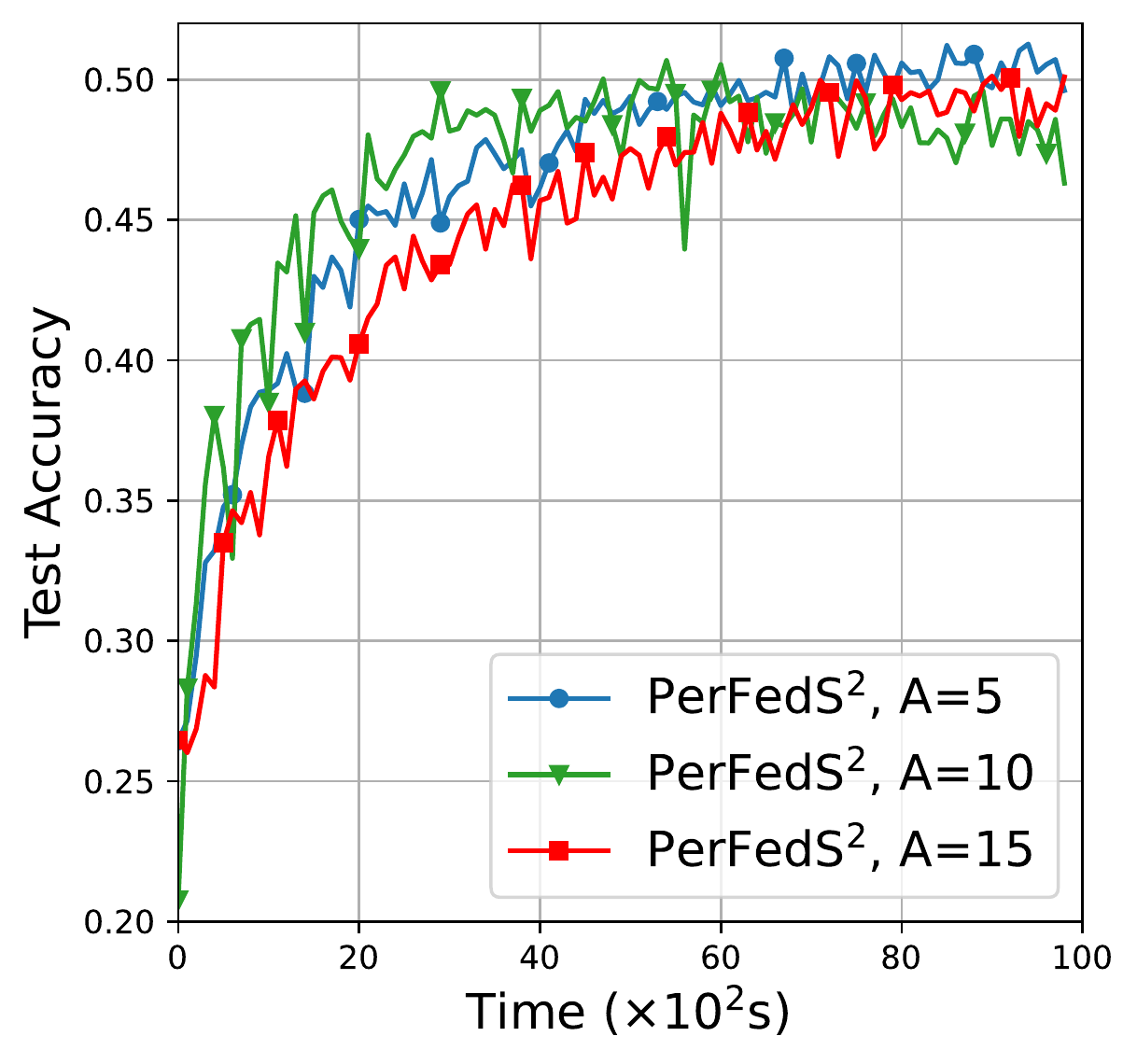}
      \label{fig_exp_5:subfig:d}}
  \caption{Convergence performance of PerFedS$^2$ with respect to the number of UEs $A$ that participate in the global model update in each round using MNIST and CIFAR-100 datasets. In this case, the distance from UEs to the server obeys the random distribution from 0 to 200 m. Meanwhile, we compare the results $A=5,10,15$.}
  \label{fig:exp_5}
\end{figure*}

Fig.~\ref{fig:exp_4} and Fig.~\ref{fig:exp_5} show the convergence performance of PerFedS$^2$ with respect to different number of participation UEs $A$ in each round, where Fig.~\ref{fig:exp_4} is under the case that all UEs have the same $\eta_i$, whereas Fig.~\ref{fig:exp_5} is under the case that the $\eta_i$ of each UE is determined by its distance to the central server that follows a random distribution.

As for the MNIST dataset, the result shown in Fig.~\ref{fig:exp_4} and Fig.~\ref{fig:exp_5} indicates a situation that the larger number of participation UEs in each round, the poorer the convergence performance is. This conclusion is not always true, given that the relative participation frequency vector $\eta=[\eta_i,\eta_2,\dots,\eta_n]$ in Fig.~\ref{fig:exp_5} is generated randomly according to the distances from UEs to the central server, and thus the optimal $A$ to minimize the overall training time is random. We can only conclude that in this very specific case of $\eta$, the larger number of participation UEs in each round, the better. Nevertheless, the benefits gained from a smaller value of $A$ is slight in Fig.~\ref{fig:exp_5}. This is reasonable because, the randomly generated $\eta$ may result in a scheduling pattern that degrades the influences caused by different number of participation UEs in each round.

However, as for the CIFAR-100 dataset, although Fig.~\ref{fig_exp_4:subfig:c} and~\ref{fig_exp_4:subfig:c} still indicate the same conclusion as that in the MNIST dataset, Fig.~\ref{fig_exp_5:subfig:c} and~\ref{fig_exp_5:subfig:d} indicate another situation where the convergence performance of PerFedS$^2$ wins when $A=10$. This result just verified the conclusion we mentioned above, that the conclusion obtained from the MNIST dataset is not always true. The result shown in Fig.~\ref{fig_exp_5:subfig:c} and~\ref{fig_exp_5:subfig:d} indicate a specific case when $A=10$ is approaching the optimal $A^*$.

\subsubsection{Effect of the staleness threshold $S$}

\begin{figure*}[!t]
  \centering
  \subfloat[MNIST training loss]{
      \includegraphics[width=1.7in]{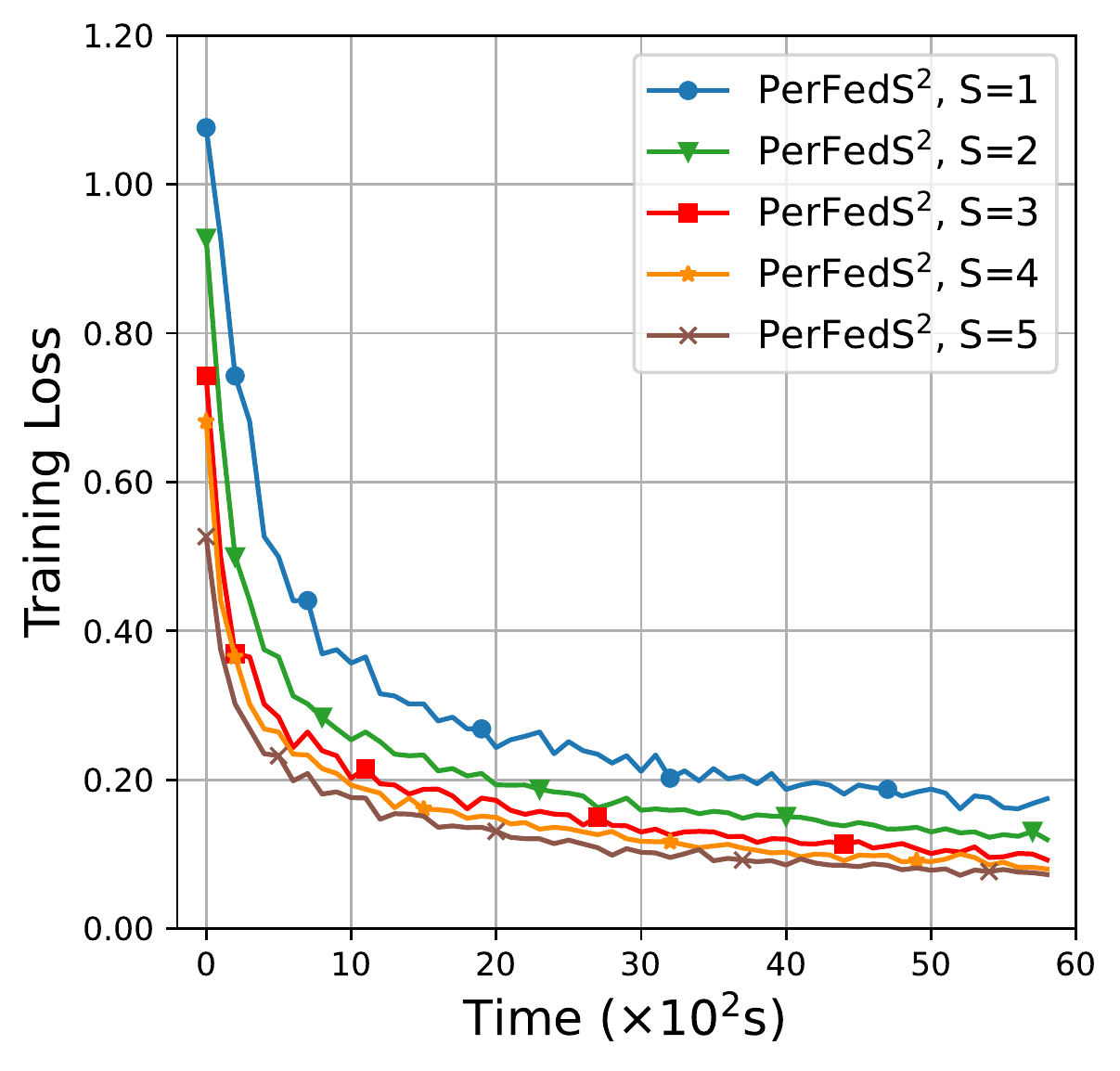}
      \label{fig_exp_6:subfig:a}}
  \subfloat[MNIST test accuracy]{
      \includegraphics[width=1.7in]{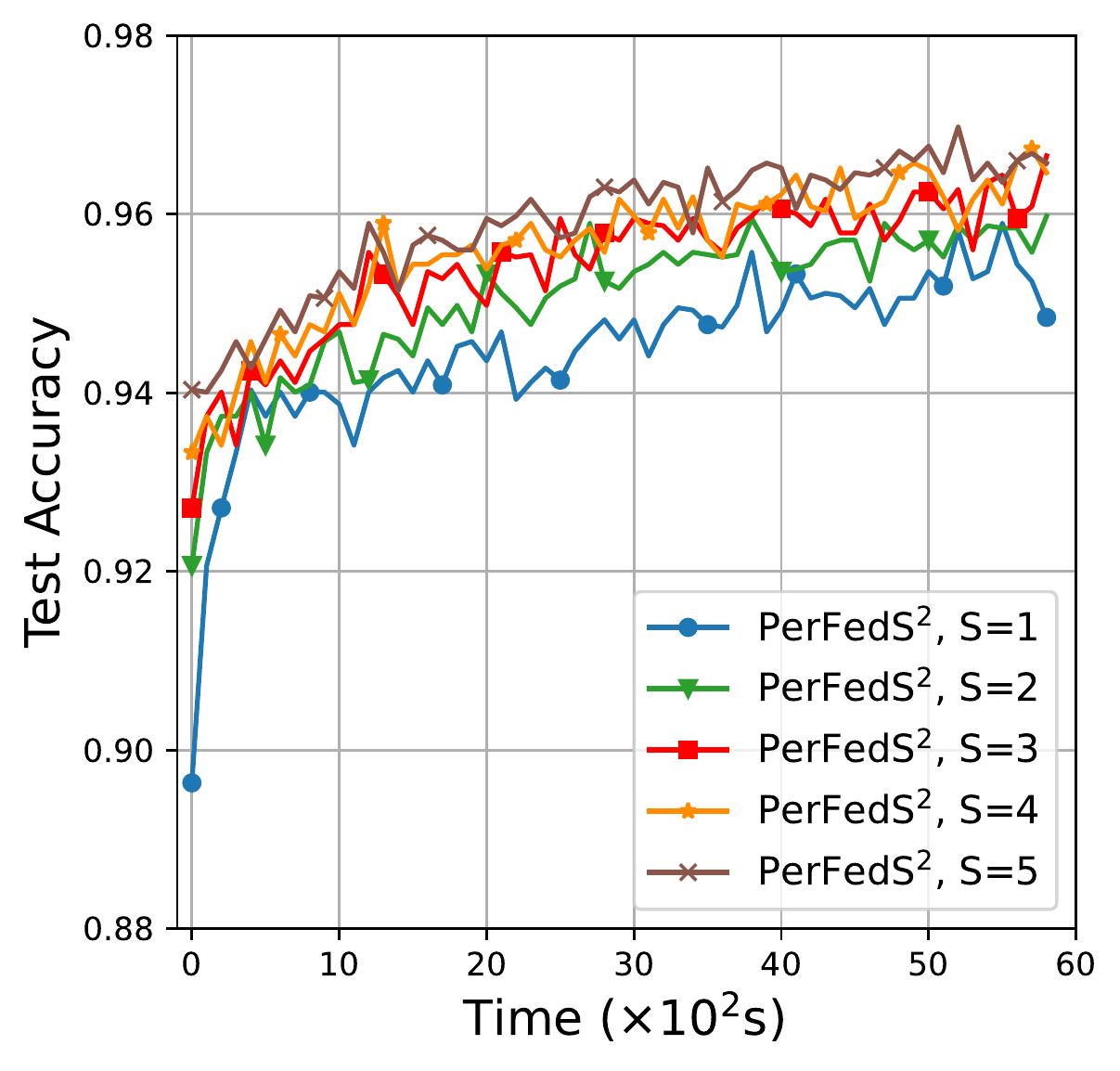}
      \label{fig_exp_6:subfig:b}}
  \subfloat[CIFAR-100 training loss]{
      \includegraphics[width=1.7in]{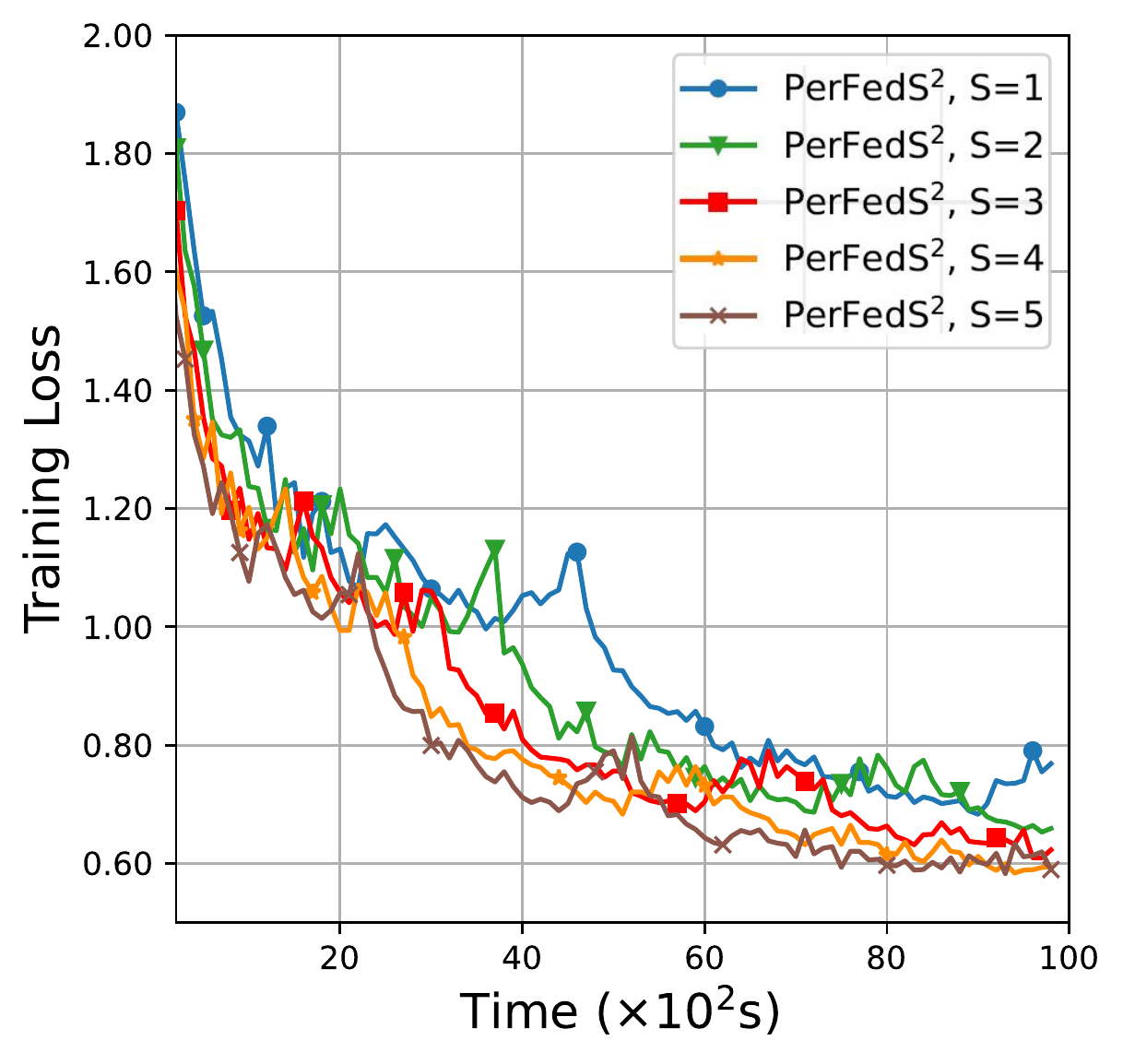}
      \label{fig_exp_6:subfig:c}}
  \subfloat[CIFAR-100 test accuracy]{
      \includegraphics[width=1.7in]{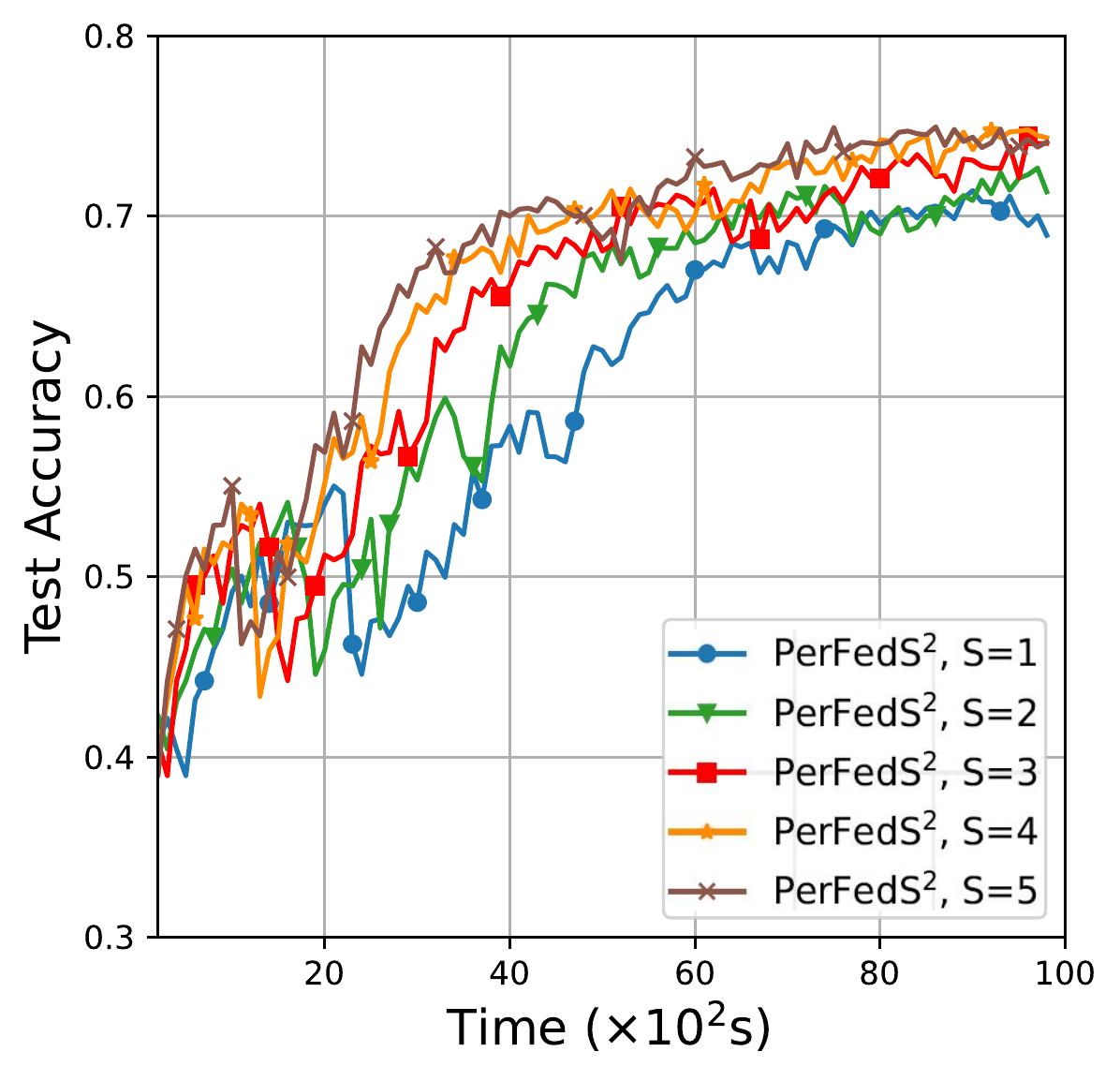}
      \label{fig_exp_6:subfig:d}}
  \caption{Convergence performance comparison of PerFedS$^2$ with respect to the staleness threshold $S$ using the MNIST and CIFAR-100 datasets. In this case, $\eta_1=\eta_2=\dots=\eta_n$, $A$=5. Meanwhile, we compare the results $S=1,2,3,4,5$.}
  \label{fig:exp_6}
\end{figure*}

Finally, we evaluate the effect of the staleness threshold $S$ on the convergence performance of PerFedS$^2$, where the results are shown in Fig.~\ref{fig:exp_6}. Here, in order to make the effect of $S$ more clear, we use the simpler setting when all UEs have the same $\eta_i$, and $A=5$. Therefore, when $S\geq 5$, all the scheduled UEs would arrive the server within $S$ rounds. Consequently, we study change of the total training time when $S=1,2,3,4,5$.

Note that in the theoretical analysis, we have the constraint that $\eta_i \geq S/K$. This constraint eliminates the situations when the staleness $\tau_k^i$ is larger than the staleness bound $S$, and thus no updates would be dropped by the central server. However, in practice, $\eta_i$ is determined by a number of elements, for example, the distances from UEs to the server or the transmit power of individual UEs. Therefore, in practice, the constraint $\eta_i \geq S/K$ cannot be always satisfied. When this happens to UE $i$, in order to keep $\eta_i$ constant, other UEs may have to wait until the updates from UE $i$ finally arrives the server, thereby prolonging the overall training time. This conclusion is verified through the results shown in Fig.~\ref{fig:exp_6}, where the larger $S$ is, the better the convergence performance PerFedS$^2$ has.

\section{Conclusions} \label{sec:7}

We have proposed a new semi-synchronous PFL algorithm over mobile edge networks, PerFedS$^2$, that not only mitigates the straggler problem caused by the synchronous training, but also ensures a convergent training loss that may not be guaranteed in the asynchronous training.
This is achieved by optimizing the joint bandwidth allocation and UE scheduling problem.
In order to solve such an optimization problem, we first have analysed the convergence rate of PerFedS$^2$, and have proved that there exist a convergent upper bound on the convergence rate.
Then, based on the convergence analysis, we have solved the optimization problem by decoupling it into two sub-problems: the bandwidth allocation problem and the UE scheduling problem.
For a given scheduling policy, the bandwidth allocations problem has been proved to have infinitely many solutions.
Meanwhile, based on the convergence analysis of PerFedS$^2$, the optimal UE scheduling policy can be determined using a greedy algorithm.
We have conducted extensive experiments to verify the effectiveness of PerFedS$^2$ in saving training time, compared with synchronous and asynchronous FL and PFL algorithms.

\section*{Appendix}

\noindent\textbf{Proof of Theorem 1}

\vspace{0.1cm}
Using Lemma~\ref{lem:1}, we have
\begin{align} \label{equ:L-lips}
  & F(w_{k+1})- F(w_k) \nonumber \\
  \leq & \langle\nabla F(w_k), w_{k+1}-w_k \rangle + \frac{L_F}{2} \|w_{k+1} - w_k\|^2 \nonumber \\
  = & - \left\langle\nabla F(w_k),\frac{\beta}{A} \sum_{i\in\mathcal{A}_k} \tilde{\nabla} F_i(w_{k-\tau_k^i}) \right\rangle \nonumber \\
  & + \frac{L_F}{2} \left\| \frac{\beta}{A}\sum_{i\in\mathcal{A}_k} \tilde{\nabla} F_i(w_{k-\tau_k^i}) \right\|^2.
\end{align}
From the above inequality, it is obvious that the key is to bound the term $\sum_{i\in\mathcal{A}_k} \tilde{\nabla} F_i(w_{k-\tau_k^i})$. Let
\begin{equation}
  \frac{1}{A} \sum_{i\in\mathcal{A}_k} \tilde{\nabla} F_i(w_{k-\tau_k^i}) = X+Y+\frac{1}{A}\sum_{i\in\mathcal{A}_k} \nabla F(w_{k-\tau_k^i}),
\end{equation}
where
\begin{align}
  X & = \frac{1}{A} \sum_{i\in\mathcal{A}_k} (\tilde{\nabla} F_i(w_{k-\tau_k^i}) - \nabla F_i(w_{k-\tau_k^i})), \nonumber \\
  Y & = \frac{1}{A} \sum_{i\in\mathcal{A}_k} (\nabla F_i(w_{k-\tau_k^i}) - \nabla F(w_{k-\tau_k^i})).
\end{align}
Our next step is to upper bound $\mathbb{E}[\|X\|^2]$ and $\mathbb{E}[\|Y\|^2]$ respectively. Recall the Cauchy-Schwarz inequality $\|\sum_{i=1}^{n} a_i b_i\|^2 \leq (\sum_{i=1}^{n}\|a_i\|^2)(\sum_{i=1}^{n}\|b_i\|^2)$, as for $X$, consider the Cauchy-Schwarz inequality with $a_i =\frac{1}{\sqrt{A}} (\tilde{\nabla} F_i(w_{k-\tau_k^i}) - \nabla F_i(w_{k-\tau_k^i}))$ and $b_i = \frac{1}{\sqrt{A}}$, we have
\begin{equation}
  \|X\|^2 \leq \frac{1}{A} \left( \sum_{i\in\mathcal{A}_k} \|\tilde{\nabla} F_i(w_{k-\tau_k^i}) - \nabla F_i(w_{k-\tau_k^i}) \|^2\right).
\end{equation}
Let $\mathcal{F}_k$ denote the information up to round $k$.
Given that the set of scheduled UEs $\mathcal{A}_k$ is selected according to their relative participation frequency $\eta_i$ ($i\in\mathcal{A}_k$), hence, by using Lemma~\ref{lem:2} along with the tower rule, we have
\begin{equation}
    \mathbb{E}[\|X\|^2] = \mathbb{E}[\mathbb{E}[\|X\|^2|\mathcal{F}_k]] \leq \sigma_F^2\sum_{i\in\mathcal{A}_k}\eta_i.
\end{equation}
Meanwhile, as for $Y$, consider the Cauchy-Schewarz inequality with $a_i = \frac{1}{\sqrt{A}} (\nabla F_i(w_{k-\tau_k^i}) - \nabla F(w_{k-\tau_k^i}))$ and $b_i = \frac{1}{\sqrt{A}}$, we have
\begin{equation}
  \|Y\|^2 \leq \frac{1}{A} \left( \sum_{i\in\mathcal{A}_k} \|\nabla F_i(w_{k-\tau_k^i}) - \nabla F(w_{k-\tau_k^i})\|^2 \right).
\end{equation}
In a similar way, the mean of $\|Y\|^2$ is the weighted average sum of $\mathbb{E}[\|Y\|^2|\mathcal{F}_k]$, where the weight is the relative participation frequency of UE $i\in\mathcal{A}_k$. By using Lemma~\ref{lem:3} along with the tower rule, we have
\begin{equation}\label{}
  \mathbb{E}[\|Y\|^2] = \mathbb{E}[\mathbb{E}[\|Y\|^2]|\mathcal{F}_k] \leq \gamma_F^2 \sum_{i\in\mathcal{A}_k}\eta_i.
\end{equation}
Now getting back to the inequality (\ref{equ:L-lips}), from the fact $\langle a, b\rangle = \frac{1}{2} (\|a\|^2 + \|b\|^2 - \|a-b\|^2)$, we have

\begin{align} \label{equ:complicated}
  & F(w_{k+1}) - F(w_k) \nonumber \\
  \leq & -\frac{\beta}{2} \|\nabla F(w_k)\|^2 - \frac{\beta}{2}\left\| \frac{1}{A} \sum_{i\in\mathcal{A}_k}\tilde{\nabla} F_i(w_{k-\tau_k^i}) \right\|^2 \nonumber \\
  & + \frac{\beta}{2} \left\| \nabla F(w_k) - X - Y - \frac{1}{A}\sum_{i\in\mathcal{A}_k} \nabla F(w_{k-\tau_k^i}) \right\|^2 \nonumber \\
  & + \frac{L_F \beta^2}{2} \left\| \frac{1}{A} \sum_{i\in\mathcal{A}_k}\tilde{\nabla} F_i(w_{k-\tau_k^i}) \right\|^2 \nonumber \\
  \leq & -\frac{\beta}{2} \|\nabla F(w_k)\|^2 + L_F\beta^2 \underbrace{\|X+Y\|^2}_{T_1} \nonumber \\
  & + \beta\underbrace{\left\|\nabla F(w_k) - \frac{1}{A}\sum_{i\in\mathcal{A}_k} \nabla F(w_{k-\tau_k^i})\right\|^2}_{T_2}  \nonumber \\
  & + (L_F\beta^2-\beta) \left\| \frac{1}{A} \sum_{i\in\mathcal{A}_k} \nabla F(w_{k-\tau_k^i}) \right\|^2.
\end{align}
Our next step is to estimate the upper bounds of $\mathbb{E}[T_1]$ and $\mathbb{E}[T_2]$, respectively.
As for $T_1$, we have
\begin{equation}
  \mathbb{E}[T_1] \leq  2\mathbb{E}[\|X\|^2] + 2\mathbb{E}[\|Y\|^2] = 2(\sigma_F^2 + \gamma_F^2).
\end{equation}
As for $T_2$, we have
\begin{align}
  T_2 = & \frac{1}{A^2} \left\|\sum_{i\in \mathcal{A}_k} (\nabla F(w_k) - \nabla F(w_{k-\tau_k^i}))\right\|^2 \nonumber \\
  \leq & \frac{1}{A} \sum_{i\in\mathcal{A}_k} \left\|\nabla F(w_k) - \nabla F(w_{k-\tau_k^i}) \right\|^2 \nonumber \\
  \leq & \frac{1}{A} \sum_{i\in \mathcal{A}_k} \left\| L_F (w_k - w_{k-\tau_k^i})\right\|^2 \nonumber \\
  \leq & \max_{i\in\mathcal{A}_k} \|L_F(w_k - w_{k-\tau_k^i})\|^2 \nonumber \\
  = & L_F^2\|(w_k - w_{k-\tau_k^\mu})\|^2,
\end{align}
where $\mu = \arg\max_{i\in\mathcal{A}_k}\|L_F(w_k - w_{k-\tau_k^i})\|^2$, the first inequality is obtained from the fact that $\|\sum_{i=1}^{n} a_i\|^2\leq n \sum_{i=1}^{n} \|a_i\|^2$, the second inequality is derived from Lemma~\ref{lem:1}, and the third inequality comes from the fact that $\frac{1}{n} \sum_{i=1}^{n} \|a_i\| \leq \max_{i}\|a_i\|$. It follows that
\begin{align} \label{equ:T_2}
  T_2 \leq & L_F^2\|w_k-w_{k-\tau_k^\mu}\|^2  \nonumber \\
  = & L_F^2 \left\| \sum_{j=k-\tau_k^\mu}^{k-1} (w_{j+1} - w_j)\right\|^2 \nonumber\\
  = & L_F^2\beta^2 \left\|\sum_{j=k-\tau_k^\mu}^{k-1} \frac{1}{A} \sum_{i\in\mathcal{A}_j} \tilde{\nabla} F_i(w_{j-\tau_j^i}) \right\|^2 \nonumber \\
  \leq & L_F^2 \beta^2 S \sum_{j=k-S}^{k-1} \left\|\frac{1}{A} \sum_{i\in\mathcal{A}_j} \tilde{\nabla} F_i(w_{j-\tau_j^i}) \right\|^2 \nonumber \\
  \leq & 2 L_F^2 \beta^2 S^2 \|X+Y\|^2  \nonumber \\
  & + 2 L_F^2 \beta^2 S^2 \left\|\frac{1}{A} \sum_{i\in\mathcal{A}_j} \nabla F(w_{j-\tau_j^i}) \right\|^2
\end{align}
Taking expectation on both sides of (\ref{equ:T_2}), we have
\begin{align}
  \mathbb{E}[T_2] \leq & 4 L_F^2\beta^2S^2(\sigma_F^2+\gamma_F^2) \sum_{i\in\mathcal{A}_k}\eta_i \nonumber \\
  & + 2L_F^2\beta^2 S^2 \mathbb{E}\left[\left\|\frac{1}{A} \sum_{i\in\mathcal{A}_k} \nabla F(w_{k-\tau_k^i}) \right\|^2\right].
\end{align}
Note that $\sum_{i\in\mathcal{A}_k} \eta_i =\sum_{i\in\mathcal{U}} \pi_k^i\eta_i$, we have
\begin{align}
  & (\sum_{i\in\mathcal{U}} \pi_k^i\eta_i)^2 \leq \sum_{i\in\mathcal{U}} (\pi_k^i)^2 \sum_{i\in\mathcal{U}} \eta_i^2 \nonumber \\
  & = \sum_{i\in\mathcal{U}}\pi_k^i \sum_{i\in\mathcal{U}}\eta_i^2 = A\sum_{i\in\mathcal{U}}\eta_i^2 \leq A,
\end{align}
where the first equation is derived from the fact that $(\pi_k^i)^2 = \pi_k^i$, the second equation is derived from the fact that $\sum_{i\in\mathcal{U}}\pi_k^i = A$, the last inequation is derived from the fact that $\eta_i<1$ and $\sum_{i\in\mathcal{U}}\eta_i=1$.
As a result, we have
\begin{equation}
  \sum_{i\in\mathcal{A}_k} \eta_i \leq \sqrt{A}.
\end{equation}
Now getting back to (\ref{equ:complicated}), we have
\begin{align}\label{equ:hh}
  & \mathbb{E}[F(w_{k+1})] - \mathbb{E}[F(w_k)] \nonumber \\
  \leq & -\frac{\beta}{2} \mathbb{E}[\|\nabla F(w_k)\|^2] \nonumber \\
  & + (2L_F\beta^2 + 4 L_F^2\beta^3S^2) (\sigma_F^2+\gamma_F^2) \sqrt{A} \nonumber \\
  & + (L_F \beta^2 - \beta + 2L_F^2\beta^2S^2) \mathbb{E}\left[\left\|\frac{1}{A} \sum_{i\in\mathcal{A}_j} \nabla F(w_{j-\tau_j^i}) \right\|^2\right]
\end{align}
Summarizing the inequality from $k=0$ to $k=K-1$, we have
\begin{align}
  & \mathbb{E}[F(w_{K})] - f(w_0) \nonumber \\
  \leq & -\frac{\beta}{2}\sum_{k=1}^{K} \mathbb{E}[\|\nabla F(w_k)\|^2]+ \nonumber \\
  & K(2L_F\beta^2 + 4 L_F^2\beta^3S^2)(\sigma_F^2+\gamma_F^2)\sqrt{A}+ \nonumber \\
  & \sum_{k=1}^{K} (L_F \beta^2 - \beta + 2L_F^2\beta^2S^2)\mathbb{E} \left[\left\|\frac{1}{A} \sum_{i\in\mathcal{A}_k} \nabla F(w_{k-\tau_k^i})\right\|^2\right] \nonumber \\
  \leq & -\frac{\beta}{2}\sum_{k=0}^{K-1} \mathbb{E}[\|\nabla F(w_k)\|^2] \nonumber \\
  & + K (2L_F\beta^2 + 4 L_F^2\beta^3S^2)(\sigma_F^2+\gamma_F^2) \sqrt{A},
\end{align}
where the last inequality is due to (\ref{equ:thm_constraint}). As a result, the desired result is obtained.

{
\footnotesize
\bibliographystyle{IEEEtran}
\bibliography{IEEEabrv,IEEEexample}
}

\begin{IEEEbiography}[{\includegraphics[width=1in,height=1.25in,keepaspectratio] {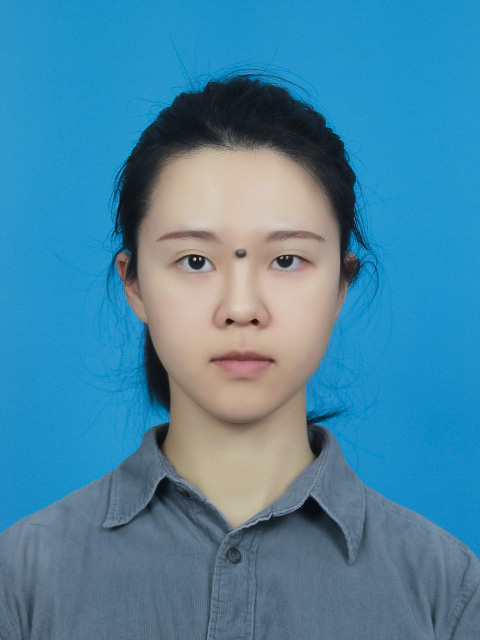}}]
{Chaoqun You}(S'13--M'20) is a postdoctoral research fellow in Singapore University of Technology and Design (SUTD). She received the B.S. degree in communication engineering and the Ph.D. degree in communication and information system from University of Electronic Science and Technology of China (UESTC) in 2013 and 2020, respectively. She was a visiting student at the University of Toronto from 2015 to 2017. Her current research interests include mobile edge computing, network virtualization, federated learning, meta-learning, and 6G.
\end{IEEEbiography}

\begin{IEEEbiography}[{\includegraphics[width=1in,height=1.25in,keepaspectratio] {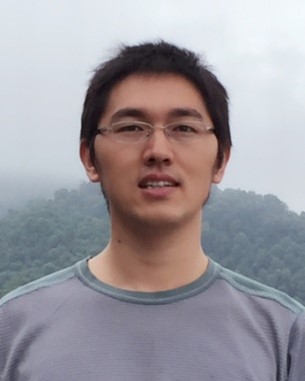}}]
{Daquan Feng} received the Ph.D. degree in information engineering from the National Key Laboratory of Science and Technology on Communications, University of Electronic Science and Technology of China, Chengdu, China, in 2015. From 2011 to 2014, he was a visiting student with the School of Electrical and Computer Engineering, Georgia Institute of Technology, Atlanta, GA, USA. After graduation, he was a Research Staff with State Radio Monitoring Center, Beijing, China, and then a Postdoctoral Research Fellow with the Singapore University of Technology and Design, Singapore. He is now an associate professor with the Shenzhen Key Laboratory of Digital Creative Technology, the Guangdong Province Engineering Laboratory for Digital Creative Technology, the Guangdong-Hong Kong Joint Laboratory for Big Data Imaging and Communication, College of Electronics and Information Engineering, Shenzhen University, Shenzhen, China. His research interests include URLLC communications, MEC, and massive IoT networks. Dr. Feng is an Associate Editor of IEEE COMMUNICATIONS LETTERS, Digital Communications and Networks and ICT Express.
\end{IEEEbiography}

\begin{IEEEbiography}[{\includegraphics[width=1in,height=1.25in,keepaspectratio] {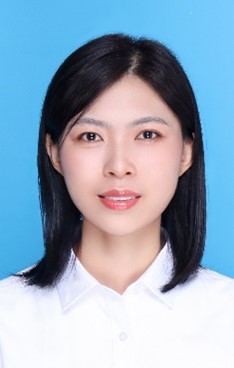}}]
{Kun Guo} (Member, IEEE) received the B.E. degree in Telecommunications Engineering from Xidian University, Xi'an, China, in 2012, where she received the Ph.D. degree in communication and information systems in 2019. From 2019 to 2021, she was a Post-Doctoral Research Fellow with the Singapore University of Technology and Design (SUTD), Singapore. Currently, she is a Zijiang Young Scholar with the School of Communications and Electronics Engineering at East China Normal University, Shanghai, China. Her research interests include edge computing, caching, and intelligence.
\end{IEEEbiography}

\begin{IEEEbiography}[{\includegraphics[width=1in,height=1.25in,keepaspectratio] {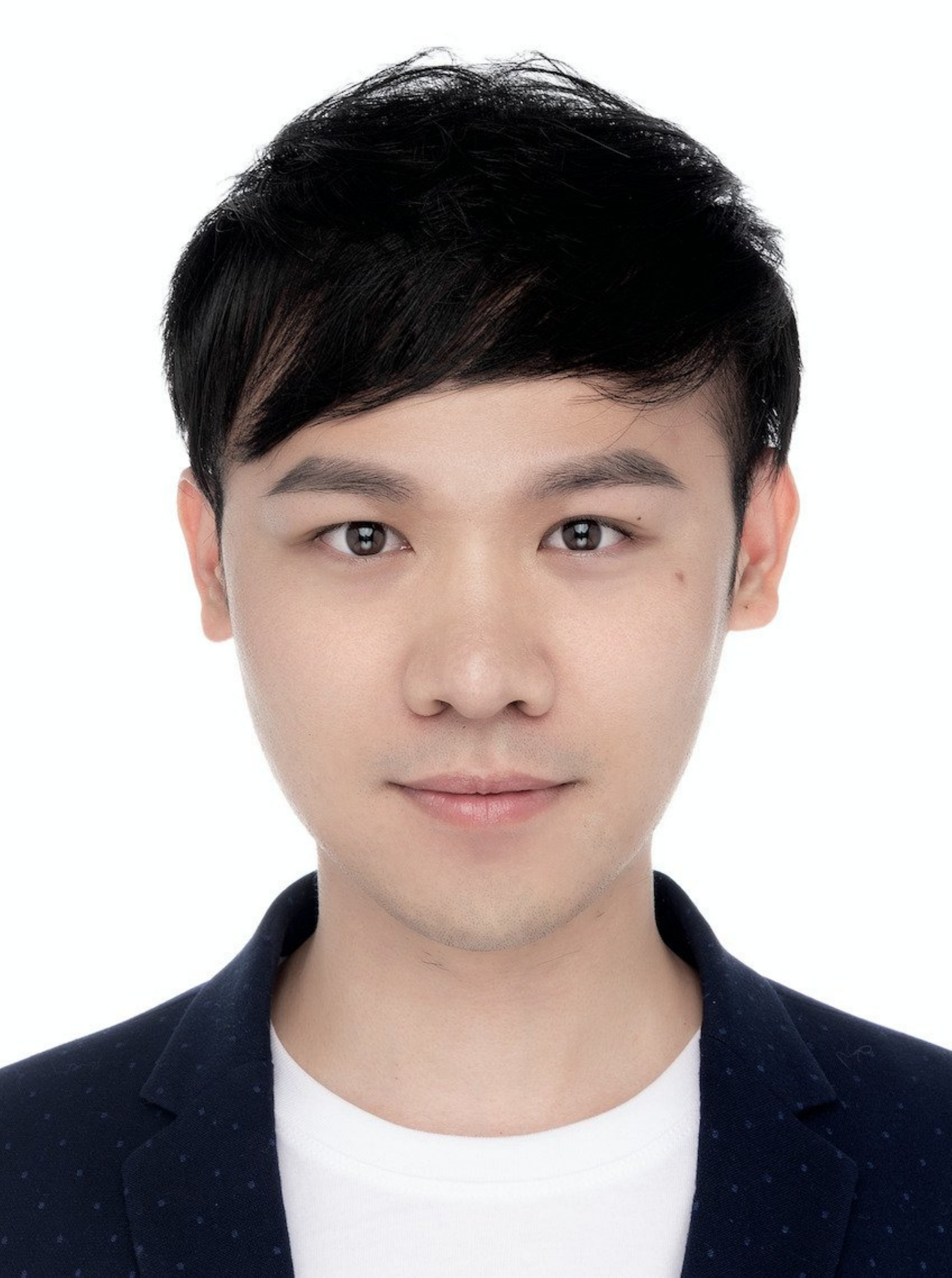}}]
{Howard H. Yang}(S'13--M'17) received the B.E. degree in Communication Engineering from Harbin Institute of Technology (HIT), China, in 2012, and the M.Sc. degree in Electronic Engineering from Hong Kong University of Science and Technology (HKUST), Hong Kong, in 2013. He earned the Ph.D. degree in Electrical Engineering from Singapore University of Technology and Design (SUTD), Singapore, in 2017. He was a Postdoctoral Research Fellow at SUTD from 2017 to 2020, a Visiting Postdoc Researcher at Princeton University from 2018 to 2019, and a Visiting Student at the University of Texas at Austin from 2015 to 2016. Currently, he is an assistant professor with the  Zhejiang University/University of Illinois at Urbana-Champaign Institute (ZJU-UIUC Institute), Zhejiang University, Haining, China. He is also an adjunct assistant professor with the Department of Electrical and Computer Engineering at the University of Illinois at Urbana-Champaign, IL, USA

Dr. Yang's research interests cover various aspects of wireless communications, networking, and signal processing, currently focusing on the modeling of modern wireless networks, high dimensional statistics, graph signal processing, and machine learning.
He serves as an editor for the {\scshape IEEE Transactions on Wireless Communications}.
He received the IEEE WCSP 10-Year Anniversary Excellent Paper Award in 2019 and the IEEE WCSP Best Paper Award in 2014.
\end{IEEEbiography}

\begin{IEEEbiography}[{\includegraphics[width=1in,height=1.25in,keepaspectratio] {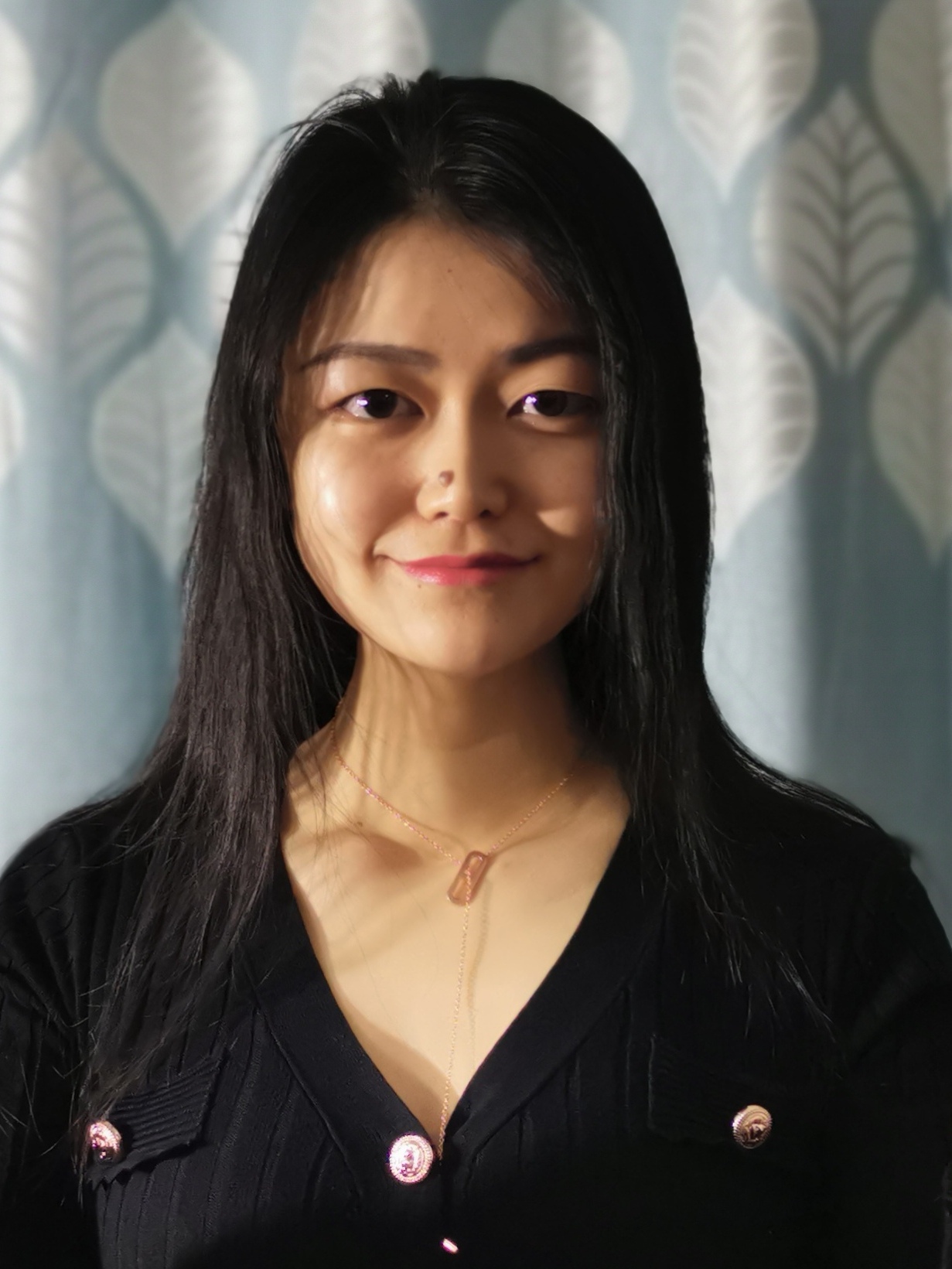}}]
{Chenyuan Feng} (S'16-M'21) received the B.E. degree in electrical and electronics engineering from the University of Electronic Science and Technology of China (UESTC), Chengdu, China, in 2016, and the Ph.D.\ degree in information system technology and design from Singapore University of Technology and Design (SUTD), Singapore, in 2021, respectively. Currently she has been doing postdoctoral work at Shenzhen Key Laboratory of Digital Creative Technology in Shenzhen University. Her research interests include edge computing, federated learning, graph signal processing and recommendation systems. She received the IEEE ComComAp Best Paper Award in 2021.
\end{IEEEbiography}

\begin{IEEEbiography}[{\includegraphics[width=1in,height=1.25in,keepaspectratio]
{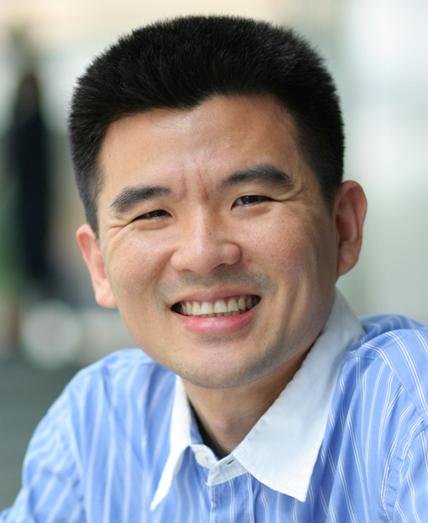}}]
{Tony Q.S. Quek}(S'98-M'08-SM'12-F'18) received the B.E.\ and M.E.\ degrees in electrical and electronics engineering from the Tokyo Institute of Technology in 1998 and 2000, respectively, and the Ph.D.\ degree in electrical engineering and computer science from the Massachusetts Institute of Technology in 2008. Currently, he is the Cheng Tsang Man Chair Professor with Singapore University of Technology and Design (SUTD). He also serves as the Director of the Future Communications R\&D Programme, the Head of ISTD Pillar, and the Deputy Director of the SUTD-ZJU IDEA. His current research topics include wireless communications and networking, network intelligence, internet-of-things, URLLC, and 6G.

Dr.\ Quek has been actively involved in organizing and chairing sessions, and has served as a member of the Technical Program Committee as well as symposium chairs in a number of international conferences. He is currently serving as an Area Editor for the {\scshape IEEE Transactions on Wireless Communications} and an elected member of the IEEE Signal Processing Society SPCOM Technical Committee. He was an Executive Editorial Committee Member for the {\scshape IEEE Transactions on Wireless Communications}, an Editor for the {\scshape IEEE Transactions on Communications}, and an Editor for the {\scshape IEEE Wireless Communications Letters}.

Dr.\ Quek was honored with the 2008 Philip Yeo Prize for Outstanding Achievement in Research, the 2012 IEEE William R. Bennett Prize, the 2015 SUTD Outstanding Education Awards -- Excellence in Research, the 2016 IEEE Signal Processing Society Young Author Best Paper Award, the 2017 CTTC Early Achievement Award, the 2017 IEEE ComSoc AP Outstanding Paper Award, the 2020 IEEE Communications Society Young Author Best Paper Award, the 2020 IEEE Stephen O. Rice Prize, the 2020 Nokia Visiting Professor, and the 2016-2020 Clarivate Analytics Highly Cited Researcher. He is a Fellow of IEEE.
\end{IEEEbiography}

\end{document}